\documentclass[a4paper,fleqn]{cas-dc}

\usepackage{textcomp}
\usepackage{svg}
\usepackage{url}
\usepackage{hyperref}
\usepackage{cuted}
\usepackage[commandnameprefix=ifneeded, final]{changes}
\usepackage{booktabs}
\usepackage{array}
\usepackage{multirow}

\usepackage[acronym]{glossaries}

\glsdisablehyper

\usepackage{booktabs}
\usepackage{graphicx}   
\usepackage{amssymb}    
\usepackage{pifont}     
\usepackage{float}
\newcommand{\cmark}{\checkmark}
\newcommand{\xmark}{\ding{55}}

\usepackage{subcaption}
\usepackage{tabularx}
\usepackage{multirow}

\hypersetup{
    urlcolor=blue
}

\newcommand{\percentdonut}[3][0.2cm]{%
  \begin{tikzpicture}
    \pgfmathsetmacro{\angle}{#2*3.6}
    \fill[gray!20] (0,0) circle (#1);
    
    \ifdim#2pt>0pt
      \fill[#3] (0,0) -- (90:#1) arc (90:90-\angle:#1) -- cycle;
    \fi
    
    \fill[white] (0,0) circle ({#1*0.6});
  \end{tikzpicture}%
}

\newacronym{DL}{DL}{deep learning}
\newacronym{3DGS}{3DGS}{3D Gaussian Splatting}
\newacronym{CAD}{CAD}{Computer-Aided Design}
\newacronym{HDRI}{HDRI}{High Dynamic Range Image}
\newacronym{ML}{ML}{machine learning}
\newacronym{CNN}{CNN}{convolutional neural network}
\newacronym{CNNs}{CNNs}{convolutional neural networks}
\newacronym{YOLO}{YOLO}{You Only Look Once}
\newacronym{mAP}{mAP}{Mean Average Precision}
\newacronym{DR}{DR}{Domain Randomization}
\newacronym{DA}{DA}{Domain Adaptation}
\newacronym{GDR}{GDR}{Guided Domain Randomization}
\newacronym{sim-to-real}{sim-to-real}{Simulation-to-Reality}
\newacronym{real-to-sim}{real-to-sim}{Reality-to-Simulation}
\newacronym{Sim-to-Real}{Sim-to-Real}{Simulation-to-Reality}
\newacronym{iasset}{I-AsSET}{Industrial Assets for Sim-to-Real Evaluation and Transfer}

\newacronym{IoU}{IoU}{Intersection over Union}
\newacronym{PBR}{PBR}{Physically Based Rendering}
\newacronym{HMLV}{HMLV}{High-Mix, Low-Volume}
\newacronym{SDG}{SDG}{Synthetic Data Generation}
\newacronym{KLT}{KLT}{Kleinladungsträger}

\usepackage[numbers]{natbib}

\def\tsc#1{\csdef{#1}{\textsc{\lowercase{#1}}\xspace}}
\tsc{WGM}
\tsc{QE}

\begin{document}
\let\WriteBookmarks\relax
\def\floatpagepagefraction{1}
\def\textpagefraction{.001}

\shorttitle{Bidirectional Sim-Real Object Perception}    

\shortauthors{J.M. Araya-Martinez et~al.}  

\title [mode = title]{SynthRender and I-AsSET: Open-Source Framework and Dataset for Bidirectional Sim–Real Transfer in Industrial Object Perception}  



%



\author[1,2]{Jose Moises Araya-Martinez}[orcid=0009-0006-5507-9512]
\cormark[1]
\ead{araya.martinez@campus.tu-berlin.de}
\credit{Conceptualization, Methodology, Validation, Investigation, Resources, Data curation, Writing - original draft, Writing - review \& editing, Project administration}

\author[2]{Thushar Tom}
\fnmark[1] 
\credit{Methodology, Software, Investigation, Data curation, Writing - original draft, Writing - review \& editing, Visualization}

\author[2]{Adrián Sanchis Reig}
\fnmark[1] 
\credit{Methodology, Software, Investigation, Data curation, Writing - original draft, Writing - review \& editing, Visualization}

\author[2]{Pablo Rey Valiente}
\credit{Software, Investigation, Data curation, Writing - original draft, Visualization}

\author[3]{Jens Lambrecht}
\credit{Validation, Writing - review \& editing, Supervision, Funding acquisition}

\author[1]{Jörg Krüger}
\credit{Validation, Writing - review \& editing, Supervision, Funding acquisition}


\affiliation[1]{organization={Technical University Berlin, Industrial Automation Technology},
            city={Berlin}, 
            country={Germany}}

\affiliation[2]{organization={Mercedes-Benz AG, Future Manufacturing Technologies},
            city={Sindelfingen}, 
            country={Germany}}


\affiliation[3]{organization={Technical University Braunschweig, Institute for Cognitive Robotics},
            city={Braunschweig}, 
            country={Germany}}


\cortext[1]{Corresponding author}

\fntext[1]{Authors contributed equally to this work.}



\begin{abstract}
Object perception is fundamental for tasks such as robotic material handling and quality inspection. However, modern supervised deep-learning models require large annotated datasets for robust automation under semi-uncontrolled conditions; a major barrier for widespread deployment with proprietary industrial parts. We address this through an integrated framework combining synthetic data generation and structured empirical evaluation for systematic investigation of bidirectional sim-to-real transfer. Our method integrates 2D-to-3D \acrlong{real-to-sim} techniques for 3D asset creation from physical parts with programmatic \acrfull{GDR} via SynthRender, an open-source synthetic image generation framework. Structured ablation studies across multiple benchmarks quantify the impact of individual rendering design choices, yielding practical guidelines for data-efficient synthetic training. To support evaluation under realistic industrial conditions, we introduce \acrfull{iasset}, a 32-class dataset with diverse textures, intra-class variation, strong inter-class similarities, and 19{,}672 annotations, providing both CAD models and reconstructed meshes for bidirectional \acrshort{sim-to-real} benchmarking. Across three industrial benchmarks, the proposed framework achieves highly competitive performance, reaching 98.7\% mAP@50 on a public robotics dataset, 97.9\% mAP@50 on an automotive benchmark, and 95.1\% mAP@50 on \acrshort{iasset}.
\end{abstract}


\maketitle

\section{Introduction}\label{intro}

\replaced{Visual object perception is essential for robust automation of complex tasks in semi-uncontrolled industrial environments. Tasks such as robot-based bin-picking and box handling~\cite{toper2025leveraging}, as well as quality inspection~\cite{araya-martinez2025xai}, exhibit high automation potential. Recent foundational models enable training-free pose estimation~\cite{wen2024foundationpose} and semantic segmentation~\cite{Kirillov_2023_ICCV}, but they still require prior object detection to handle novel objects. Specifically, FoundationPose initializes both translation and rotation search from a 2D bounding box supplied by an external detector~\cite{wen2024foundationpose}, and the Segment Anything Model requires a point, box, or text prompt to specify which object instance to segment~\cite{Kirillov_2023_ICCV}. Object detection therefore functions as an upstream localization step that these foundation models depend on to operate on previously unseen objects, rather than a task independent from them. Modern detectors rely on supervised, data-intensive learning~\cite{Hussain2023, Huang_2025_CVPR}, contributing to a bottleneck for widespread industrial automation~\cite{simeth2024hmlv, Tremblay2018}. Therefore, this work targets data-centric, synthetic detector training as the enabling step to further perception pipelines that rely on 2D bounding boxes as a primary input modality.}{Visual object perception is essential for robust automation of complex tasks in semi-uncontrolled industrial environments. Tasks such as robot-based bin-picking and box handling~\cite{toper2025leveraging}, as well as quality inspection~\cite{araya-martinez2025xai}, exhibit high automation potential. Recent foundational models enable training-free pose estimation~\cite{wen2024foundationpose} and semantic segmentation~\cite{Kirillov_2023_ICCV}, but they still require prior object detection to handle novel objects. Modern detectors rely on supervised, data-intensive learning~\cite{Hussain2023, Huang_2025_CVPR}, contributing to a bottleneck for widespread industrial automation~\cite{simeth2024hmlv, Tremblay2018}.}

\added{Rendering synthetic data that reflects real-world features is a promising direction to alleviate this data bottleneck, yet closing the gap between simulated and real-world environments remains a major open challenge in computer vision and robotics~\cite{Tremblay2018}. \acrfull{sim-to-real} transfer refers to deploying simulation-trained models directly in the real world~\cite{Peng2018}, whereas \acrfull{real-to-sim} transfer reconstructs real-world scenes or assets into simulation to generate reality-anchored synthetic data~\cite{James2019}. Together, these complementary directions form bidirectional \acrshort{sim-to-real} transfer~\cite{Truong2021}. In this loop, real assets provide realistic geometry and appearance, while synthetic augmentation increases variability without incurring prohibitive acquisition and annotation costs. This work instantiates such a bidirectional loop for industrial object detection and introduces the following contributions:}

\begin{itemize}
    \item The \acrshort{iasset}: Designed for \acrshort{sim-to-real} benchmarking in semi-\hspace{0pt}uncontrolled industrial environments, \acrshort{iasset} contains \acrshort{CAD} models and 2D-to-3D reconstructed meshes for 32 objects to support structured evaluation of \acrshort{sim-to-real} workflows. Its test set comprises 508 high-resolution RGB-D real images annotated with 19{,}672 object-detection instances. \acrshort{iasset} also includes checkpoints of our best-performing models and multiple synthetic training sets of 4{,}000 domain-randomized images each, generated with SynthRender and corresponding to the configurations used to obtain our results. The dataset and associated assets are publicly available at
\href{https://huggingface.co/datasets/moiaraya/I-AsSET}{the I-AsSET repository}~\cite{araya_martinez_iris_2026}.

    \item Automated \acrshort{DA} Methods: We benchmark the \acrshort{real-to-sim} potential of existing 2D-to-3D reconstruction methods as \acrshort{DA} strategies, providing an alternative to manual adaptation of the geometry and appearance of 3D assets. Our experiments include GenAI-based 2D-to-3D generation~\cite{TRELLIS}, Gaussian Splatting for mesh generation~\cite{tobiasz2025meshsplats}, and texture inference for reality-anchored materials~\cite{meshyai2025}.

    \item Systematic \acrshort{DR} Ablations via SynthRender: We conduct structured ablation studies using SynthRender, an open-source, scriptable \acrshort{DR} engine built on BlenderProc~\cite{Denninger2023, blender}, to quantify the impact of individual rendering design choices across three benchmarks. Our experiments identify physics-based placement, exponential light sampling, RGB lighting, and material randomization as consistently beneficial factors for industrial \acrshort{sim-to-real} transfer. Across the evaluated settings, the results suggest that the construction of synthetic variability can influence transfer performance more strongly than dataset scale alone, and generalizes across the evaluated detector architectures. Our framework is publicly available at the
\href{https://github.com/Moiso/SynthRender}{SynthRender Repository}~\cite{araya2026synthrender}.

    \item Unified \acrshort{sim-to-real} Framework and Evaluation: We systematically integrate the 2D-to-3D \acrshort{DA} approaches and \acrshort{DR}-based synthesis into a unified methodological framework, and provide comprehensive ablation experiments to quantify the contribution of each component. The resulting evaluation establishes design guidelines for industrial \acrshort{sim-to-real} workflows and achieves strong performance on a public robotics dataset~\cite{horvath2022object, Zhu2025icra} and an automotive benchmark~\cite{martinez2024scap}.
\end{itemize}

The next section reviews the current state-of-the-art in industrial domain randomization and adaptation. \autoref{sec:method} describes the unified framework for 2D-to-3D generation techniques and their combination with the SynthRender framework. Also, the \acrshort{iasset} dataset is introduced. Subsequently, \autoref{sec:results} presents and discusses the results of multiple ablation studies. Finally, \autoref{sec:conclusions} draws conclusions and outlines directions for future work.

\section{Related Works}
\label{sec:related}

This section summarizes prior literature relevant to data-efficient perception dataset and methods. ~\autoref{sub:related:sim2real} compares \acrshort{sim-to-real} datasets with our proposed \acrshort{iasset} set. Moreover,~\autoref{sec:techniques_for_DA} introduces the concept of \acrlong{DA} with a focus on methods for which this technique demands low human overhead in terms of data acquisition, annotation effort and number of iterations. Furthermore,~\autoref{sec:generative_sdg} summarizes generative methods for synthetic data creation and outlines some of its current challenges. Lastly, ~\autoref{sub:s2r_pipelines} presents a structured comparison of existing \acrfull{SDG} pipelines and SynthRender.

\subsection{Datasets for Sim-to-Real Industrial Object Perception}
\label{sub:related:sim2real}

Benchmarking \acrshort{sim-to-real} transfer requires datasets that provide both synthetic training data and real\added{-world} test images under conditions representative of industrial deployment. Early efforts such as T-LESS~\cite{hodan2017tless} and the BOP benchmark~\cite{hodan2020bop} established the practice of pairing CAD-based synthetic training sets with annotated real\added{-world} test images, primarily targeting pose estimation. More recent datasets have extended this paradigm to object detection in cluttered scenes~\cite{deroovere2024dimo, huang2025xyzibd}, yet several dimensions relevant to industrial automation remain underrepresented: high inter-class visual ambiguity, strong lighting variability, and support for bidirectional \acrshort{sim-to-real} workflows that go beyond CAD-only synthesis.

\autoref{tab:dataset_comparison} situates existing datasets along these dimensions. For the real-image subset, we assess visual complexity through five criteria. Lighting variability: number of distinct illumination conditions present in the listed real images. Pose variability: angular coverage of object orientations. Inter-class ambiguity: proportion of class pairs sharing geometry or material properties. Material variation: number of distinct surface finish types represented. Background clutter: mean number of non-target objects visible per image. Scores reflect the authors' assessment based on these criteria, and are intended as qualitative, relative indicators to support high-level dataset selection rather than a precise quantitative benchmark.

\begin{table*}[hbt]
\centering
\caption{Comparison of datasets for sim-to-real industrial object perception. Special attention is given to public datasets with dynamic light and pose conditions, as well as a high degree of inter-class similarity. Possible scores across these categories are low \percentdonut[0.15cm]{25}{black}, medium \percentdonut[0.15cm]{50}{black}, high \percentdonut[0.15cm]{75}{black}, and very high \percentdonut[0.15cm]{100}{black} variability, relative to other state-of-the-art datasets. Also a high diversity in the data modalities can support future sim-to-real benchmarking with novel technologies, such as the bidirectional sim-real gap exploration offered by \acrshort{iasset}. The instances variable reports the total number of annotated object occurrences across the listed real-image subset.}

\label{tab:dataset_comparison}
\footnotesize
\setlength{\tabcolsep}{2.45pt}
\renewcommand{\arraystretch}{1.5}

\begin{tabularx}{\textwidth}{@{}l c c c c c c c c c c c c c c c c}
\toprule
& & &
\multicolumn{9}{c}{\textbf{\replaced{Real-Image Subset}{Test Set: Real Images}}} &
\multicolumn{5}{c}{\textbf{Train Set: Synthetic Images}} \\
\cmidrule(lr){4-12}
\cmidrule(lr){13-17}
& & &
\multicolumn{2}{c}{} &
\multicolumn{5}{c}{\textbf{Visual Complexity}} &
\multicolumn{2}{c}{\textbf{Image Quality}} &
\multicolumn{1}{c}{} &
\multicolumn{4}{c}{\textbf{Modalities}} \\
\cmidrule(lr){6-10}
\cmidrule(lr){11-12}
\cmidrule(lr){14-17}
\textbf{Dataset Name} &
\rotatebox{90}{\textbf{Availability}} &
\rotatebox{90}{\textbf{\# Classes}} &
\rotatebox{90}{\textbf{\# Images}} &
\rotatebox{90}{\textbf{Instances$^{*}$}} &
\rotatebox{90}{\textbf{Light Var.}} &
\rotatebox{90}{\textbf{Pose Var.}} &
\rotatebox{90}{{\makecell[l]{\textbf{Inter-Class}\\\textbf{Ambiguity}}}} &
\rotatebox{90}{{\makecell[l]{\textbf{Material}\\\textbf{Variation}}}} &
\rotatebox{90}{{\makecell[l]{\textbf{Background}\\\textbf{Clutter}}}} &
\rotatebox{90}{{\makecell[l]{\textbf{Highest RGB}\\\textbf{Resolution}}}} &
\rotatebox{90}{{\makecell[l]{\textbf{Depth Acc.}\\\textbf{at 0.5\,m}}}} &
\rotatebox{90}{\textbf{\# Images}} &
\rotatebox{90}{\makecell[l]{\textbf{PBR}\\\textbf{Materials}}} &
\rotatebox{90}{\textbf{RGB}} &
\rotatebox{90}{\textbf{Depth}} &
\rotatebox{90}{\textbf{3D CADs}} \\
\midrule
\textbf{T-LESS \cite{hodan2017tless}} &
Public & 30 &
48{,}960 & \replaced{108{,}432}{N/A} &
\percentdonut[0.15cm]{50}{black} & \percentdonut[0.15cm]{75}{black} & \percentdonut[0.15cm]{100}{black} &
\percentdonut[0.15cm]{25}{black} & \percentdonut[0.15cm]{50}{black} &
3264$\times$2448 & $\sim$1.2\,mm$^{**}$ &
50{,}000 & \checkmark$^{*}$ &
\checkmark & \checkmark & \checkmark \\
\textbf{SIP15-OD \cite{Zhu2025icra}} &
Proprietary & 15 &
321 & 877 &
\percentdonut[0.15cm]{75}{black} & \percentdonut[0.15cm]{75}{black} & \percentdonut[0.15cm]{50}{black} &
\percentdonut[0.15cm]{50}{black} & \percentdonut[0.15cm]{75}{black} &
4032$\times$3024 & --- &
$\leq$8{,}000 & \checkmark &
\checkmark & $\times$ & $\times$ \\
\textbf{Automotive \cite{martinez2024scap}} &
Proprietary & 3 &
75 & \replaced{225}{N/A} &
\percentdonut[0.15cm]{75}{black} & \percentdonut[0.15cm]{50}{black} & \percentdonut[0.15cm]{25}{black} &
\percentdonut[0.15cm]{75}{black} & \percentdonut[0.15cm]{50}{black} &
4320$\times$3240 & --- &
$\leq$1{,}000 & \checkmark &
\checkmark & $\times$ & $\times$ \\
\textbf{SORDI.ai \cite{abouakar2024sordi}} &
Public & 111 &
0 & 0 &
\percentdonut[0.15cm]{50}{black} & \percentdonut[0.15cm]{50}{black} & \percentdonut[0.15cm]{75}{black} &
\percentdonut[0.15cm]{50}{black} & \percentdonut[0.15cm]{75}{black} &
--- & --- &
1{,}191{,}893 & \checkmark &
\checkmark & $\times$ & $\times$ \\
\textbf{Robotics \cite{horvath2022object}} &
Public & 10 &
190 & 920 &
\percentdonut[0.15cm]{25}{black} & \percentdonut[0.15cm]{25}{black} & \percentdonut[0.15cm]{50}{black} &
\percentdonut[0.15cm]{25}{black} & \percentdonut[0.15cm]{25}{black} &
1280$\times$720 & --- &
$\leq$8{,}000 & $\times$ &
\checkmark & \checkmark & \checkmark \\
\textbf{DIMO \cite{deroovere2024dimo}} &
Public & 6 &
31{,}200 & \replaced{158{,}496}{N/A} &
\percentdonut[0.15cm]{25}{black} & \percentdonut[0.15cm]{75}{black} & \percentdonut[0.15cm]{75}{black} &
\percentdonut[0.15cm]{50}{black} & \percentdonut[0.15cm]{25}{black} &
2560$\times$2048 & $\sim$2.5\,mm$^{\dagger}$ &
$>$500{,}000 & \checkmark &
\checkmark & \checkmark & \checkmark \\
\textbf{FOD-S2R \cite{vashist2025fods2r}} &
Public & 14 &
3{,}114 & \replaced{5{,}530}{N/A} &
\percentdonut[0.15cm]{25}{black} & \percentdonut[0.15cm]{50}{black} & \percentdonut[0.15cm]{25}{black} &
\percentdonut[0.15cm]{25}{black} & \percentdonut[0.15cm]{25}{black} &
1920$\times$1080 & --- &
3{,}137 & $\times$ &
\checkmark & $\times$ & $\times$ \\
\textbf{RT-Less \cite{he2023rtless}} &
Public & 38 &
38{,}392 & \replaced{72{,}088}{N/A} &
\percentdonut[0.15cm]{25}{black} & \percentdonut[0.15cm]{50}{black} & \percentdonut[0.15cm]{100}{black} &
\percentdonut[0.15cm]{25}{black} & \percentdonut[0.15cm]{25}{black} &
2448$\times$2048 & --- &
250{,}800 & $\times$ &
\checkmark & $\times$ & \checkmark \\
\textbf{XYZ-IBD \cite{huang2025xyzibd}} &
Public & 15 &
22{,}000 & 273{,}000 &
\percentdonut[0.15cm]{25}{black} & \percentdonut[0.15cm]{75}{black} & \percentdonut[0.15cm]{75}{black} &
\percentdonut[0.15cm]{75}{black} & \percentdonut[0.15cm]{100}{black} &
1920$\times$1080 & $\sim$2.5\,mm$^{\dagger}$ &
50{,}000 & \checkmark &
\checkmark & \checkmark & \checkmark \\
\textbf{ITODD \cite{drost2017itodd}} &
Public & 28 &
822 & 3{,}876 &
\percentdonut[0.15cm]{25}{black} & \percentdonut[0.15cm]{75}{black} & \percentdonut[0.15cm]{75}{black} &
\percentdonut[0.15cm]{75}{black} & \percentdonut[0.15cm]{50}{black} &
4064$\times$3044 & $\sim$0.1mm &
50{,}000 & \checkmark$^{*}$ &
$\times$ & \checkmark & \checkmark \\
\textbf{IPD \cite{kalra2024ipd}} &
Public & 20 &
30{,}000 & 100{,}000 &
\percentdonut[0.15cm]{100}{black} & \percentdonut[0.15cm]{75}{black} & \percentdonut[0.15cm]{75}{black} &
\percentdonut[0.15cm]{75}{black} & \percentdonut[0.15cm]{50}{black} &
6048$\times$4024 & $\sim$0.15mm &
150{,}000 & \checkmark &
\checkmark & \checkmark & \checkmark \\
\textbf{\acrshort{iasset} (ours)} &
Public & 32 &
508 & 19{,}672 &
\percentdonut[0.15cm]{100}{black} & \percentdonut[0.15cm]{100}{black} & \percentdonut[0.15cm]{100}{black} &
\percentdonut[0.15cm]{100}{black} & \percentdonut[0.15cm]{100}{black} &
1224$\times$1024 & $\sim$0.1\,mm$^{\ddagger}$ &
$\leq$8{,}000 & \checkmark &
\checkmark & \checkmark & \checkmark \\
\bottomrule
\noalign{\smallskip}
\multicolumn{17}{@{}l}{\scriptsize $^{*}$PBR materials added retroactively for the BOP Challenge 2020 \cite{hodan2020bop}; original release has none.} \\
\multicolumn{17}{@{}l}{\scriptsize $^{**}$Primesense Carmine 1.09 (structured light): 1.2\,mm depth resolution at 0.5\,m (2$\sigma$) \cite{primesense2013carmine}; Kinect v2 (ToF): GSD $\approx$1.4\,mm at 0.5\,m \cite{toth2022kinect}.} \\
\multicolumn{17}{@{}l}{\scriptsize $^{\dagger}$Intel RealSense D415 (active stereo): $\approx$2.5\,mm agreement with ground truth at 150--500\,mm range \cite{carfagni2019realsense}.} \\
\multicolumn{17}{@{}l}{\scriptsize $^{\ddagger}$Zivid 2+ MR60 (structured light): global planarity trueness $<$0.10\,mm at focus distance (600\,mm) \cite{zivid2024mr60}.} \\
\end{tabularx}
\end{table*}

Furthermore, we report image quality in terms of RGB resolution and depth accuracy. For the synthetic train set, we report dataset scale, availability of \acrfull{PBR} materials, and supported modalities. Notably, none of the existing datasets considered provide reconstructed meshes alongside CAD models, which precludes structured evaluation of \acrshort{real-to-sim} pipelines based on 2D-to-3D reconstruction.

Among public datasets, T-LESS~\cite{hodan2017tless} is notable for its high inter-class ambiguity across 30 texture-less objects, though it offers limited lighting and material variation; PBR materials were added retroactively for the BOP Challenge~\cite{hodan2020bop} rather than being part of the original design. DIMO~\cite{deroovere2024dimo} provides a large synthetic train set with strong pose variability and PBR support, but its real images are captured under controlled, low-clutter conditions. XYZ-IBD~\cite{huang2025xyzibd} offers rich background clutter and RGB-D modalities with a large instance count, yet does not include reconstructed meshes for \acrshort{real-to-sim} evaluation. RT-Less~\cite{he2023rtless} covers a broad set of texture-less objects but similarly lacks depth and reconstruction modalities. FOD-S2R~\cite{vashist2025fods2r} is one of the few datasets explicitly framed around sim-to-real transfer, though its visual complexity and modality coverage remain limited. Datasets used in our own benchmarking experiments, the Robotics dataset~\cite{horvath2022object} and the Automotive benchmark~\cite{martinez2024scap}, are either proprietary or low in visual complexity, limiting their use as general-purpose sim-to-real benchmarks. \added{Furthermore, large-scale industrial datasets, such as SORDI.ai \cite{abouakar2024sordi}, provide over a million synthetic renders comprising more than one hundred object classes, however, the corresponding real-world test set is not publicly released alongside the synthetic corpus, which limits its applicability to sim-to-real research.}

\added{To broaden the comparison beyond datasets that are uniformly outperformed by \acrshort{iasset}, Table~\ref{tab:dataset_comparison} additionally includes two public benchmarks that each stress a specific axis of difficulty. ITODD~\cite{drost2017itodd} targets industrial bin-picking with 28 texture-less, visually near-identical metal parts captured across over 800 scenes, making it a particularly demanding testbed for inter-class ambiguity; however, it provides only monochrome imagery and is released under a non-commercial license, limiting its use as a direct RGB-based training or benchmarking source. The Industrial Plenoptic Dataset (IPD)~\cite{kalra2024ipd} covers 20 industrial parts across 2{,}300 physical scenes and more than 100{,}000 object views, captured with 13 multi-modal cameras under four exposures and three lighting conditions spanning 100 to 100{,}000\,lux, constituting a substantially wider illumination range than any other dataset considered here, and one that directly challenges the light-variability advantage claimed for \acrshort{iasset}. Including these two datasets provides a more balanced picture of the state-of-the-art: while \acrshort{iasset} remains the only dataset offering reconstructed-mesh annotations alongside CAD models and combining high scores across all five visual-complexity categories simultaneously, ITODD and IPD each match or exceed it along a single, well-documented dimension (inter-class ambiguity and lighting variability, respectively), underscoring that no existing public dataset yet unifies all of these challenging properties at once.}

A consistent gap across all reviewed datasets is the absence of reconstructed meshes as a complement to CAD models, which precludes structured evaluation of \acrshort{real-to-sim} pipelines that rely on 2D-to-3D reconstruction rather than manual asset preparation. Furthermore, no existing public dataset simultaneously achieves high variability across all five visual complexity dimensions while providing RGB-D imagery at industrial sensor quality. \acrshort{iasset} addresses both limitations: it provides CADs and 2D-to-3D reconstructed meshes for all 32 object classes, supports evaluation across the full bidirectional \acrshort{sim-to-real} pipeline, and is captured with a high-accuracy structured-light sensor achieving depth trueness below 0.1\,mm~\cite{zivid2024mr60}, i.e. an order of magnitude more accurate than active stereo alternatives~\cite{carfagni2019realsense}.

\subsection{Automated 2D-to-3D Domain Adaptation}
\label{sec:techniques_for_DA}

The first stage of synthetic data generation with real-world relevant features requires 3D assets whose geometry and appearance approximate those of physical objects. Previous approaches typically relied on CAD models with manual~\cite{martinez2024scap, Mayershofer2021}, randomized~\cite{horvath2022object, Zhu2025icra}, or mixed~\cite{Eversberg2021} texture application. However, CAD models are not always available, and manual modeling of synthetic assets with precise geometry and materials is both time-intensive and demands specialized expertise.

Recent advances in neural 3D reconstruction offer alternatives without manual geometry or material assignment. \acrfull{3DGS}~\cite{kerbl2023gaussian} represents scenes as collections of anisotropic Gaussians optimized from multi-view images, enabling high-fidelity geometry and appearance reconstruction without manual intervention. \added{Each Gaussian is parameterized by a position, an anisotropic covariance, an opacity, and a view-dependent color, and renders them with a fast differentiable rasterizer; unlike implicit neural radiance fields, this explicit formulation achieves real-time, high-fidelity novel-view synthesis and, after adaptive densification and pruning, yields a compact geometry that can be converted into textured meshes for downstream rendering.} Subsequent work has extended Gaussian splatting to produce explicit mesh representations suitable for rendering pipelines~\cite{tobiasz2025meshsplats}. Generative approaches such as TRELLIS~\cite{TRELLIS} push automation further by inferring 3D structure and texture directly from one or a small number of input images, bypassing the need for multi-view capture entirely. For texture-only adaptation, vision-language-guided material generation tools such as MeshyAI~\cite{meshyai2025} can produce \acrfull{PBR} materials from a single RGB image and wrap them onto existing geometry, offering a practical middle ground when accurate CAD models are available but realistic textures are not.

The impact of scene context on sim-to-real transfer has received comparatively less attention than object appearance. Background randomization is a common baseline~\cite{tobin2017domain}, but photorealistic background reconstruction via 3DGS offers an alternative that anchors synthetic scenes in the \deleted{real} test environment, potentially reducing the domain gap contributed by scene context rather than object appearance alone~\cite{kerbl2023gaussian}.

\added{Model-level cross-domain adaptation follows a different and complementary strategy from the asset-reconstruction and domain-randomization approaches studied in this work. Rather than modifying synthetic assets or the source-data distribution, model-level methods modify the detector or its training procedure by exploiting labeled source-domain data together with unlabeled or sparsely labeled target-domain images. Representative approaches include adversarial feature alignment at the image and instance levels}~\cite{chen2018domainadaptive,saito2019strongweak}\added{, target-domain pseudo-labeling and self-training}~\cite{khodabandeh2019robust}\added{, and teacher--student adaptation, in which a teacher model generates target-domain supervision for a student model}~\cite{deng2021unbiased,cao2023contrastive}\added{.}

\added{These methods operate at the model-training level. In contrast, SynthRender operates at the data-generation level and produces fully annotated synthetic source data. They are therefore complementary rather than directly interchangeable. A controlled empirical comparison would require a separate target-data access protocol and method-specific implementation and tuning across the evaluated detector architectures, which is outside the experimental scope of this work. Combining domain-randomized synthetic pretraining with model-level adaptation remains a relevant direction for future work.}

\subsection{2D Generative Synthetic Data Generation}
\label{sec:generative_sdg}

\added{Image-generative methods constitute a complementary, purely 2D paradigm that bypasses the 2D-to-3D reconstruction step entirely, directly synthesizing appearance-diversified images without producing an intermediate 3D representation. GAN-based approaches such as SimGAN refine simulator outputs toward the appearance of unlabeled real images while employing self-regularization to preserve the information contained in the original synthetic image~\cite{shrivastava2017simgan}.}

\added{More recent diffusion-based approaches can generate or transform images using text, layout, depth, edge, or mask conditioning, providing greater control over scene composition than unconstrained image generation}~\cite{chen2024geodiffusion}\added{. When applied to diversify object or scene appearance rather than to perform a single targeted image translation, this generative image synthesis is sometimes referred to as generative domain randomization, in contrast to the parametric \acrfull{DR}~\cite{tobin2017domain} approach, which instead perturbs a fixed set of rendering parameters, such as textures, lighting, and object poses, sampled from predefined, hand-crafted distributions. Rather than sampling from such a distribution, generative domain randomization uses a generative model to synthesize novel appearance variations directly, extending diversity beyond what can be manually parameterized, but without ever constructing the 3D representation that the 2D-to-3D methods rely on. This distinction is central to the present work: generative domain randomization should not be confused with the \acrfull{GDR} adopted here, which is a physically guided form of parametric \acrshort{DR}. Both \acrshort{DR} and \acrshort{GDR} still render from an explicit 3D scene and therefore yield exact, geometry-consistent annotations by construction; they differ only in how the rendering parameters are sampled, with \acrshort{DR} drawing from uniform hand-crafted ranges and \acrshort{GDR} replacing them with physically plausible sampling of lighting, materials, and object placement. Generative domain randomization, by contrast, dispenses with the 3D scene altogether and operates purely in the 2D image space.} 

\added{This flexibility, however, comes with two notable drawbacks. First, because generation is not inherently constrained to preserve exact object shape, it can introduce unintended deformation of object geometry, boundaries, or scale unless explicitly conditioned on structural priors}~\cite{fang2024controlaug}\added{, a failure mode that the explicit 3D representations used in 2D-to-3D adaptation, and in \acrshort{DR} and \acrshort{GDR}, avoid by construction. Second, the iterative denoising process underlying diffusion-based generation incurs substantially higher computational cost than the lightweight, real-time parameter sampling used in parametric domain randomization}~\cite{cao2024survey, araya-martinez2025genai}\added{.}

\added{Compared with these image-generative approaches, 3D rendering provides exact annotations directly from the scene state, including object identities, poses, segmentation masks, and bounding boxes. It also preserves geometric and multi-view consistency because every image is generated from an explicit 3D scene. Generative models can provide broader appearance and contextual diversity, but may alter object boundaries, geometry, scale, or class-specific details; consequently, annotations propagated from conditioning inputs may require additional validation. Conversely, the diversity of rendering-based pipelines is constrained by the available 3D assets and the configured randomization ranges, although their physical and camera parameters remain explicitly controllable.}


\subsection{Synthetic Data Generation \replaced{Systems}{Pipelines}}
\label{sub:s2r_pipelines}

Synthetic data generation pipelines have emerged as a practical alternative to large-scale real annotation, particularly in domains where data acquisition is costly or hazardous. The core idea is to render training images from 3D assets under randomized conditions, a strategy known as \acrfull{DR}, so that the resulting distribution covers the variability encountered at deployment~\cite{tobin2017domain, tremblay2018training}. More recently, physically-grounded variants using \acrshort{GDR} have replaced uniform randomization with physically plausible sampling of lighting, materials, and object placement, yielding more realistic synthetic distributions and stronger sim-to-real transfer~\cite{Denninger2023}.

\added{In this section, the term ``system'' is used as an umbrella term for heterogeneous software categories. More specifically, Isaac Sim is an interactive robotics simulation application, CARLA is an autonomous-driving simulator, SynthDet is a dataset-generation example built on the Unity Perception package, and BlenderProc, Kubric, BlendTorch, SynMfg, and SynthRender provide frameworks or pipelines for synthetic dataset generation.}

\autoref{tab:sdg_comparison_integrated} compares representative \replaced{systems}{pipelines} along five axes: the underlying rendering engine, supported features, automatic annotation modalities, application focus, and the sim-to-real integration mechanisms they expose. Annotation richness is particularly relevant for multi-task learning and for downstream tasks such as pose estimation and depth completion, which may benefit from the same synthetic pass that produces detection labels.

\begin{table*}[bht]
\centering
\caption{\label{tab:sdg_comparison_integrated}Technical comparison of the proposed SynthRender against related \replaced{synthetic data generation systems and simulation applications}{SDG pipelines} with sim-to-real integration. The modalities category comprises automatic annotations and scene information that can be generated with each \replaced{system}{pipeline}. They are abbreviated as color (C), depth (D), pose annotations (P), semantic masks (SM), bounding boxes (BB), surface normals (N), and LiDAR data (L).}
\footnotesize
\renewcommand{\arraystretch}{1.5}
\setlength{\tabcolsep}{0.9pt}
\newcommand{\ann}[1]{\makebox[1.2em][c]{#1}}
\newcommand{\no}{\ann{--}}
\begin{tabular}{l l l l l r}
\toprule
\textbf{\replaced{System}{Pipeline}} & \textbf{Engine} & \textbf{Features} & \textbf{Auto. Annotations} & \textbf{Focus} & \textbf{Sim-to-Real Functions} \\
\midrule
SynthDet \cite{borghes2020synthdet}    & Unity \cite{unity}                & Domain Randomization  & \ann{C}\ann{D}\no\ann{SM}\ann{BB}\no\no       & Consumer      & Unity Perception, ML-Agents       \\
CARLA \cite{dosovitskiy2017carla}      & Unreal \cite{unreal}              & Urban Sim, Weather    & \ann{C}\ann{D}\ann{P}\ann{SM}\ann{BB}\no\ann{L} & Driving     & Python API, ROS bridge            \\
Isaac Sim \cite{isaacsim}              & Omniverse                         & DR/GDR, Physics       & \ann{C}\ann{D}\ann{P}\ann{SM}\ann{BB}\ann{N}\ann{L} & Industrial & Omniverse Replicator, ROS      \\
CAD2Render \cite{moonen2023cad2render} & Unity (HDRP)                      & DR, GPU-accelerated   & \ann{C}\ann{D}\no\ann{SM}\no\ann{N}\no        & Manufacturing & GUI Generator, custom pass        \\
SynosIs \cite{fulir2024synosis}        & Appleseed \cite{appleseed_zenodo} & Procedural Textures   & \ann{C}\no\no\ann{SM}\no\no\no                & Inspection    & Parametric control                \\
\added{Kubric} \cite{greff2022kubric}  & \added{Blender + PyBullet}         & \added{Physics, scalable video} & \ann{C}\ann{D}\ann{P}\ann{SM}\ann{BB}\ann{N}\no & \added{General vision} & \added{Python API, distributed generation} \\
\added{BlendTorch} \cite{heindl2020blendtorch} & \added{Blender (Eevee)}      & \added{Adaptive DR, online} & \ann{C}\no\no\no\ann{BB}\no\no & \added{Industrial} & \added{PyTorch streaming, training feedback} \\
SynMfg \cite{Zhu2025icra}              & BlenderProc \cite{Denninger2023}   & DR, PBR, scriptable   & \ann{C}\no\no\ann{SM}\ann{BB}\no\no           & Manufacturing & Python API, scene control scripts \\
SynthRender (ours)                     & BlenderProc \cite{Denninger2023}   & DR/GDR, Programmatic  & \ann{C}\ann{D}\ann{P}\ann{SM}\ann{BB}\ann{N}\no & Industrial  & Python API, randomization scripts \\
\bottomrule
\end{tabular}
\end{table*}

\added{Among Blender-based approaches, Kubric}~\cite{greff2022kubric}\added{ combines Blender rendering with PyBullet physics and emphasizes scalable image and video generation with rich ground-truth information. Its distributed architecture supports large generation workloads and a broad range of vision tasks, although it is not specifically tailored to industrial sim-to-real object perception. BlendTorch}~\cite{heindl2020blendtorch}\added{ follows a different design by coupling Blender's Eevee renderer directly with PyTorch. It generates domain-randomized data online and supports bidirectional communication through which the training process can influence the simulation. This avoids the storage requirements of offline datasets and enables adaptive randomization, whereas BlenderProc-based approaches prioritize physically-based rendering and reusable offline datasets with richer annotation modalities.}

\replaced{Game-engine-based systems}{Game-engine-based pipelines} such as SynthDet~\cite{borghes2020synthdet} and CARLA~\cite{dosovitskiy2017carla} offer real-time rendering and broad ecosystem support, but are oriented toward consumer object recognition and autonomous driving respectively, and provide limited configurability for industrial part geometry. CARLA stands out for its sensor realism, including LiDAR and weather simulation, but its scene model is inherently road-centric. \added{As an interactive robotics simulation application rather than a dataset-generation pipeline, }Isaac Sim~\cite{isaacsim} offers the most complete annotation modalities, including surface normals and LiDAR, and provides native ROS integration for robotics deployment; however, its dependency on the Omniverse ecosystem introduces significant infrastructure overhead. CAD2Render~\cite{moonen2023cad2render} \replaced{is a modular toolkit targeting}{targets} manufacturing and provides GPU-accelerated rendering via Unity HDRP, but lacks pose annotations and bounding box output natively. SynosIs~\cite{fulir2024synosis} is a synthetic image generation pipeline focused on inspection tasks and procedural texture generation, but its annotation modalities are limited to color and semantic masks. SynMfg~\cite{Zhu2025icra}, the most closely related prior work, is a BlenderProc-based synthetic data generation framework targeting manufacturing with \acrshort{PBR} materials, but does not output depth, pose, or surface normal annotations in its current implementation.

SynthRender shares the BlenderProc~\cite{Denninger2023} foundation with SynMfg\added{ and should be understood as an application-specific framework built on top of BlenderProc rather than as a new rendering or simulation engine}. \added{It inherits BlenderProc's Blender/Cycles rendering, physics integration, procedural scene manipulation, and annotation capabilities.} Its \replaced{implementation-specific additions comprise}{contribution lies in the integration of} guided \acrshort{DR}, automated scene generation, keyframe-based scene management, batch rendering, parallel execution, pre-render scene visualization, and single-pass multi-output annotation into a reproducible framework tailored to industrial object perception. \added{Compared with SynMfg, SynthRender additionally targets guided randomization, keyframe-based scene construction, pre-render inspection, and the simultaneous output modalities required by the experiments in this work. Isaac Sim, by contrast, is a broader interactive robotics simulation application with native ROS integration and sensor modalities such as LiDAR, capabilities that are outside SynthRender's offline dataset-generation scope.} As summarized in~\autoref{tab:synthrender_positioning}, this makes it directly applicable to the multi-task requirements of industrial perception. Beyond these technical and usability optimizations, the contribution of this work is not SynthRender as a standalone tool, but the systematic ablation of its design choices and their quantified impact on sim-to-real transfer, as detailed in~\autoref{sec:results}. \added{Accordingly, we do not claim the engineering integration of these existing components as a methodological novelty.}

As reviewed in this section, research has demonstrated the potential of \acrshort{DR}~\cite{Zhu2025icra, tremblay2018training} and manual \acrshort{DA}~\cite{martinez2024scap} towards narrowing the \acrshort{sim-to-real} gap. However, some challenges remain insufficiently addressed:

\begin{itemize}
    \item Bidirectional \acrshort{sim-to-real} workflows that combine automated \acrshort{DA} from physical parts with \acrshort{DR} are needed to enable perception in tasks where \acrfull{CAD} files have non-realistic appearance or are unavailable.
    \item An extensible benchmark capturing semi-uncontrolled industrial conditions, i.e. varying illumination, object perspectives, clutter, and changing backgrounds, is needed to evaluate and compare sim-to-real approaches, including analysis of inter-class failure modes.
\end{itemize}

\deleted{As described in the next section, this work addresses these identified gaps through: (i)~a systematic investigation of 2D-to-3D reconstruction techniques for \acrshort{DA}; (ii)~structured ablation studies quantifying the contribution of individual \acrshort{DR} design choices via SynthRender; and (iii)~\acrshort{iasset}, a 32-class dataset providing both CAD models and reconstructed meshes, synthetic train sets, and real\added{-world} test images captured under semi-uncontrolled industrial conditions.}

\section{Methodology}
\label{sec:method}

\begin{figure}[thpb]
\centerline{\includegraphics[trim={0.7cm 1.2cm 0.7cm 1.2cm},clip, width=1\linewidth]{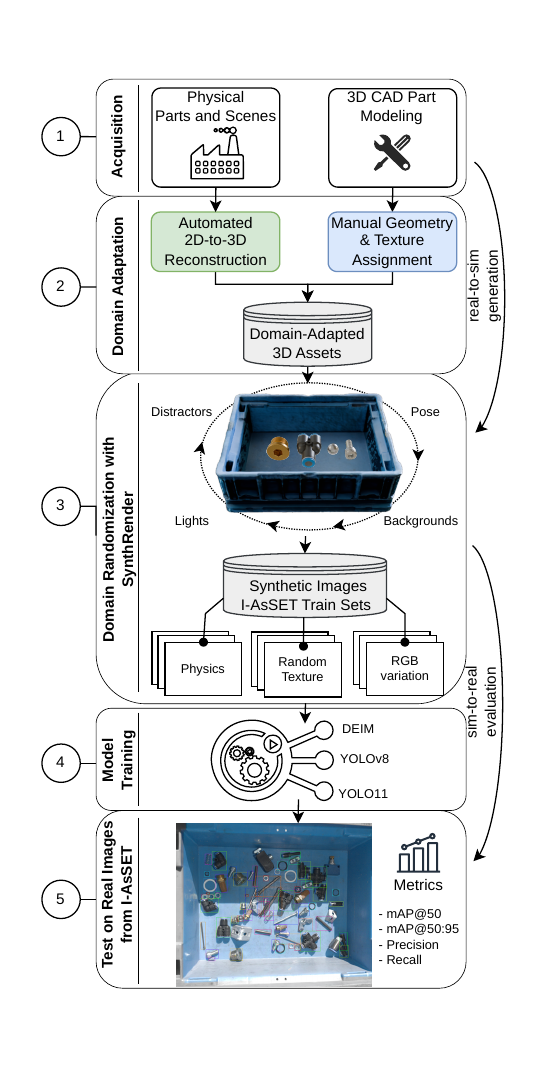}}
\caption{Proposed framework for benchmarking and ablation studies, comprising five stages: data acquisition, \acrshort{DA} via 2D-to-3D physical scene reconstruction and manual assignment, different \acrshort{DR} settings generate separate synthetic train-set variants using SynthRender, training across multiple object detectors, and evaluation on the introduced \acrshort{iasset} dataset for bidirectional \acrshort{sim-to-real} gap benchmarking.}
\label{fig:intro}
\end{figure}

\deleted{Based on the introductory ~Figure\ref{fig:intro_graph},} ~\autoref{fig:intro} delves into the five main stages followed in this work. These methodological steps are elaborated in the upcoming sections, organized as follows:

\begin{itemize}
    \item \textbf{Acquisition:} First, we collect \acrshort{CAD} files and physical parts corresponding to the objects under test, distractors or scene elements such as small load carriers, also known as \acrfull{KLT}, as per their German designation. We retrieve \acrshort{CAD} models either from public catalogs with explicit written permission from the intellectual property holders or we model them from procured physical functional shapes. \replaced{The real-world test set of \acrshort{iasset} (the physical images on which every model is evaluated, while all training and validation use synthetic images) is acquired and annotated in this stage.}{The real test set of \acrshort{iasset} is acquired and annotated in this stage, as described in~\mbox{\autoref{sub:iasset}}.}
    
    \item \textbf{\acrlong{DA}:} In this stage \acrshort{DA} follows two different paths: i) traditional manual assignment of materials and shape corrections to improve the appearance, geometry and details of initial \acrshort{CAD} models, and ii) our proposed usage of modern 3D reconstruction methods to generate 3D assets using \acrfull{3DGS} \cite{tobiasz2025meshsplats} and GenAI methods for 3D reconstruction from 2D images\deleted{, as explained in \mbox{\autoref{method:sub:techniques_for_DA}}}. As a result, the domain-adapted 3D assets of \acrshort{iasset} are generated in this stage.
    
    \item \textbf{\acrshort{DR} with SynthRender:} in this stage, we develop the open-source \acrshort{SDG} engine SynthRender. \replaced{I}{As explained in \mbox{\autoref{sub:synthrender}}, i}t includes physics simulation~\cite{bullet} for realistic object placement and three-point lighting with controllable intensity. Additionally, we explore multiple design decisions and parameters variations that led to our best results\deleted{, as described in \mbox{\autoref{sub:sota_benchmark}}}. These synthetic train sets are included in \acrshort{iasset} for reproducibility.

    \item \textbf{Model Training:} Experiments \deleted{in~\mbox{\autoref{sub:synth_render_across_models}}} evaluate the \acrshort{sim-to-real} generalization of SynthRender across three detector families: YOLOv8, YOLO11, and DEIM, as well as multiple model sizes, providing a basis for isolating the effect of synthetic data design choices from architectural factors.

    \item \textbf{Testing on Real Data:} Leveraging the real and synthetic modalities of \acrshort{iasset}, the real\added{-world} test set is used for evaluation, following standard object detection metrics (mAP@50~\cite{everingham2010pascal} and mAP@50:95~\cite{lin2014microsoft}) to quantify the bidirectional \acrshort{sim-to-real} gap across all experimental conditions.
    
\end{itemize}

\subsection{2D-to-3D Domain Adaptation Methods}
\label{method:sub:techniques_for_DA}

\begin{figure}[htbp]
\centerline{\includegraphics[trim={0cm 13.2cm 0.2cm 0cm},clip, width=1\linewidth]{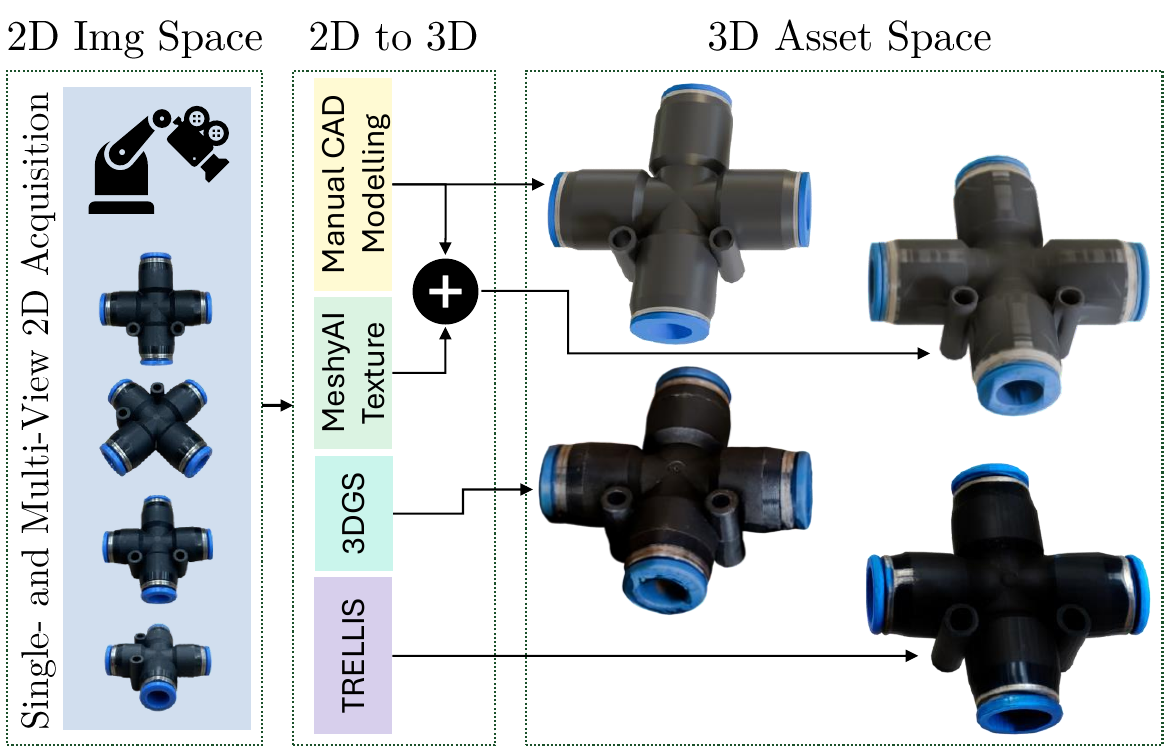}}
\caption{\label{fig:automated_DA}\replaced{3D asset and texture generation as \acrshort{DA} approaches. The four pipelines cover three geometry paradigms: manual modeling, \acrshort{3DGS}, and TRELLIS; and four texture sources: manual \acrshort{PBR}, MeshyAI, 3DGS, or TRELLIS.}{3D asset and texture
generation as DA approaches.}}
\end{figure}

As shown in the DA stage of~\autoref{fig:intro}, we explore low-overhead techniques for accurate 3D asset generation from physical objects. \replaced{\mbox{Figure~\ref{fig:automated_DA}} compares four asset-generation pipelines that span three geometry paradigms (manual modeling, \acrshort{3DGS}, and GenAI (TRELLIS) reconstruction) and additionally differ in how object textures are produced, ordered by increasing automation and decreasing human effort.}{\mbox{Figure~\ref{fig:automated_DA}} compares four approaches for 2D-to-3D transformation with increasing levels of automation and decreasing human effort.}

Empirically evaluating the pipelines illustrated in~Figure~\ref{fig:automated_DA} on all \acrshort{iasset} classes allows us to compare bidirectional \acrshort{sim-to-real} transfer towards domain-adapted 3D synthetic data generation from 2D captures. \replaced{In the following, we describe the four pipelines, grouped by their geometry-reconstruction paradigm: manual modeling (the first two, which share \acrshort{CAD} geometry and differ only in texturing), \acrshort{3DGS}, and TRELLIS.}{In the following, we describe the four pursued approaches:}

\paragraph{Manual Modeling\added{.}}
High-quality CAD representations are modeled from physical functional shapes or retrieved, with written permission, from their intellectual property holders. Hand-crafted \acrfull{PBR} materials~\cite{pharr2016pbr} are then applied in Blender to visually represent the appearance of physical parts. This method provides ideal digital twins, but is time-consuming, requires expert knowledge, and misses part-production artifacts and imperfections.

\paragraph{Manual CAD + MeshyAI.}
Accurate CAD geometry is retained, and PBR materials are generated automatically from a single real RGB image using MeshyAI~\cite{meshyai2025}. The resulting textures are wrapped onto the CAD objects. Since only one image is required for texture generation, the method can be easily automated; however, surface regions not visible in the input view (such as the back face) are inferred by the model rather than observed, which may introduce texture inconsistencies on occluded surfaces. Despite this limitation, the method offers a good compromise between ideal geometry and realistic appearance for the visible object regions.

\paragraph{\acrshort{3DGS}.}
Multi-view images are collected from a physical part. A 3D Gaussian Splatting pipeline~\cite{tobiasz2025meshsplats} implemented in the KIRI Engine~\cite{kiri2025} then generates 3D mesh representations encoding both geometry and texture. This method avoids manual texturing and produces meshes with a realistic appearance. However, geometric artifacts introduced during reconstruction require post-processing, including data cleaning and noise removal.

\paragraph{TRELLIS.}
Both, mesh and texture can be generated with the TRELLIS model~\cite{TRELLIS} directly from one or multiple input images. Thus, when CAD models are unavailable, this GenAI method constitutes a convenient, automated approach to produce semantically correct 3D assets.\\

In addition to object modeling, we evaluate the impact of background realism using 3DGS-reconstructed scenes from the \deleted{real} test environment. This approach provides contextualized background information as an alternative to manual or randomized scene generation, and allows us to isolate how much of the domain gap originates from object appearance versus scene context. \deleted{ A full benchmark of the bidirectional \acrshort{sim-to-real} capabilities of these methods is presented in~\mbox{\autoref{sub:DA_results}}.}

\subsection{SynthRender Framework}
\label{sub:synthrender}

\begin{figure}[thpb]
    \centering
    \includegraphics[width=\linewidth]{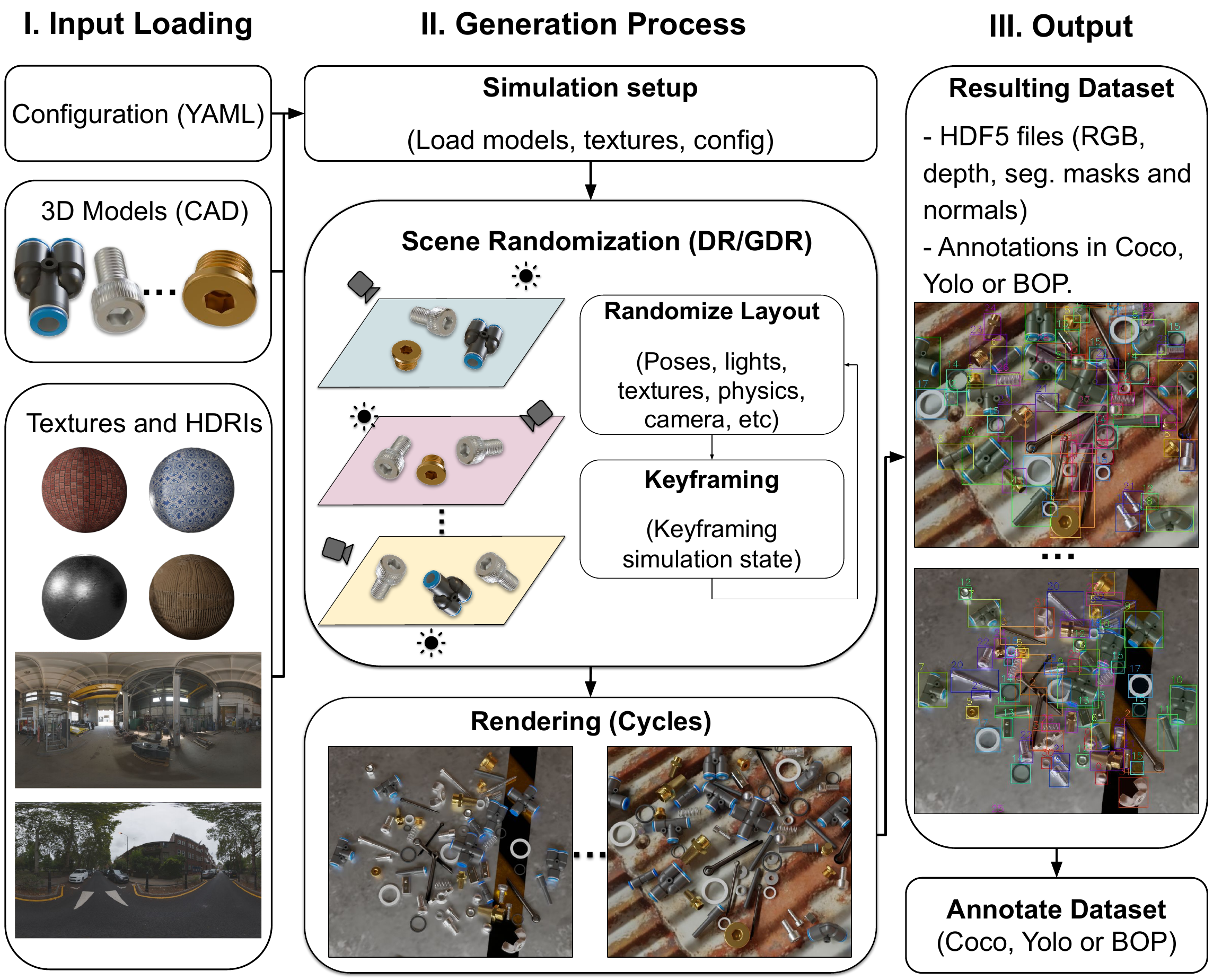}
    \caption{Functional diagram of SynthRender, illustrating the input, generation, and output stages.}
    \label{fig:synthrender-diagram}
\end{figure}

The SynthRender framework is a programmatic generator of synthetic data with a \acrshort{sim-to-real} transfer focus. As summarized in~\autoref{tab:synthrender_positioning}, its contribution lies in the integration of \acrshort{GDR}, automated scene generation, keyframe-based scene management, batch rendering, parallel execution, pre-render scene visualization for debugging \acrshort{GDR} configurations, and single-pass multi-output annotation into a reproducible framework tailored to industrial object perception. As illustrated in the first stage of~\autoref{fig:synthrender-diagram}, it takes three main inputs: i) a configuration file defining the parameters listed in~Table~\ref{tab:synthrender_params}, ii) 3D meshes or CAD models of the target objects for which annotations will be generated, and iii) contextual scene information, including textures, \acrfull{HDRI} files~\cite{debevec1997recovering}, and distractor objects. Each loaded model may appear multiple times per scene at randomized poses to increase data diversity and scene complexity.

\begin{table}[htpb]
\caption{\label{tab:synthrender_positioning}Positioning of SynthRender with respect to related synthetic data generation \replaced{systems}{pipelines}. BProc: BlenderProc; \replaced{Isaac: NVIDIA Isaac Sim, built on the Omniverse platform}{Omniv.: NVIDIA Omniverse/Isaac Sim}; SynthR.: SynthRender (ours). \added{The comparison describes the documented functionality of each system rather than ranking their overall capabilities; ``No'' does not imply that a feature could not be implemented through custom extensions.}}
\centering
\scriptsize
\renewcommand{\arraystretch}{1.25}
\setlength{\tabcolsep}{1.0pt}
\begin{tabular}{lccccc}
\toprule
\textbf{Feature} & \textbf{SCAP}\cite{ours} & \textbf{SynMfg}\cite{Zhu2025icra} & \textbf{BProc}\cite{Denninger2023} & \textbf{Isaac}\cite{isaacsim}& \textbf{SynthR.}\cite{araya2026synthrender} \\
\midrule
Open source & No & Yes & Yes & No & Yes \\
Guided DR & Yes & No & No & Yes & Yes \\
\makecell[l]{Workflow\\automation} & Partial & Partial & Yes & Yes & Yes \\
\makecell[l]{Configurable\\pipeline} & Partial & Yes & Yes & Yes & Yes \\
\makecell[l]{Keyframe-based\\generation} & No & No & Partial & Partial & Yes \\
\makecell[l]{Batch\\rendering} & No & No & Partial & Yes & Yes \\
\makecell[l]{Single-pass\\annotation} & No & No & Yes & Yes & Yes \\
\makecell[l]{Reproducible\\config + seed} & Partial & Partial & Yes & Yes & Yes \\
\makecell[l]{Parallel\\implementation} & No & Yes & No & Yes & Yes \\
\makecell[l]{Pre-render\\visualization /\\debug mode} & Partial & No & No & Yes & Yes \\
\added{Native ROS\\integration} & No & No & No & Yes & No \\
\added{LiDAR sensor\\output} & No & No & No & Yes & No \\
\added{Interactive\\simulation} & No & No & No & Yes & No \\
\bottomrule
\end{tabular}
\end{table}

As depicted in the second stage of~\autoref{fig:synthrender-diagram}, SynthRender applies DR or GDR according to user-defined rules and ranges. Each frame has a unique, temporally discontinuous configuration of layout and simulation parameters. Both simulation and rendering are executed in Blender through the BlenderProc API, leveraging the Cycles path tracing engine~\cite{cycles}.

\replaced{For each randomized scene, the rendered modalities and annotation-relevant scene metadata are generated during the same rendering stage and stored in per-frame HDF5 containers}{For each randomized scene, metadata is computed during rendering and later used to generate the corresponding annotations}, as shown in the third stage of~\autoref{fig:synthrender-diagram}. \replaced{Each container includes RGB images, depth maps, normal maps, segmentation masks, object identities, poses, and the parameter values associated with the frame.}{After rendering, RGB images and metadata are stored in HDF5 files. Each file contains RGB images, depth maps, normal maps, segmentation masks, and the parameter values used for each frame.} \added{A subsequent format-specific export step reads these containers and produces the requested dataset annotations.} The following subsections describe the main components of the pipeline.

\subsection{Load and process data}

The framework loads a configuration file that defines all internal simulation parameters, including paths to CAD models, materials, and \acrshort{DR} settings. Each CAD model undergoes preprocessing to ensure compatibility with the framework. Models are parented to a textureless cube-like proxy mesh used for faster collision-free placement. Optionally, all sub-parts of a model can be merged into a single mesh. Additional attributes, such as scale, texture, and category identifiers for annotation, are also assigned at this stage.

Synthetic distractors can also be added to the scene. These are altered versions of existing assets and are used as distractors. They may consist of simple geometric primitives or deformed variants of existing CAD models. Deformations modify the geometry while preserving the original texture, ensuring visual similarity without semantic equivalence.

Finally, auxiliary simulation elements are loaded, including a default digital twin scene, area lights arranged in a three-point studio configuration, and HDRI environment maps. If enabled, rigid-body physics are assigned to all models. \replaced{Target models and target-derived synthetic distractors are modeled as dynamic objects, whereas regular distractors are modeled as kinematic collision objects whose poses remain fixed during each physics simulation.}{Target models and synthetic distractors use active rigid bodies, while regular distractors use passive rigid bodies.}

\subsection{Set-up random scene}
\label{sub:synthrender:scene_setup}

A total of \replaced{$N_{\mathrm{scene}}$}{$n$} scenes are generated according to the configuration parameters shown in~Table~\ref{tab:synthrender_params}. \added{Here, $N_{\mathrm{scene}}$ denotes the total number of randomized scene states generated in a run. Scene generation is controlled through a YAML configuration file containing the asset paths, asset whitelists and blacklists, camera and lighting settings, object-specific configurations, sampling bounds, physics settings, output modalities, and random seed.}

\added{For every scene, parameters are generated in a fixed order: the environmental illumination and support-plane material are selected; the anchor, camera, and lights are configured; target and distractor models are sampled; collision-free candidate poses are generated; optional rigid-body simulation is performed; and the resulting state is stored at its corresponding keyframe. Python's \texttt{random} module and NumPy are initialized with the configurable seed, making the generated sequence deterministic when a fixed seed is provided. Setting the seed to $-1$ instead selects a new random seed. When only a frame interval is rendered, the random operations preceding the interval are still evaluated, although their states are not keyframed, so that a given frame remains identical to the same frame generated as part of the complete sequence.}

Randomization affects model selection and visibility, collision-free placement, camera pose and depth of field, lighting intensity and color, environmental backgrounds, and support-plane materials. To improve rendering efficiency, each randomized scene state is mapped to a unique keyframe on the animation timeline. Since HDRI environment maps cannot be keyframed in the same way as object transforms and light parameters, scenes sharing the same HDRI are grouped and rendered in batches. The following operations are performed for each generated scene:

\paragraph{\textbf{Environmental background:}} HDRI images are selected from the asset directory after applying the configured whitelist and blacklist. Since HDRIs cannot be keyframed, scenes are rendered in batches sharing the same HDRI, and the world background is changed between batches.

\paragraph{\textbf{World lighting:}} The HDRI light strength is sampled from the user-defined range in increments of $0.01$. HDRI-based lighting provides an ambient illumination component that complements the three area lights, producing softer shadow transitions and a more spatially distributed illumination of the scene.

\paragraph{\textbf{Plane sampling:}} Each loaded plane represents one candidate support material. One plane is selected uniformly for each scene and made visible, while the remaining planes are hidden. This reproduces material sampling without changing a material data block between keyframes.

\paragraph{\textbf{Anchor pose:}} \replaced{The anchor position is sampled using BlenderProc's spherical-shell sampler within the configured center, radius, and elevation bounds, while its rotation around the vertical axis is randomized.}{The anchor position and rotation are randomly sampled from a spherical volume defined in the configuration file.} The anchor acts as the local reference for the target spawn volume, camera, and three-point lighting arrangement.

\paragraph{\textbf{Camera:}} The camera position is sampled from a spherical shell centered on the anchor, using the configured camera-distance and elevation bounds. Its optical axis is oriented toward the anchor, and its f-stop is sampled uniformly from the configured interval. The image resolution and sensor width are fixed through the configuration file; when a camera-intrinsics YAML file is provided, its intrinsic matrix is loaded and used for rendering. \added{Although both the anchor and camera positions are randomized, the camera optical axis is always directed toward the anchor. This models viewpoint variation but not an independent camera-mounting misalignment or an offset of the projection center. Blender's horizontal and vertical lens-shift parameters are therefore not randomized in the current experiments. Future extensions could randomize these parameters or perturb the camera pointing direction to model calibration and mounting errors.}

\paragraph{\textbf{Area lights:}} A three-point lighting setup is used. Light positions and directions remain fixed relative to the anchor, while their intensity and, when enabled, color are varied. Let $E_l^{\mathrm{min}}$ and $E_l^{\mathrm{max}}$ denote the configured minimum and maximum total irradiance values.

By default, exponential lighting is enabled and the total irradiance follows a polynomial schedule over the scene sequence. Given the normalized frame coordinate $u_f = f/(N_{\mathrm{scene}}-1)$ with $0 \leq f < N_{\mathrm{scene}}$, the irradiance is
\[
E_l(f) = E_l^{\mathrm{max}} \, u_f^{\gamma_l},
\qquad
u_f = \frac{f}{N_{\mathrm{scene}}-1}.
\]
Here, $\gamma_l \geq 0$ is the exponential factor from Table~\ref{tab:synthrender_params}. For $\gamma_l > 1$, a larger proportion of the generated scenes receives low irradiance while the sequence still reaches $E_l^{\mathrm{max}}$. The edge case $\gamma_l = 1$ reduces the schedule to a linear increase of irradiance with the frame index. This non-uniform schedule compensates for the tendency of uniform randomization to overrepresent high-intensity configurations. Since industrial cameras operate with a linear sensor response limited by pixel well capacity and exposure tuning optimized for moderate illumination~\cite{emva1288,nakamura2006image}, increasing the proportion of moderate and low irradiance scenes reduces the number of saturated synthetic images.

Exponential lighting can optionally be disabled, in which case the total irradiance is instead sampled uniformly and independently of frame index,
\[
E_l \sim \mathcal{U}\left(E_l^{\mathrm{min}}, E_l^{\mathrm{max}}\right).
\]

The resulting total irradiance is randomly divided into three non-negative components, one per area light, whose sum equals $E_l(f)$ (or $E_l$, in the uniform case). Each component is converted to the corresponding Blender light energy using the fixed light-to-anchor distance.

If light-color randomization is enabled, a selector $h$ is sampled from $\{0,0.1,\ldots,1\}$. For $h<1$, the saturation $s$ is sampled from $\{0.5,0.8\}$; $h=1$ denotes the achromatic option and uses $s=0$. The value component is fixed at $v=1$. HSV is used as the sampling representation because it allows hue and saturation to be controlled independently while leaving illumination magnitude to the separately sampled light energy. Since BlenderProc's light-color interface expects an RGB triplet, the sampled HSV color is converted to RGB before
assignment. If color randomization is disabled, the fixed light color stored in the scene is retained.

\paragraph{\textbf{Target models:}} The number of real and synthetic target models is sampled from the corresponding integer bounds. Models are then selected uniformly without replacement from their respective asset pools. A value of $-1$ as an upper sampling bound denotes the complete available pool. Each position and Euler-angle component is sampled uniformly between its configured lower and upper bounds. Positions are expressed relative to either the dynamic anchor or a configured static origin.

A candidate target pose is accepted only if its reference point and at least six of the eight corners of its transformed 3D bounding box lie inside the camera frustum and its bounding box does not intersect previously accepted objects. Up to 1,000 candidate poses are evaluated per model. If no valid pose is found within this limit, that model is omitted from the current scene rather than accepting an invalid placement.

\paragraph{\textbf{Distractor models:}} Real and procedurally generated distractors are sampled independently from their corresponding asset pools. Their counts, scales, copies, position bounds, and orientation bounds are defined separately in the configuration file. Distractor positions and Euler rotations are sampled component-wise from uniform distributions within these bounds. Their vertical offset is calculated from the bottom of their bounding box so that the sampled position corresponds to the support surface. Distractor poses are checked against the targets, previously placed distractors, and fixed scene geometry, but they are not required to lie inside the camera frustum. \added{For each pool, the number of distractors is drawn uniformly from its configured integer interval and the corresponding models are selected uniformly without replacement; an upper bound of $-1$ selects the complete available pool.}

\paragraph{\textbf{Physics simulation:}} Candidate poses are validated before invoking Blender's rigid-body simulation. Collision detection first applies bounding-sphere and axis-aligned bounding-box tests as broad-phase filters, followed by a separating-axis-theorem test between oriented bounding boxes as the narrow-phase check. This preliminary validation is implemented outside Blender's physics solver to avoid repeatedly modifying the Blender scene while searching for valid initial poses. \added{BlenderProc already provides native collision-aware pose sampling~\cite{Denninger2023}; the custom validation stage is therefore not intended to replace a missing capability. It was retained because SynthRender jointly evaluates camera-frustum coverage and collisions against fixed scene geometry and previously accepted objects, while allowing configurable bounding-sphere, axis-aligned bounding-box, and separating-axis-theorem checks. Candidate poses are evaluated numerically from their transformed bounding volumes before modifying the Blender scene, assigning keyframes, or invoking rigid-body simulation. Consequently, invalid candidates can be rejected without executing a physics simulation, reducing unnecessary scene-update and simulation overhead.}

When physics is enabled, target models and target-derived synthetic distractors are configured as dynamic objects, whereas regular distractors and support geometry are configured as kinematic collision objects whose poses remain fixed during each simulation. \replaced{These terms replace Blender's implementation-specific ``active'' and ``passive'' rigid-body terminology.}{The dynamic and fixed collision-body configurations correspond to Blender's ``active'' and ``passive'' rigid-body types, respectively.} A manually modeled child mesh whose name begins with \texttt{\#Collider} is used when available. \added{A model can contain multiple such collider children constructed from simple convex primitives, such as boxes, cylinders, and spheres. Blender treats these children as a compound collider, allowing openings and concave regions to be represented without enclosing the complete object in a single hull.} Otherwise, a convex-hull proxy is generated from the model meshes. The optional convex-hull decomposition setting generates multiple convex components for objects that require a closer collision approximation.

The simulation is evaluated every 2 simulated seconds and is stopped when all dynamic objects satisfy the configured translational and rotational stability thresholds over the final simulated second, or when the maximum simulation duration of 10 seconds is reached. The reference configuration uses per-coordinate thresholds of $0.5$ Blender length units for translation and $2$ radians for Euler rotation. After termination, the final poses of the dynamic objects are fixed and stored at the scene's keyframe. If camera reorientation is enabled, the camera is redirected toward the centroid of the settled dynamic objects while preserving its sampled viewing distance.

\begin{table}[htpb]
\caption{SynthRender available parameters for DR and GDR. \added{The superscripts $\mathrm{min}$ and $\mathrm{max}$ denote the user-defined lower and upper bounds of each parameter, respectively.}}
\label{tab:synthrender_params}
\centering
\scriptsize
\renewcommand{\arraystretch}{1.15}
\setlength{\tabcolsep}{3pt}
\begin{tabular}{>{\raggedright\arraybackslash}m{2.35cm} >{\centering\arraybackslash}m{1.9cm} >{\centering\arraybackslash}m{0.9cm} >{\raggedright\arraybackslash}m{2.1cm}}
\toprule
\textbf{Parameter} & \textbf{Range} & \textbf{Unit} & \textbf{Description} \\
\midrule
\multicolumn{4}{l}{\textit{\textbf{General Settings}}} \\
Output format & $\{\mathrm{seg,rgb,norm}\}$ & -- & Output channels \\
Physics & $\{0,1\}$ & \replaced{--}{Bool} & Physics simulation \\
Background light &
\replaced{$s_{\mathrm{bg}} \ge 0$}{$s \ge 0$} &
-- & Env. light intensity \\

\midrule
\multicolumn{4}{l}{\textit{\textbf{Anchor Spawn}}} \\
Center &
\replaced{$\mathbf{c}_{a} \in \mathbb{R}^3$}{$c \in \mathbb{R}^3$} &
m & Anchor center \\
Radius &
\replaced{$r_{a} \ge 0$}{$r \ge 0$} &
m & Anchor radius \\
Elevation &
\replaced{$\epsilon_{a} \in [-90,90]$}{$\epsilon \in [-90,90]$} &
$^\circ$ & Elevation angle \\

\midrule
\multicolumn{4}{l}{\textit{\textbf{Cam \& Lighting}}} \\
Camera elevation &
\replaced{$\epsilon_{c} \in [-90,90]$}{$\epsilon \in [-90,90]$} &
$^\circ$ & Cam-to-anchor angle \\
Camera distance &
\replaced{$d_{c} \ge 0$}{$r \ge 0$} &
m & Distance to anchor \\
Light distance &
\replaced{$d_{l} \ge 0$}{$r \ge 0$} &
m & Distance to anchor \\
Light intensity &
\replaced{$E_{l} \ge 0$}{$E \ge 0$} &
W/m$^2$ & Irradiance \\
Light exponential &
\replaced{$\gamma_{l} \ge 0$}{$e \ge 0$} &
-- &
\replaced{Exponential factor}{Falloff factor} \\
Light color rand. & $\{0,1\}$ & \replaced{--}{Bool} & Random RGB color \\

\midrule
\multicolumn{4}{l}{\textit{\textbf{Object Spawn}}} \\
Target count &
\replaced{$n_{t}\in[n_{t}^{\mathrm{min}},n_{t}^{\mathrm{max}}]$}{$n\in[\min,\max]$} &
-- & No. of targets \\
Distractor count &
\replaced{$n_{d}\in[n_{d}^{\mathrm{min}},n_{d}^{\mathrm{max}}]$}{$n\in[\min,\max]$} &
-- & No. of distractors \\
Synth. distractor count &
\replaced{$n_{s}\in[n_{s}^{\mathrm{min}},n_{s}^{\mathrm{max}}]$}{$n\in[\min,\max]$} &
-- & No. of synth. distractors \\
Position &
\replaced{$\mathbf{p}_{o}\in[\mathbf{p}_{o}^{\mathrm{min}},\mathbf{p}_{o}^{\mathrm{max}}]$}{$p\in[\min,\max]^3$} &
m & Offset from anchor \\
Orientation &
\replaced{$\boldsymbol{\theta}_{o}\in[-180,180]^3$}{$\theta\in[-180,180]^3$} &
$^\circ$ & 3D Euler orientation \\

\midrule
\multicolumn{4}{l}{\textit{\textbf{Object Config}}} \\
Join children & $\{0,1\}$ & \replaced{--}{Bool} & Merge into one mesh \\
Scale & -- & -- & Object scale factor \\
Copies & -- & -- & Copies per object \\
\bottomrule
\end{tabular}
\end{table}


\subsection{Render randomized scenes}

Once all scenes have been generated, only the relevant frames are rendered, as shown in the output stage of~\autoref{fig:synthrender-diagram}. Rendering is restricted to an interval $[f_s, f_e]$, with \replaced{$0 \le f_s \le f_e < N_{\mathrm{scene}}$}{$0 \le f_s \le f_e \le n-1$}, avoiding rendering outside the region of interest. Furthermore, the pipeline renders multiple outputs simultaneously, including RGB images, semantic and instance segmentation masks, depth maps, normal maps, and simulation metadata containing poses, lighting parameters, and camera settings.

In addition, to reduce unnecessary hardware delays, instead of issuing a separate render call per scene, the entire sequence is rendered in a single batch. This avoids repeated loading of meshes, materials, and metadata into GPU memory, significantly reducing RAM and VRAM transfer overhead. \replaced{Output data is stored in per-frame HDF5 containers, BlenderProc's structured output format, which groups the rendered modalities and scene metadata into an annotation-format-agnostic intermediate representation. Format-specific exporters subsequently convert these containers into COCO, BOP, or YOLO datasets, allowing the same rendered outputs to be exported to another supported format without rerendering the scenes. This design was selected for modularity, data reuse, and reproducibility rather than as a runtime optimization. The computational cost of the post-render annotation stage is reported separately in~Figure~\ref{fig:synthrender_rendering_times}; no direct comparison against annotation export performed during rendering was conducted.}{Output data is stored in HDF5 files to reduce filesystem overhead and simplify downstream processing. These files are directly used to generate datasets in COCO, BOP, and YOLO formats.}

\section{Experimental Results and Analysis}
\label{sec:results}

This section presents \added{our newly introduced \acrshort{iasset} dataset, followed by} sim-to-real experiments on three datasets: robotics~\cite{horvath2022object}, automotive~\cite{martinez2024scap}, and \acrshort{iasset}. We evaluate the effect of the detection model and perform multiple ablation studies to identify the decisive factors that influence the sim-to-real gap, before comparing our results against the state-of-the-art. \added{ Unless stated otherwise, the ablation and benchmarking experiments use manually modeled \acrshort{CAD} assets, so as to isolate the effect of the studied factor from the quality of asset reconstruction; the reconstruction methodologies themselves (\acrshort{3DGS}, TRELLIS, and MeshyAI) are evaluated as \acrshort{real-to-sim} domain-adaptation strategies in \mbox{\autoref{sub:DA_results}} (\mbox{\autoref{tab:domain_adaptation_final}}).}

\subsection{Industrial Assets for Sim-to-Real Evaluation and Transfer}
\label{sub:iasset}

\begin{figure}[thpb]
    \centering
    \begin{subfigure}[b]{\linewidth}
        \centering
        \includegraphics[width=\linewidth]{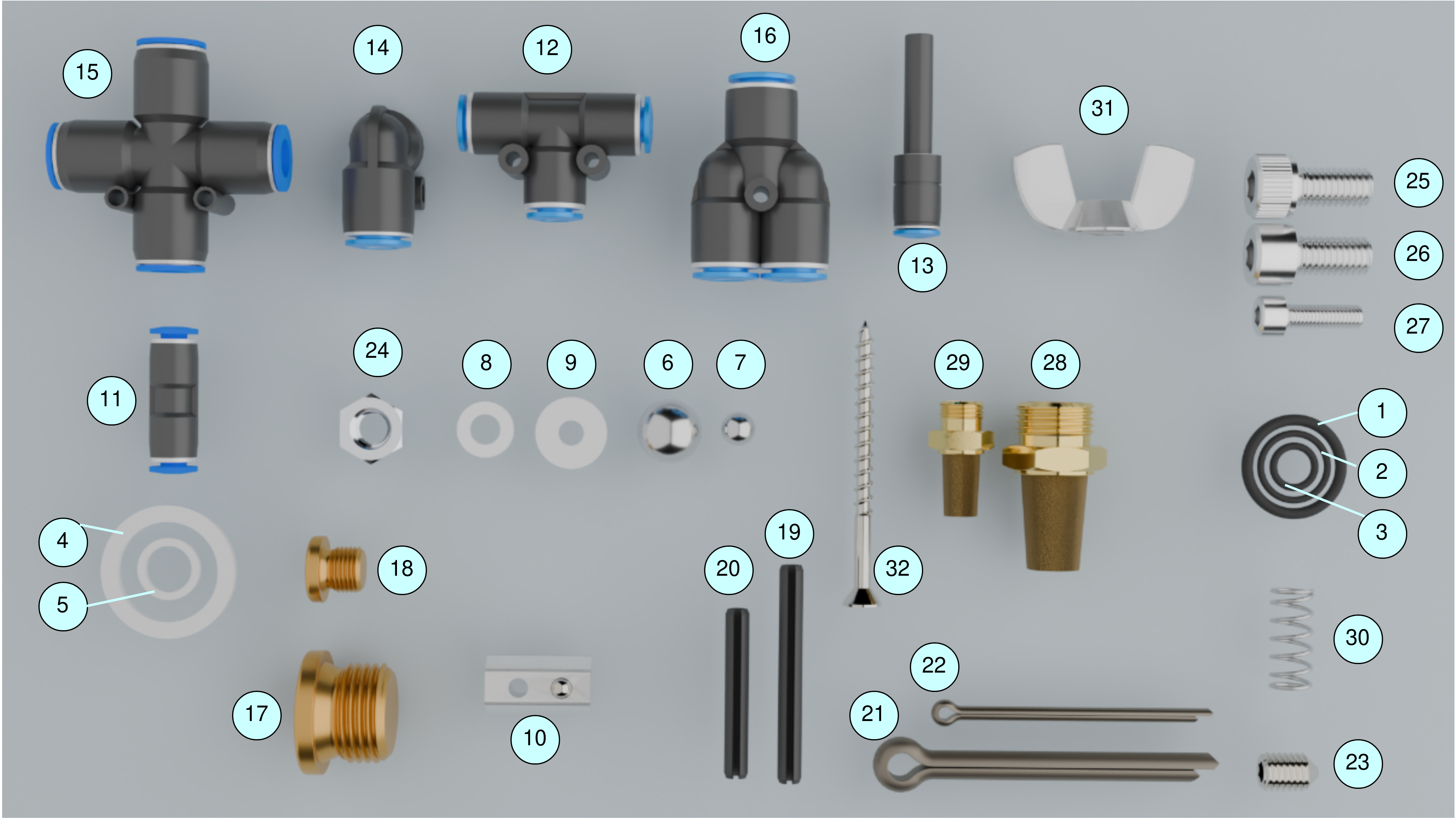}
        \caption{Synthetic CAD Renders}
        \label{fig:iasset_objects_synth}
    \end{subfigure}
    
    \vspace{0.2cm}
    
    \begin{subfigure}[b]{\linewidth}
        \centering
        \includegraphics[width=\linewidth]{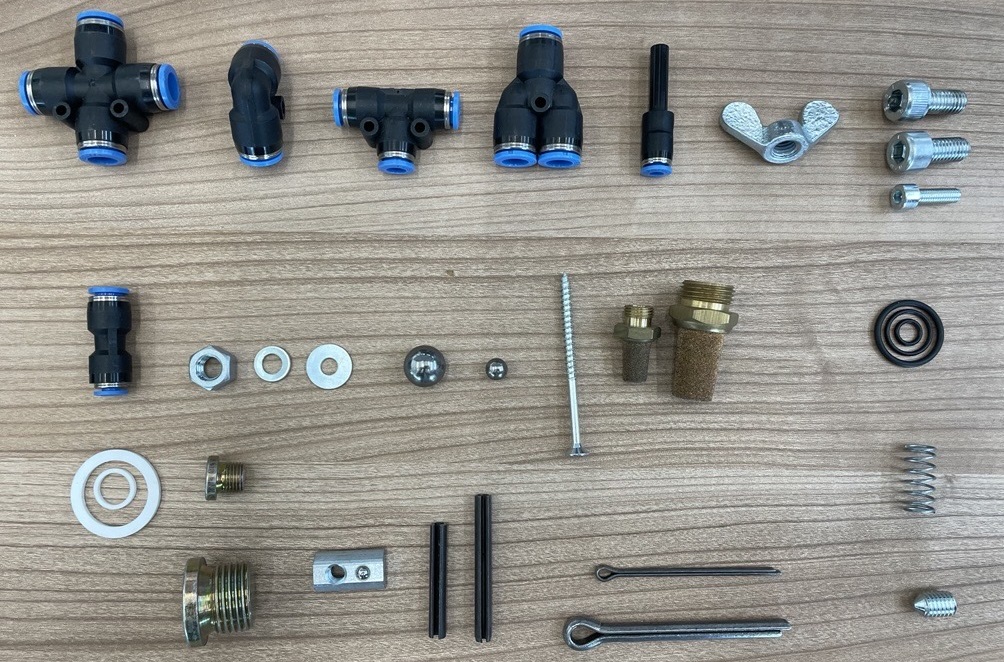} 
        \caption{Real Physical Objects}
        \label{fig:iasset_objects_real}
    \end{subfigure}
    \caption{\acrshort{iasset} objects comparison. (a) Synthetic renders of the 32 industrial CAD models used in this study, annotated with their corresponding class IDs. (b) The real physical components arranged in similar positions.}
    \label{fig:iasset}
\end{figure}

The \acrshort{iasset} industrial dataset features 32 classes of common mechanical and pneumatic components utilized in automation. A direct comparison of the synthetic CAD renders and their real physical counterparts is shown in~\autoref{fig:iasset}. \acrshort{iasset} follows a structured naming scheme to ensure clarity and traceability across all object classes. Each object identifier encodes both the component provenance (prefix) and, when applicable, its relative scale (suffix).~\autoref{tab:iasset_classes} shows the object classes along their system-level categorization and the source of every \acrshort{CAD} model.

\begin{table}[htbp]
\centering
\caption{Object classes, source and system-level family.}
\label{tab:iasset_classes}
\setlength{\tabcolsep}{4pt}
\resizebox{\columnwidth}{!}{%
\begin{tabular}{c l l l}
\hline
\textbf{ID} & \textbf{Class Name} & \textbf{Source} & \textbf{Family} \\
\hline
1  & C\_O\_Ring\_L           & Custom-Modeled    & Mechanical \\
2  & C\_O\_Ring\_M           & Custom-Modeled    & Mechanical \\
3  & C\_O\_Ring\_S           & Custom-Modeled    & Mechanical \\
4  & C\_Plastic\_Washer\_L   & Custom-Modeled    & Mechanical \\
5  & C\_Plastic\_Washer\_S   & Custom-Modeled    & Mechanical \\
6  & C\_Steel\_Ball\_L       & Custom-Modeled    & Mechanical \\
7  & C\_Steel\_Ball\_S       & Custom-Modeled    & Mechanical \\
8  & C\_Washer\_M5           & Custom-Modeled    & Mechanical \\
9  & C\_Washer\_M6           & Custom-Modeled    & Mechanical \\
\hline
10 & F\_Roll-in\_Nut\_M5     & FATH GmbH         & Mechanical \\
\hline
11 & FestoI                  & Festo SE \& Co. KG & Pneumatic \\
12 & FestoT                  & Festo SE \& Co. KG & Pneumatic \\
13 & Festo\_Torch            & Festo SE \& Co. KG & Pneumatic \\
14 & FestoV                  & Festo SE \& Co. KG & Pneumatic \\
15 & FestoX                  & Festo SE \& Co. KG & Pneumatic \\
16 & FestoY                  & Festo SE \& Co. KG & Pneumatic \\
\hline
17 & GF\_Collar\_L            & GlobalFastener Inc.& Mechanical \\
18 & GF\_Collar\_S            & GlobalFastener Inc.& Mechanical \\
19 & GF\_Slotted\_Pin\_L      & GlobalFastener Inc.& Mechanical \\
20 & GF\_Slotted\_Pin\_S      & GlobalFastener Inc.& Mechanical \\
21 & GF\_Split\_Pin\_L        & GlobalFastener Inc.& Mechanical \\
22 & GF\_Split\_Pin\_S        & GlobalFastener Inc.& Mechanical \\
23 & GF\_Cone\_Screw\_M8      & GlobalFastener Inc.& Mechanical \\
24 & GF\_Hexagon\_Nut         & GlobalFastener Inc.& Mechanical \\
25 & GF\_Knurled\_Screw\_M8   & GlobalFastener Inc.& Mechanical \\
26 & GF\_Plain\_Screw\_M8     & GlobalFastener Inc.& Mechanical \\
27 & GF\_Screw\_M5            & GlobalFastener Inc.& Mechanical \\
\hline
28 & MM\_Silencer\_L          & McMaster-Carr Supply Co. & Pneumatic \\
29 & MM\_Silencer\_S          & McMaster-Carr Supply Co. & Pneumatic \\
30 & MM\_Spring               & McMaster-Carr Supply Co. & Mechanical \\
31 & MM\_Wing                 & McMaster-Carr Supply Co. & Mechanical \\
32 & MM\_Wood\_Screw          & McMaster-Carr Supply Co. & Mechanical \\
\hline
\end{tabular}
} %
\end{table}

\acrshort{iasset} is designed to evaluate object detection under realistic sensing conditions and to study \acrshort{sim-to-real} transfer using high-quality synthetic assets. The objects in \acrshort{iasset} feature a unique set of characteristics:
\begin{itemize}
    \item \textbf{Semi-Uncontrolled Conditions and Extensibility:} Includes diverse materials, geometries, and textures, with environmental variables such as direct sunlight, camera-object poses, and changing backgrounds. The 32 selected industrial parts (e.g., pneumatic components, fasteners, and seals) are widely accessible, facilitating contributions of additional test data across new environments and imaging systems.
    
   \item \textbf{Challenging Class Selection:} Several object classes share similar    materials or geometries, increasing classification difficulty. Crucially, several parts differ primarily in scale (e.g., 6 vs. 7, 19 vs. 20, 21 vs. 22, and 28 vs. 29), testing a model's ability to differentiate size under semi-uncontrolled physical conditions. Additionally, multiple instances per class are present in the test set to introduce intra-class deviations such as scratches, rust, and other surface variations. These features, in combination
    with varying environmental conditions, make \acrshort{iasset} a challenging test case for \acrshort{sim-to-real} algorithms.

    \item \textbf{Bidirectional \acrshort{Sim-to-Real}:} 3D assets in \acrshort{iasset} are provided not just as \acrshort{CAD} models but also as reconstructed meshes via multiple methods to support research on bidirectional \acrshort{sim-to-real} gap narrowing.
\end{itemize}

The 3D representations of the 32 mechanically relevant components are collected from five sources, as per~\autoref{tab:iasset_classes}. All objects include ideal 3D geometry from \acrshort{CAD} models and 2D-to-3D reconstructed models from all methods investigated in this work (i.e., 3DGS, TRELLIS, and MeshyAI-textured \acrshort{CAD}s). This unique collection of \acrshort{CAD}-based and reconstructed geometries, paired with manual and generated textures and annotated real RGB-D imagery, enables systematic evaluation of novel bidirectional \acrshort{sim-to-real} approaches. Moreover, a series of 2D scans used as input for the generative models are included in the dataset. This facilitates the extension of the benchmark of GenAI models without physically sourcing the parts.

\acrshort{iasset} was intentionally constructed with near-uniform class frequencies ($\sim$615 instances per class) to minimize class imbalance effects during evaluation. The real dataset comprises 508 RGB-D images captured at $1224\times1024$ resolution using a Zivid 2+ MR60 sensor. To systematically evaluate detection performance under varying clutter conditions, the dataset is stratified into three complexity tiers: 96 single-object (1~Obj.) images, 210 single-instance (1x~Inst.) images containing exactly one instance of each of the 32 classes, and 202 double-instance (2x~Inst.) images containing exactly two instances of each class. \added{ Here, ``balanced'' refers to the per-class \emph{instance} frequency ($\sim$615 instances per class), not to equal image counts across the three tiers. Because each single-instance and each double-instance image contains, respectively, one and two instances of \emph{every} class, these two tiers alone determine the class balance ($\approx$210 and $\approx$404 instances per class); the single-object images contribute a single instance each and are included primarily to enable isolated single-object analysis. Their smaller number is therefore by design and does not affect the class balance.}

Due to partial boundary truncation and physical occlusions in dense scenes, this composition yields 19,672 valid annotations overall. Objects are predominantly positioned near the image center to reflect typical industrial tabletop acquisition setups, while bounding boxes generally occupy less than 15\% of the image dimensions.

\begin{figure*}[!t]
    \centering
    
    \begin{subfigure}[c]{0.03\linewidth}
        \centering
        \rotatebox{90}{\textbf{Controlled}}
    \end{subfigure}\hfill
    \begin{subfigure}[c]{0.31\linewidth}
        \centering
        \includegraphics[width=\linewidth]{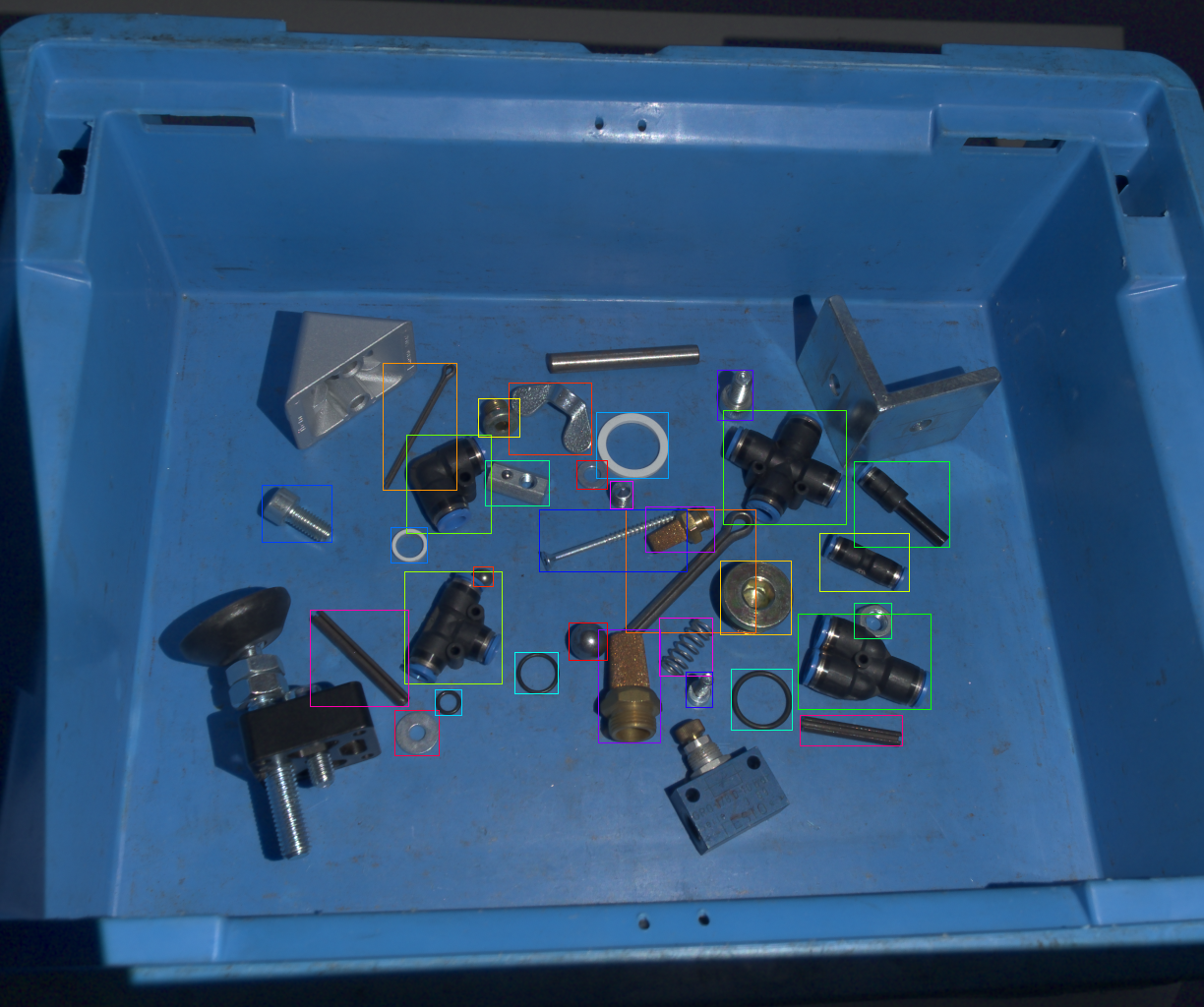}
    \end{subfigure}\hfill
    \begin{subfigure}[c]{0.31\linewidth}
        \centering
        \includegraphics[width=\linewidth]{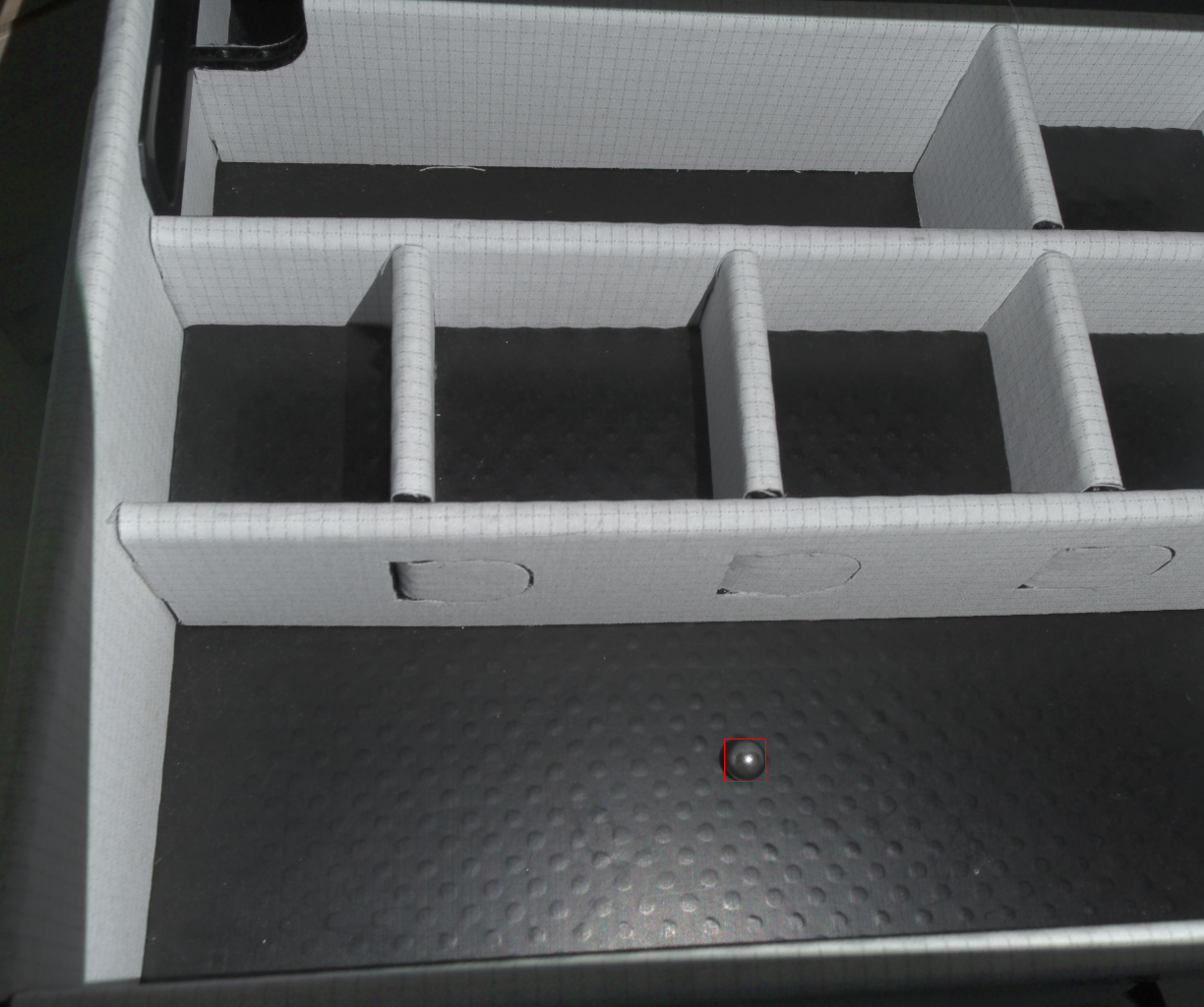}
    \end{subfigure}\hfill
    \begin{subfigure}[c]{0.31\linewidth}
        \centering
        \includegraphics[width=\linewidth]{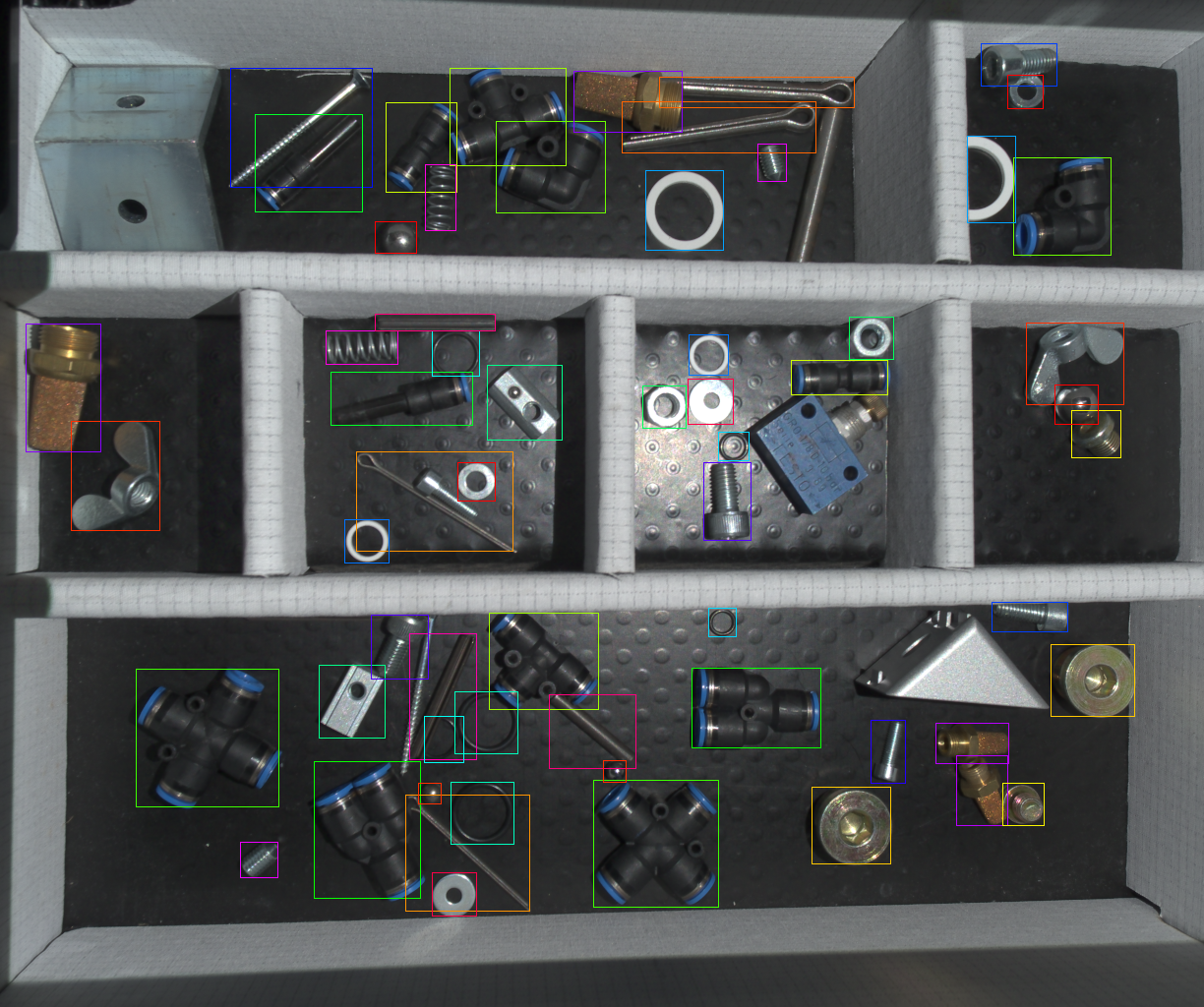}
    \end{subfigure}
    
    \vspace{0.2cm} 
    
    \begin{subfigure}[c]{0.03\linewidth}
        \centering
        \rotatebox{90}{\textbf{Sunlight}}
    \end{subfigure}\hfill
    \begin{subfigure}[c]{0.31\linewidth}
        \centering
        \includegraphics[width=\linewidth]{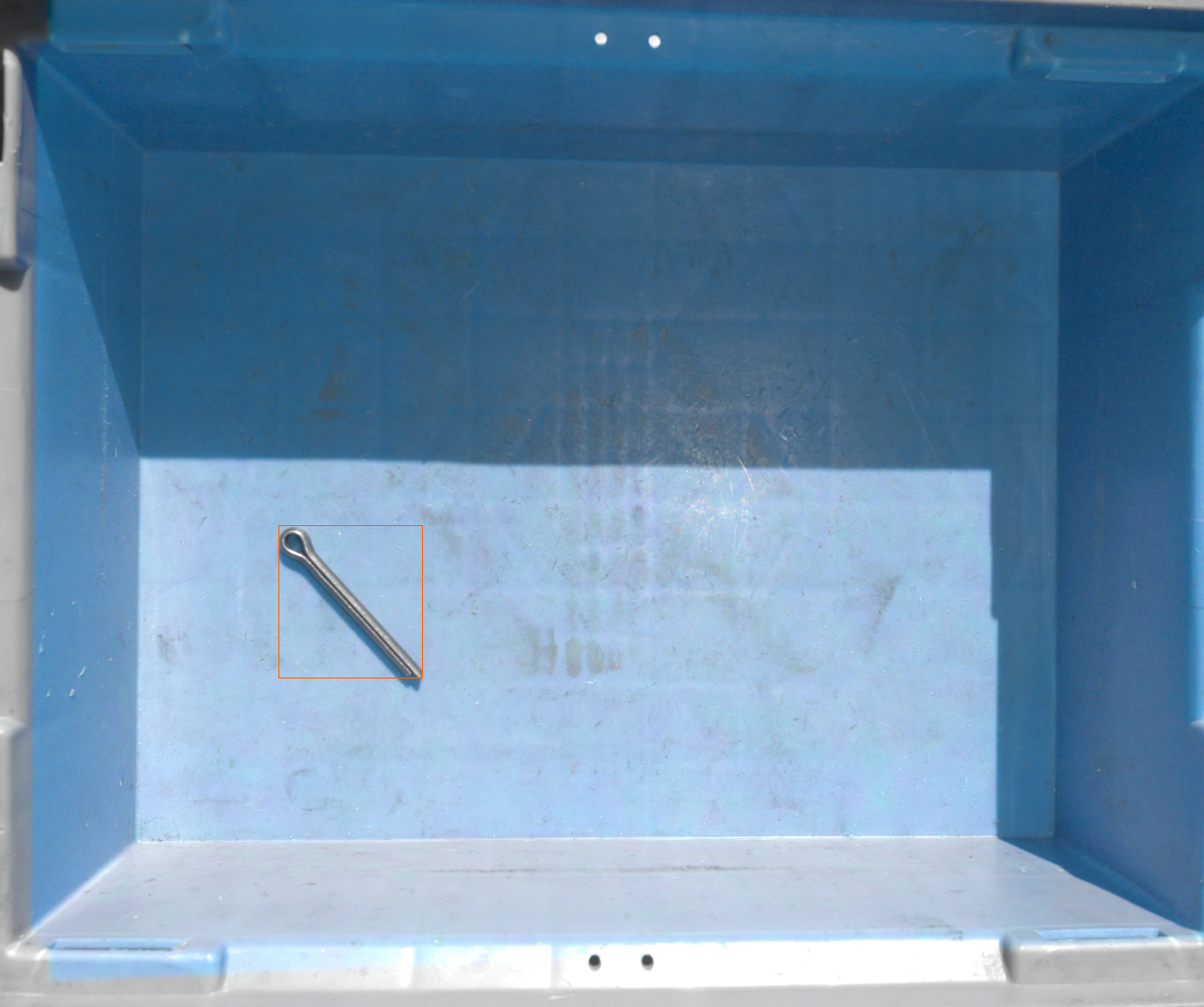}
    \end{subfigure}\hfill
    \begin{subfigure}[c]{0.31\linewidth}
        \centering
        \includegraphics[width=\linewidth]{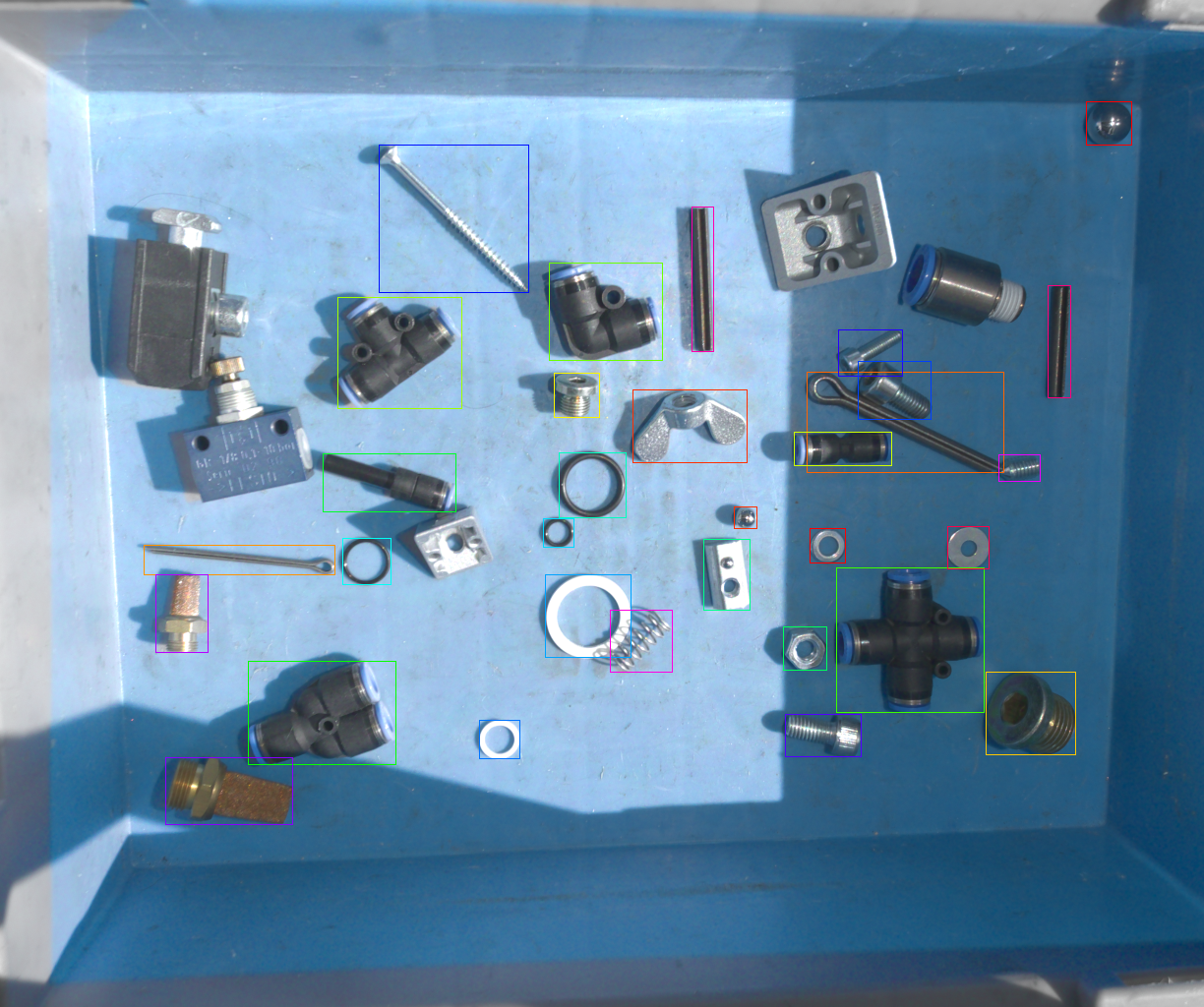}
    \end{subfigure}\hfill
    \begin{subfigure}[c]{0.31\linewidth}
        \centering
        \includegraphics[width=\linewidth]{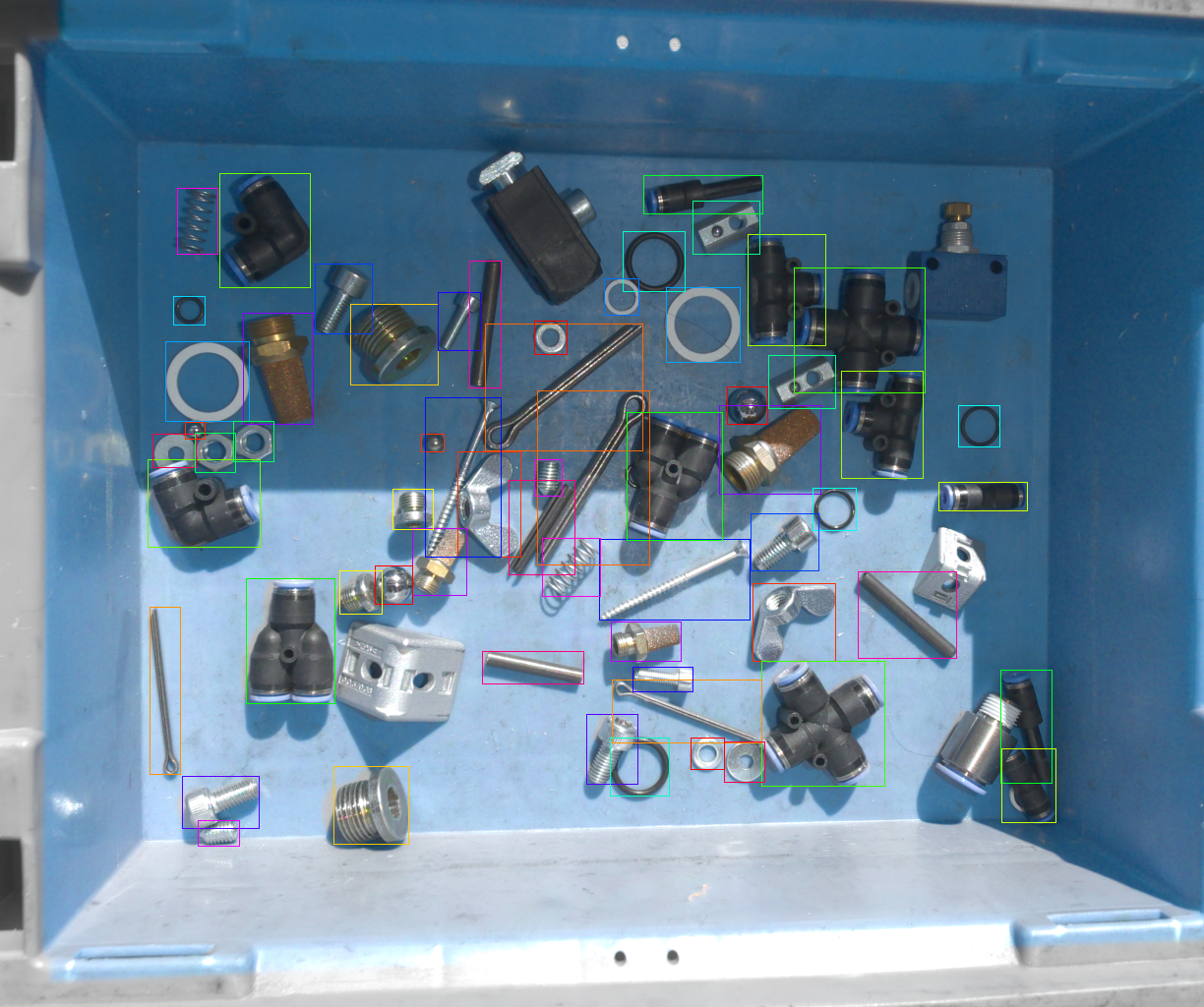}
    \end{subfigure}
    
    \vspace{0.2cm} 
    
    \begin{subfigure}[c]{0.03\linewidth}
        \centering
        \rotatebox{90}{\textbf{Backgrounds}}
    \end{subfigure}\hfill
    \begin{subfigure}[c]{0.31\linewidth}
        \centering
        \includegraphics[width=\linewidth]{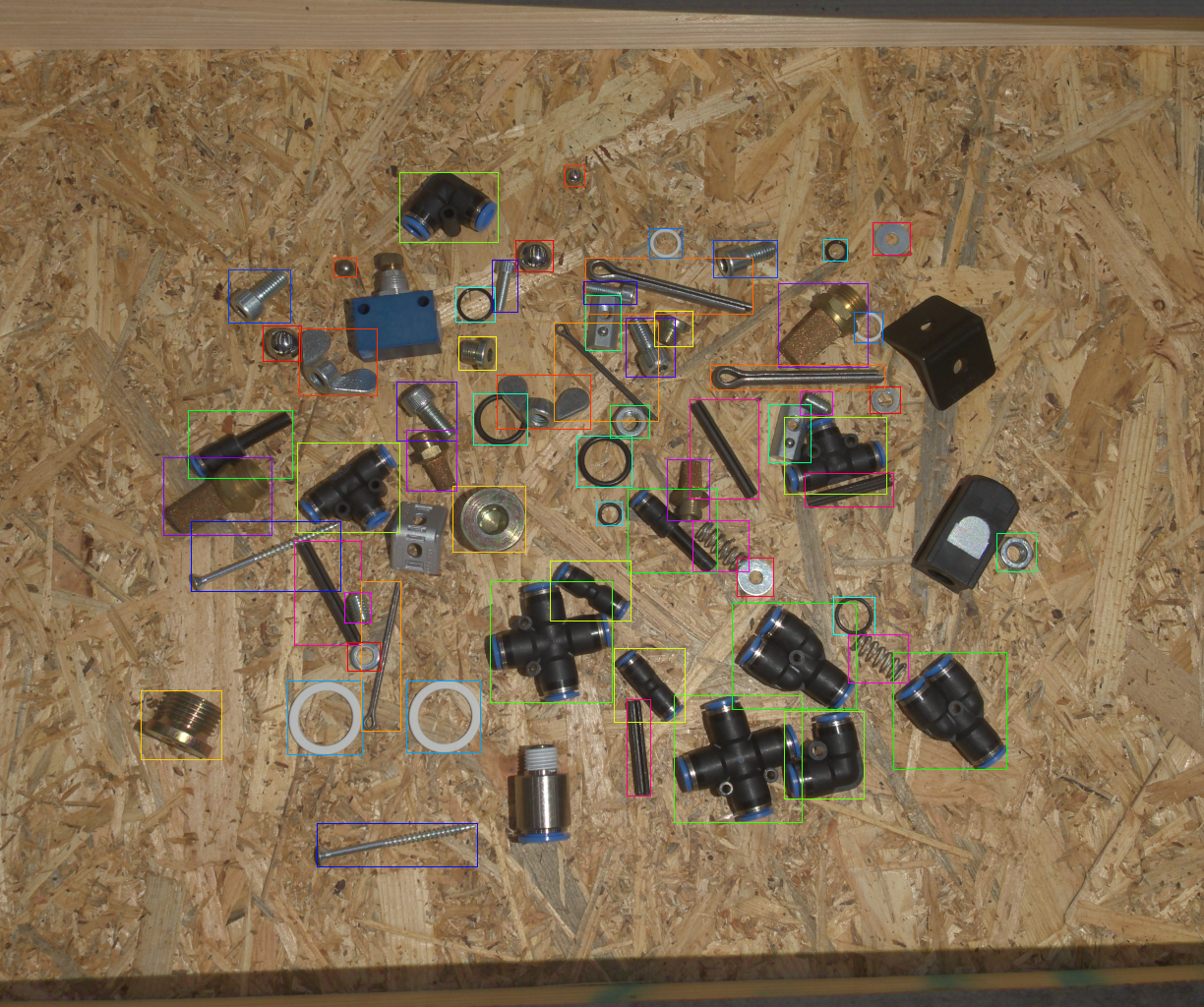}
    \end{subfigure}\hfill
    \begin{subfigure}[c]{0.31\linewidth}
        \centering
        \includegraphics[width=\linewidth]{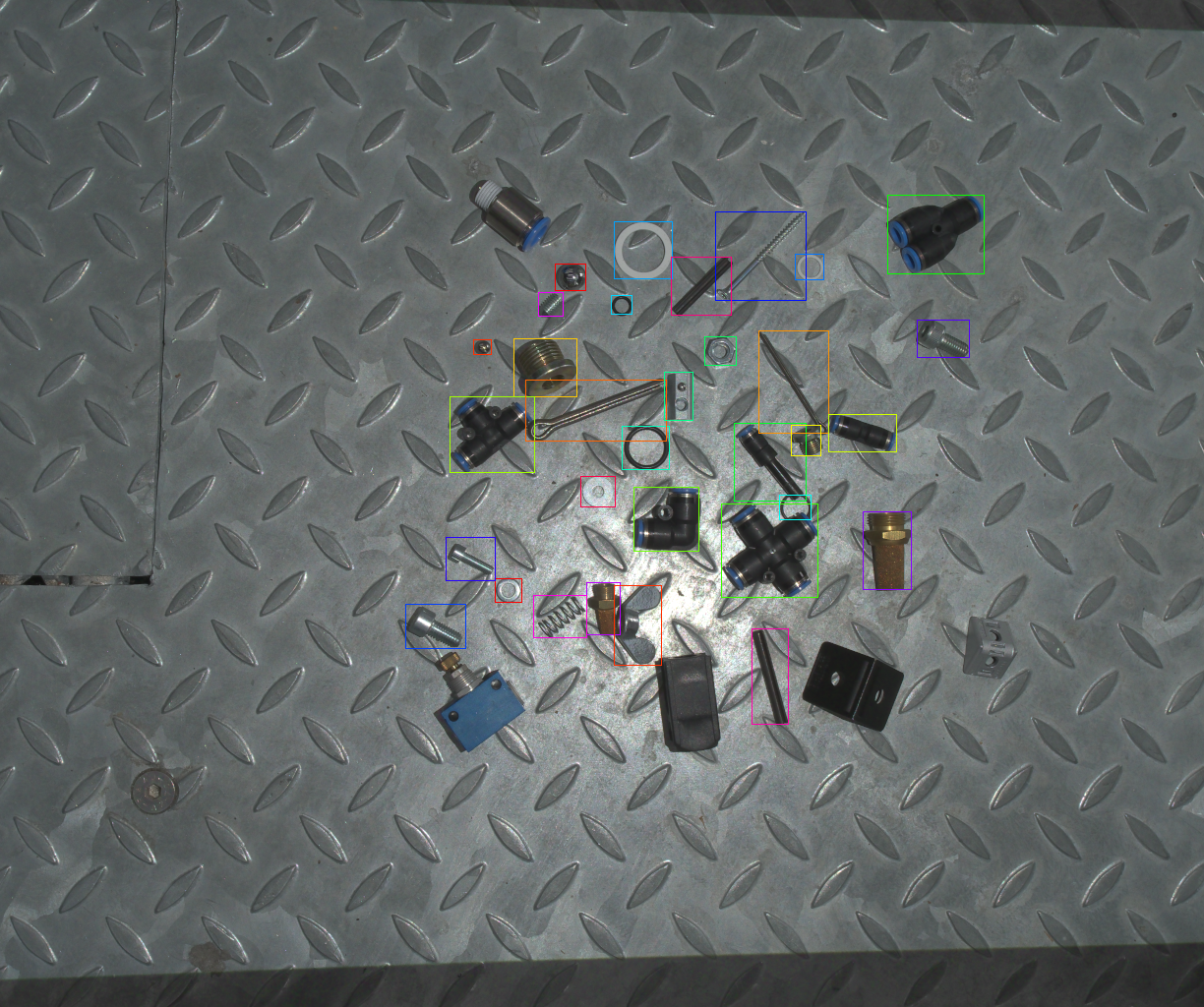}
    \end{subfigure}\hfill
    \begin{subfigure}[c]{0.31\linewidth}
        \centering
        \includegraphics[width=\linewidth]{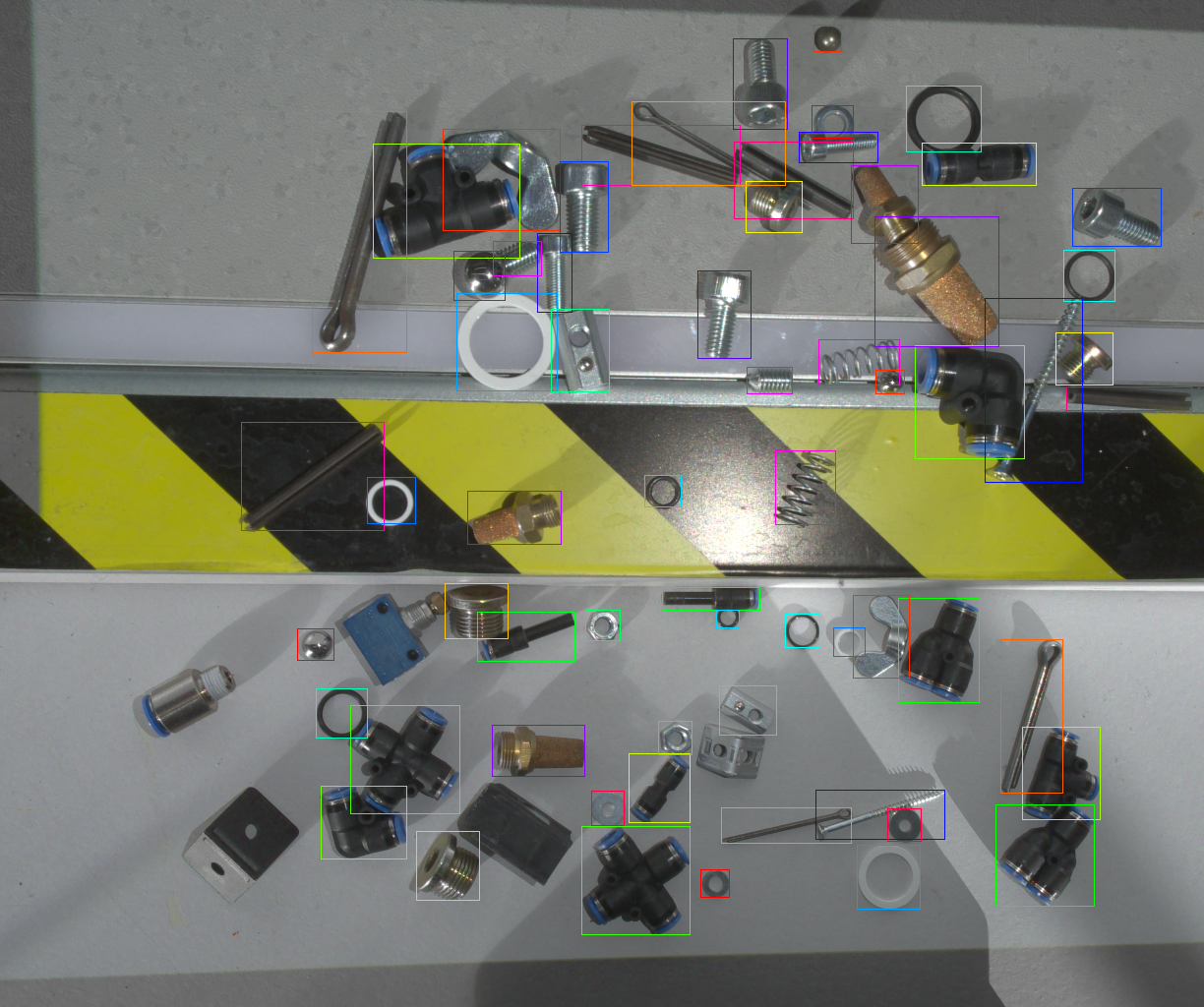}
    \end{subfigure}
    
    \vspace{0.2cm} 
    
    \begin{subfigure}[c]{0.03\linewidth}
        \centering
        \rotatebox{90}{\textbf{Robot}}
    \end{subfigure}\hfill
    \begin{subfigure}[c]{0.31\linewidth}
        \centering
        \includegraphics[width=\linewidth]{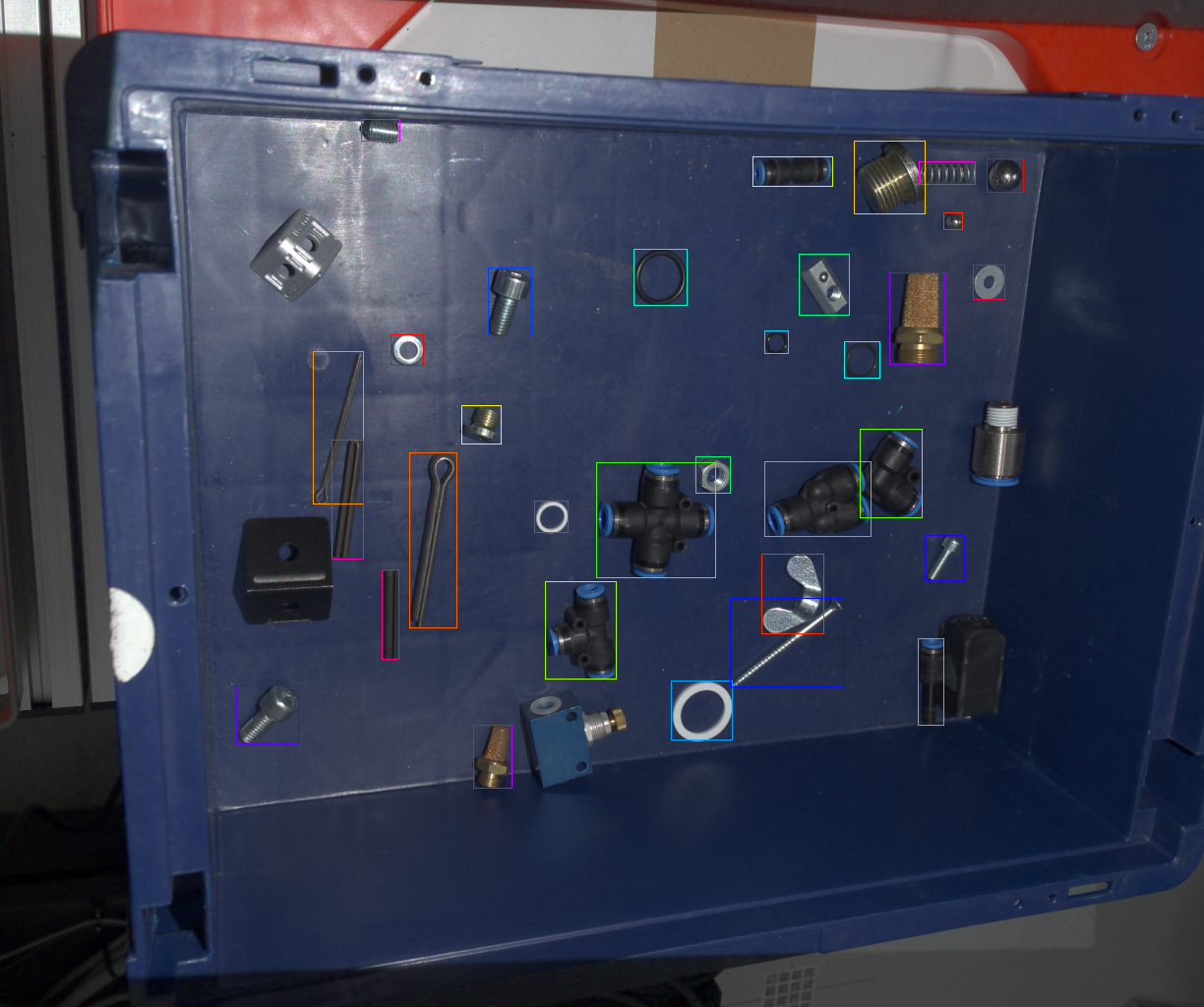}
    \end{subfigure}\hfill
    \begin{subfigure}[c]{0.31\linewidth}
        \centering
        \includegraphics[width=\linewidth]{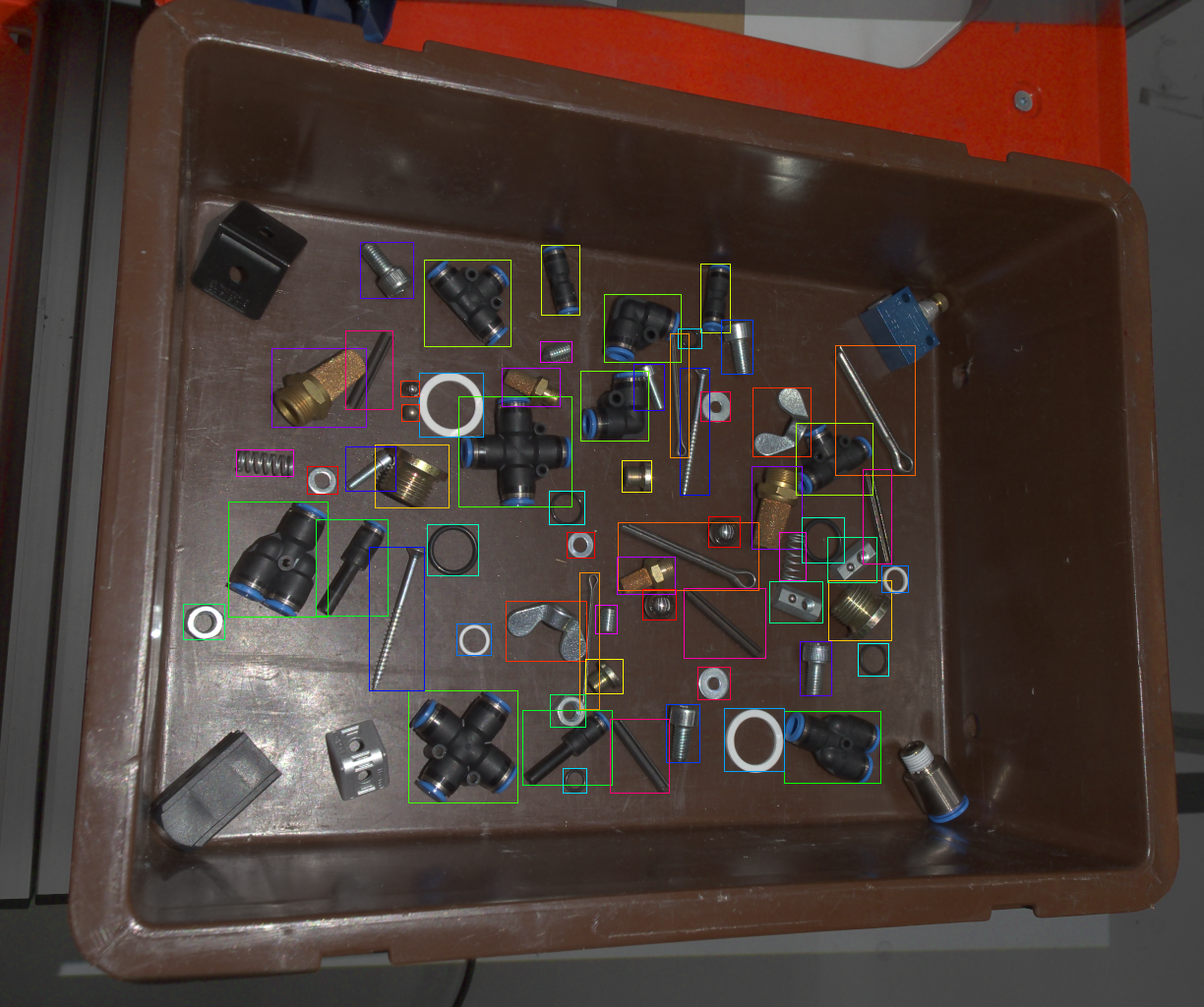}
    \end{subfigure}\hfill
    \begin{subfigure}[c]{0.31\linewidth}
        \centering
        \includegraphics[width=\linewidth]{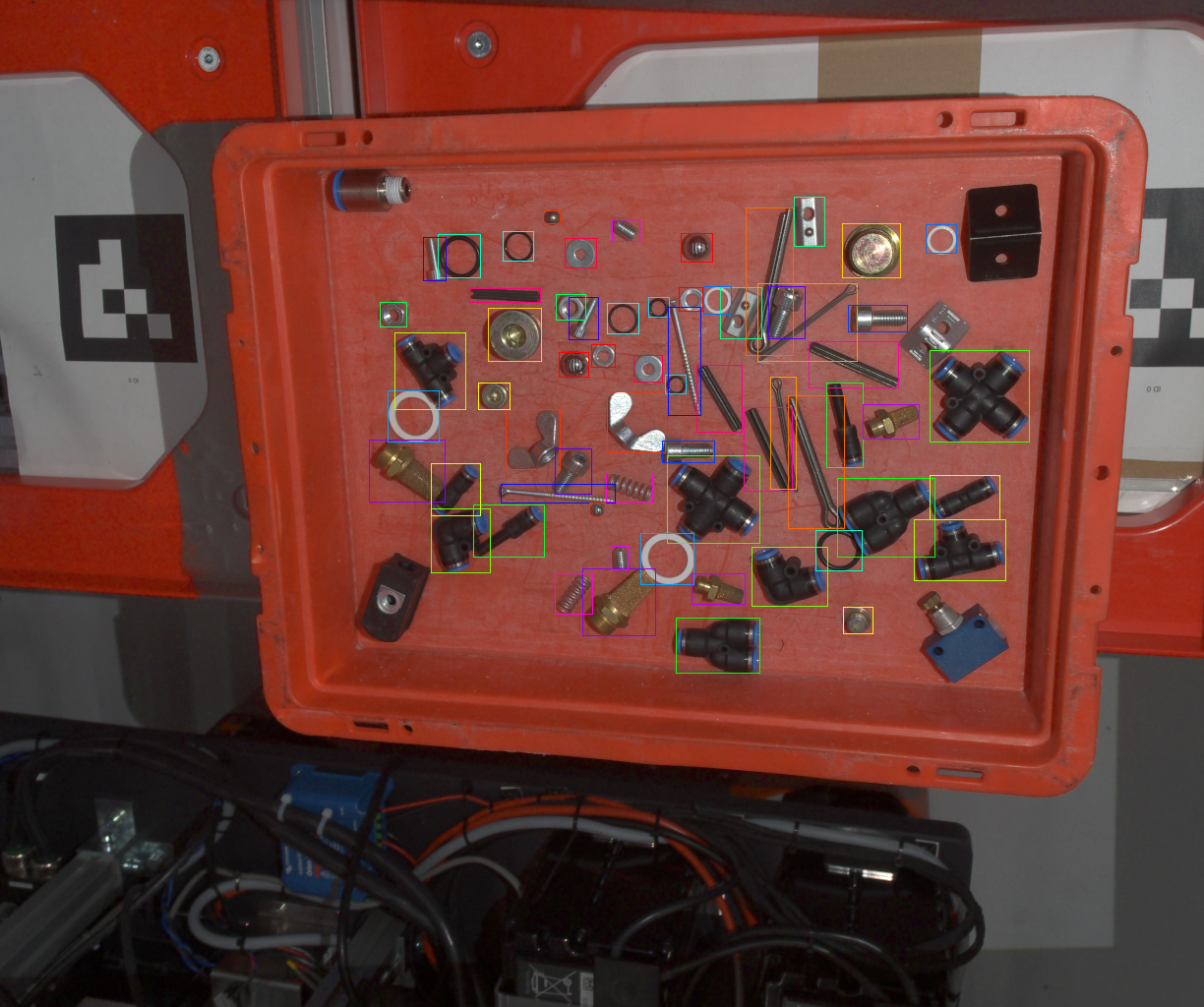}
    \end{subfigure}\hfill
    
    \caption{\label{fig:iasset_samples}Representative sample images from the \acrshort{iasset} dataset showcasing diverse real-world test scenes across our four different domains. Ground truth object detection bounding boxes are overlaid to demonstrate the density, scale variations, and annotation quality in highly cluttered scenarios.}
\end{figure*}

The test scenes cover single- and multi-object detection across four acquisition domains, summarized in~\autoref{tab:iasset_scenes}, designed to test different environmental variables. Specifically, the \textit{Controlled} domain provides a baseline using standard logistic containers under uniform studio lighting; the \textit{Sunlight} domain introduces harsh, directional natural lighting; the \textit{Backgrounds} domain tests texture invariance using five distinct industrial surfaces; and in the \textit{Robot} domain the camera is mounted on a robot and objects are placed inside six differently colored \acrshort{KLT} boxes. To illustrate the complexity and visual diversity of these environments, \autoref{fig:iasset_samples} displays a grid of representative real-world test scenes, complete with their ground truth bounding box annotations. Real images include COCO/YOLO bounding boxes, while synthetic data provides pixel-perfect depth, instance masks, and 6D poses.

\begin{table}[!t]
\centering
\caption{\label{tab:iasset_scenes}Distribution of \acrshort{iasset} \deleted{real} test images across environmental domains and object densities (Single Object / Single Instance / Double Instance).}
\resizebox{\columnwidth}{!}{%
\begin{tabular}{lcccc}
\toprule
\multirow{2}{*}{\textbf{Domain}} & \multirow{2}{*}{\textbf{Total}} & \multicolumn{3}{c}{\textbf{Image Composition}} \\
\cmidrule(lr){3-5}
& & \textbf{1 Obj.} & \textbf{1x Inst.} & \textbf{2x Inst.} \\
\midrule
Controlled & 101 & 64 & 21 & 16 \\
Sunlight & 67 & 32 & 20 & 15 \\
Backgrounds & 100 & 0 & 49 & 51 \\
Robot & 240 & 0 & 120 & 120 \\
\midrule
\textbf{Total} & \textbf{508} & \textbf{96} & \textbf{210} & \textbf{202} \\
\bottomrule
\end{tabular}%
}
\end{table} 

\subsection{Experimental Setup: Hyperparameters and Randomization Regimes}
\label{sub:exp_setup_hyper_and_rand_regimes}

In line with our data-centric approach, all experiments employ off-the-shelf detection models, including YOLO and DEIM variants, with standard hyperparameter configurations and no task-specific fine-tuning; all training experiments are run six times and the reported values are averaged. While performance may be further improved through hyperparameter optimization, such tuning is orthogonal to the contributions of this work and is therefore left out of scope. The hyperparameters used across all experiments, unless explicitly stated otherwise, are listed in~\autoref{tab:hyperparams}.

\begin{table}[htpb]
\centering
\caption{Training hyperparameters used across our experiments.}
\label{tab:hyperparams}
\begin{tabular}{lcc}
\toprule
\textbf{Hyperparameter}         & \textbf{Value} \\
\midrule
Epochs                          & 500            \\
Train image size                      & 1024$\times$1024               \\
Batch size                      & 16              \\
Learning rate                   & 0.01             \\
Patience                        & 30             \\
\bottomrule
\end{tabular}
\end{table}

In a series of ablation studies, as listed in ~\autoref{tab:ablation_config}, we evaluate which randomization parameters available in SynthRender, help reduce the \acrshort{sim-to-real} gap. Based on Table~\ref{tab:synthrender_params}, the \textbf{Baseline} configuration disables physics simulation, RGB light variation, and camera intrinsics for the target camera; randomization of target object textures is also disabled, while background texture variation remains active, following previous research \cite{araya-martinez2025xai}. Camera distance and elevation are set to spatially relevant values aligned with the target domain, alongside a tuned target anchor center and radius. The target object count is set to one with one copy per instance, yielding two target object instances per frame for each object class. The distractor count is set to its maximum value, i.e. all objects in the distractor database are used to promote scene diversity without adding excessive clutter. Target object scale, spatial orientation, and position are configured to ensure that the majority of objects remain within the frame boundaries across all rendered images. These configuration choices reflect a plausible initial \acrshort{GDR} scenario, following applicable best practices from existing literature \cite{horvath2022object, Zhu2025icra, martinez2024scap}.

\begin{table*}[hpbt]
\centering
\caption{Ablation study configurations for domain randomization in SynthRender. Only parameters that vary across ablation configurations are shown. 
Fixed parameters, such as camera to object distance, are 
constrained to realistic ranges representative of the target evaluation 
domain. The parameters listed in~Table~\ref{tab:synthrender_params} are constrained to a defined variation range.
\cmark~indicates the feature is enabled; \xmark~indicates it is disabled. The Additive Ablation category builds features incrementally, while Reverse Ablation removes specific features from the full setup or adds them individually to the target texture randomization.}
\label{tab:ablation_config}
\setlength{\tabcolsep}{5pt}
\begin{tabular}{l ccccccc cccccccc}
\toprule
& \multicolumn{7}{c}{\textbf{Additive Ablation}} & \multicolumn{8}{c}{\textbf{Reverse Ablation}} \\
\cmidrule(lr){2-8} \cmidrule(lr){9-16}
& \rotatebox{90}{\textbf{Baseline}} 
& \rotatebox{90}{\textbf{RGB}}
& \rotatebox{90}{\textbf{RGB + Exp}}
& \rotatebox{90}{\textbf{Physics}}
& \rotatebox{90}{\textbf{Phy+RGB+EXP}}
& \rotatebox{90}{\replaced{\textbf{P.+R.+E.+Intri.}}{\textbf{P.R.E.+Intri.}}}
& \rotatebox{90}{\textbf{\shortstack[l]{All + Random\\Texture}}}
& \rotatebox{90}{\textbf{All minus Intri.}}
& \rotatebox{90}{\textbf{All minus Exp}}
& \rotatebox{90}{\textbf{All minus RGB}}
& \rotatebox{90}{\textbf{RandTex + Phy}}
& \rotatebox{90}{\textbf{All minus Physics}}
& \rotatebox{90}{\textbf{RandTex + RGB}}
& \rotatebox{90}{\textbf{RandTex + Exp}}
& \rotatebox{90}{\textbf{RandTex Only}}
\\
\midrule
\textbf{Scale}                      & \cmark & \cmark & \cmark & \cmark & \cmark & \cmark & \cmark & \cmark & \cmark & \cmark & \cmark & \cmark & \cmark & \cmark & \cmark \\
\textbf{Position \& Orientation}    & \cmark & \cmark & \cmark & \cmark & \cmark & \cmark & \cmark & \cmark & \cmark & \cmark & \cmark & \cmark & \cmark & \cmark & \cmark \\
\textbf{Background Texture}                 & \cmark & \cmark & \cmark & \cmark & \cmark & \cmark & \cmark & \cmark & \cmark & \cmark & \cmark & \cmark & \cmark & \cmark & \cmark \\
\textbf{RGB Light}                  & \xmark & \cmark & \cmark & \xmark & \cmark & \cmark & \cmark & \cmark & \cmark & \xmark & \xmark & \cmark & \cmark & \xmark & \xmark \\
\textbf{Exponential Light}          & \xmark & \xmark & \cmark & \xmark & \cmark & \cmark & \cmark & \cmark & \xmark & \cmark & \xmark & \cmark & \xmark & \cmark & \xmark \\
\textbf{Physics}                    & \xmark & \xmark & \xmark & \cmark & \cmark & \cmark & \cmark & \cmark & \cmark & \cmark & \cmark & \xmark & \xmark & \xmark & \xmark \\
\textbf{Camera Intrinsics}          & \xmark & \xmark & \xmark & \xmark & \xmark & \cmark & \cmark & \xmark & \cmark & \cmark & \xmark & \cmark & \xmark & \xmark & \xmark \\
\textbf{Target Object Texture}             & \xmark & \xmark & \xmark & \xmark & \xmark & \xmark & \cmark & \cmark & \cmark & \cmark & \cmark & \cmark & \cmark & \cmark & \cmark \\
\bottomrule
\end{tabular}

\end{table*}

From this baseline, the ablation configurations are denoted as follows: \textbf{RGB} refers to RGB light color randomization; \textbf{Exp} to exponential light intensity sampling; \textbf{Physics} to physics simulation enabled in isolation; \textbf{Phy+RGB+Exp} to the combined activation of physics, RGB, and exponential light randomization; \replaced{\textbf{P.+R.+E.+Intri.}}{\textbf{P.R.E.+Intri.}} to the further addition of the camera intrinsics of the acquisition camera, specifically the Zivid 2+ MR60 used to capture the \acrshort{iasset} test set, on top of physics, RGB, and exponential light variation; and \textbf{All} to the full configuration comprising physics, RGB, exponential light, camera intrinsics, and randomized \acrshort{PBR} materials assigned to target objects. Physics simulation in this context denotes gravity-driven object placement via Blender rigid body dynamics, as described in~\autoref{sub:synthrender:scene_setup}.

\added{Moreover, to assess the statistical reliability of the reported detection performance across our experiments, each model configuration was retrained $n=6$ times with independent random seeds controlling weight initialization, following common practice in computer vision benchmarking where multiple training seeds are used to characterize run-to-run variance in mAP-based metrics \citep{roboflow100vl2025}. For each metric (mAP@50, mAP@50:95, recall and precision), we report the mean together with the 95\% percentile bootstrap confidence interval, computed using $10{,}000$ resamples of the $n=6$ per-seed scores \citep{efron1993bootstrap}. This procedure quantifies the variability attributable to stochastic training dynamics without requiring distributional assumptions on the underlying score distribution.}

\subsection{Real and Synthetic Datasets}

\autoref{tab:dataset_split} summarizes the dataset partitioning used in this work. Except for the few-shot experiments, all models are trained exclusively on synthetic data and evaluated exclusively on real data. Consequently real images are used exclusively as the test set $\mathcal{D}_{\text{test}}^{\text{real}}$, ensuring that evaluation reflects transfer from synthetic training data to real industrial imagery. The test set includes variation in object pose, background, illumination, and occlusion to reflect realistic \replaced{detection}{inspection} conditions.

\begin{table}[!H]
\centering
\caption{Dataset partitioning used for sim-to-real evaluation on \acrshort{iasset}.}
\label{tab:dataset_split}
\renewcommand{\arraystretch}{1.5}
\begin{tabular}{llccc}
\toprule
\textbf{Split} & \textbf{Notation} & \textbf{Domain} & \textbf{Images}  \\
\midrule
Training   & $\mathcal{D}_{\text{train}}^{\text{syn}}$ & Synthetic & 3{,}200 \\
Validation & $\mathcal{D}_{\text{val}}^{\text{syn}}$   & Synthetic & 800  \\
\midrule
Total synthetic & $\mathcal{D}^{\text{syn}}$ & Synthetic & 4{,}000 \\
\midrule
Test       & $\mathcal{D}_{\text{test}}^{\text{real}}$ & Real      & 508  \\
\bottomrule
\end{tabular}
\end{table}

Synthetic datasets are generated with SynthRender and are used exclusively for training and validation. Except for the ablation study on training set size, the training set contains 4{,}000  synthetic images which is split into $\mathcal{D}_{\text{train}}^{\text{syn}}$ and $\mathcal{D}_{\text{val}}^{\text{syn}}$ using an 80:20 ratio, resulting in 3{,}200 training images and 800 validation images. Each synthetic image may contain zero, one, or multiple instances of the same class, arranged randomly across the scene.

Similarly to previous literature by Zhu et al.~\cite{Zhu2025icra}, the total synthetic dataset size of 4{,}000 images was selected through an ablation study that examines the tradeoff between dataset size and model performance, which exhibits diminishing returns as the training set grows (Figure~\ref{fig:training_size_ablation_comparison}). Detection improves steadily at low image counts and the marginal gain decreases as the set approaches the selected size; this scaling behaviour is analyzed in detail in \autoref{sub:ablations}.



\subsection{SynthRender Across Detector Architectures}
\label{sub:synth_render_across_models}

In these experiments, we evaluate the ability of SynthRender's synthetic data to support \acrshort{sim-to-real} transfer across three state-of-the-art object detectors: YOLOv8~\cite{Hussain2023}, YOLO11, and DEIM~\cite{Huang_2025_CVPR}. The experimental setup follows the data partitioning for \acrshort{iasset} synthetic (train) and real (test) sets described in \autoref{tab:dataset_split}, and the models follow the hyperparameters of \autoref{tab:hyperparams}. We use CAD models as geometric references and assign textures manually to the target objects. To ensure that any performance differences are attributable solely to the choice of architecture rather than to variations in the synthetic data generation process, a fixed baseline data configuration is established.

\begin{figure}[htpb]
    \centering
    \centerline{\includegraphics[width=1\linewidth]{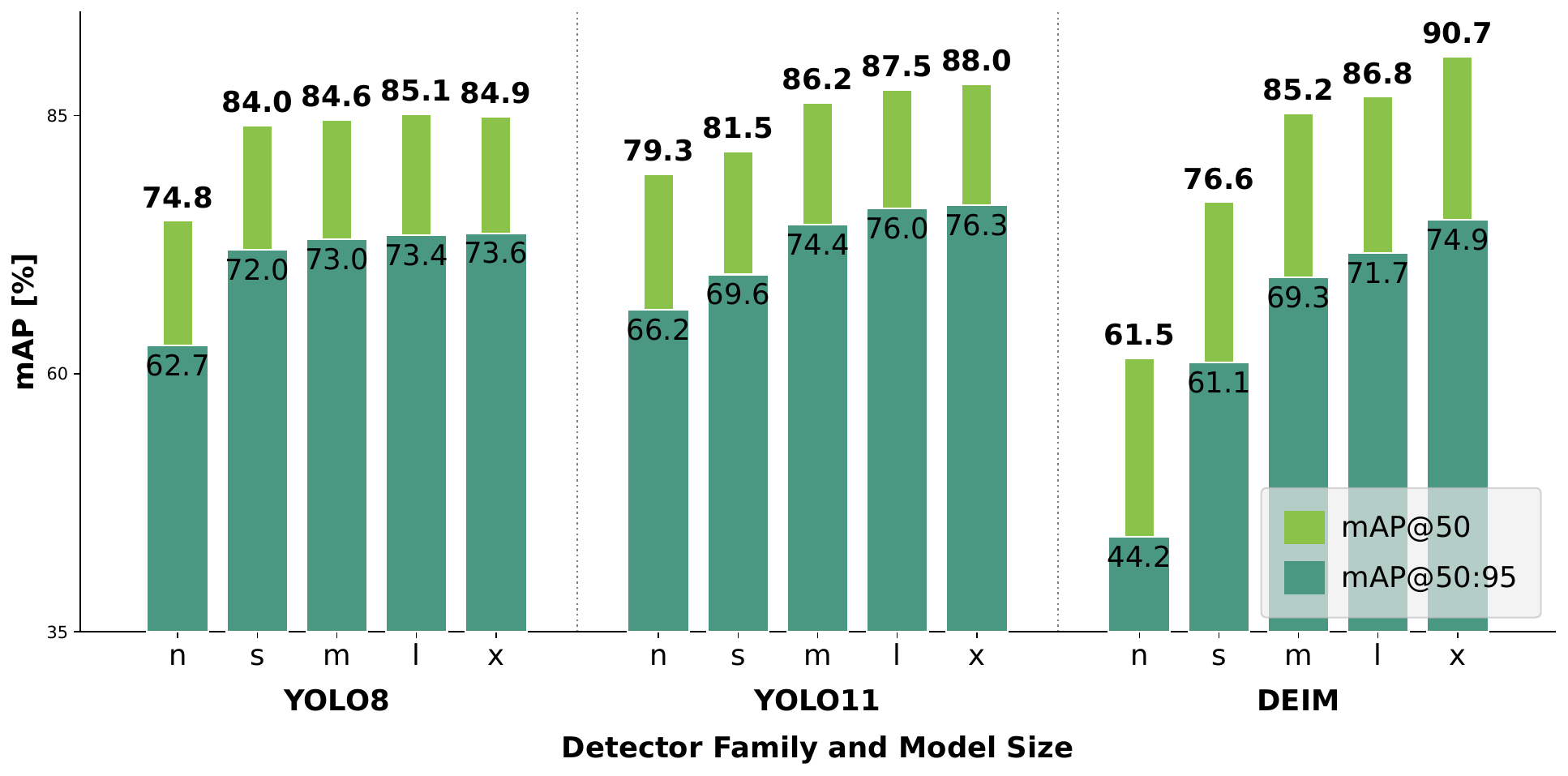}}
    \caption{\Acrshort{sim-to-real} transfer on three SOTA object detection families and model sizes trained on the \acrshort{iasset} synthetic baseline.}
    \label{fig:result-frames-sota}
\end{figure}

As per the baseline configuration described in \autoref{tab:ablation_config}, notable features such as RGB light randomization, exponential light sampling, and physics simulation are deliberately excluded at this stage, as they are systematically studied in the ablation experiments of the subsequent section. While a dependency between model size (\textit{n, s, m, l, x}) and mAP@50 is observed in~\autoref{fig:result-frames-sota} within each family, the convergence of performance across YOLOv8, YOLO11, and DEIM at larger scales suggests comparable \acrshort{sim-to-real} transfer across the studied architectures once sufficient model capacity is available. \added{Therefore, unless otherwise noted, we adopt YOLO11m across all testing scenarios. As shown in ~\autoref{fig:result-frames-sota}, mAP increases substantially up to the medium (m) variant but exhibits diminishing returns for the larger l and x variants (e.g., mAP@50:95 rises by 4.8 points from s to m, but only 1.6 and 0.3 points thereafter). YOLO11m therefore lies near the knee of the accuracy curve, offering near-peak detection performance while avoiding the additional parameters and computational cost inherent to the larger variants. This allows us to isolate the effect of the synthetic data rather than differences in relative model performance.}

\subsubsection{\replaced{SynthRender Computational Cost: Single- and Multi-GPU Execution}{SynthRender Computational Cost}}
\label{subsub:synthrender_computational_cost}

In addition to detection performance, rendering efficiency is relevant for practical synthetic dataset generation. \replaced{We therefore compare the end-to-end computational cost of the SynthRender workflow between a local single-GPU workstation and a cloud-based four-GPU cluster for the automotive, robotics, and \acrshort{iasset} benchmarks.}{We therefore evaluate the end-to-end computational cost of the SynthRender workflow for the automotive, robotics, and \acrshort{iasset} benchmarks.} The measured runtime includes three stages: rendering the synthetic images with the physically-based Cycles engine~\cite{cycles}, generating the corresponding annotations, and training the detector.

\begin{figure}[!t]
    \centering
    \centerline{\includegraphics[width=1\linewidth]{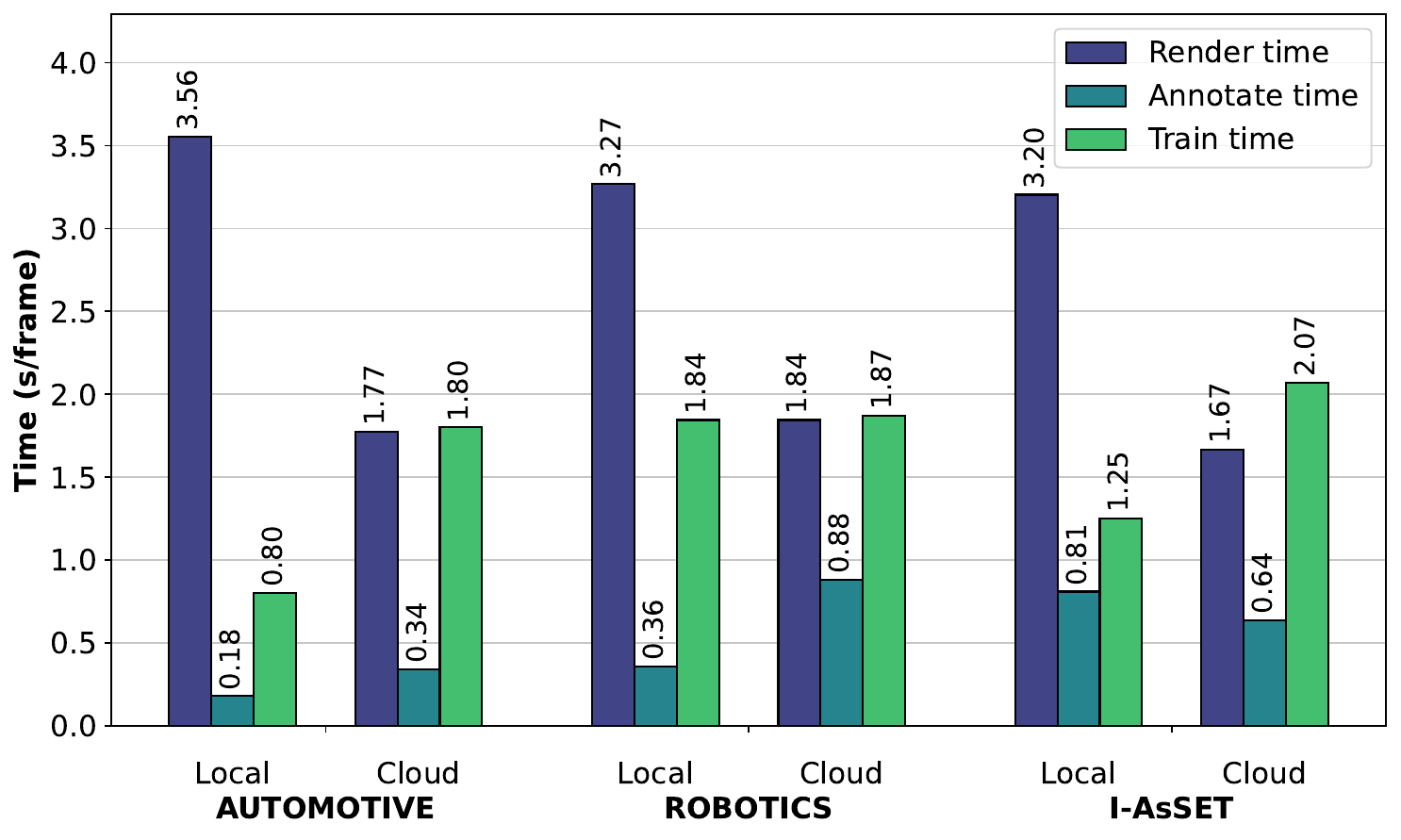}}
    \caption{\label{fig:synthrender_rendering_times}\replaced{End-to-end runtime comparison between single-GPU local execution and four-GPU cloud execution for SynthRender datasets on the automotive, robotics, and \acrshort{iasset} benchmarks. Runtime is divided into rendering, annotation generation, and detector training.}{End-to-end runtime comparison between local and cloud execution for SynthRender datasets on the automotive, robotics, and \acrshort{iasset} benchmarks. Runtime is divided into rendering, annotation generation, and detector training.}}
\end{figure}

Figure~\ref{fig:synthrender_rendering_times} compares \replaced{the single-GPU local setup with the four-GPU cloud setup}{local and cloud execution} for the three datasets at a resolution of 1024$\times$1024. The results show that cloud execution mainly reduces the rendering bottleneck, since SynthRender can distribute the rendering workload across multiple GPU-bound processes by splitting the frame range into independent intervals. This makes cloud execution particularly useful when generating multiple synthetic datasets for ablation studies or when iterating over different randomization configurations. Annotation time is dominated by the number of object instances per image (scene complexity). \added{Because the two setups differ in both GPU model and GPU count, this experiment compares the complete execution configurations rather than providing a hardware-normalized assessment of local versus cloud computing.}

For reproducibility, the local baseline was executed on a single-GPU workstation running Ubuntu 22.04.5 LTS, equipped with an Intel Core i9-14900K CPU, 125.35 GB RAM, a 2 TB SSD, and an NVIDIA GeForce RTX 4090 GPU with 24 GB VRAM. The cloud setup was executed on Azure Databricks using a \texttt{Standard NC64as T4 v3} cluster with 64 CPU cores, 440 GB memory, and four NVIDIA Tesla T4 GPUs. For cloud rendering, SynthRender was launched with four GPU-bound processes, splitting the frame interval into independent sub-ranges rendered in parallel.

However, the speed-up is not uniform across all stages. Annotation generation and training do not benefit from the same degree of parallelism in the evaluated setup and can be affected by data loading, storage, and I/O overhead. Therefore, the main practical advantage of the cloud setup is the reduction of synthetic image rendering time, while the full end-to-end pipeline still depends on the efficiency of the downstream annotation and training stages.

\subsection{Ablation Study on Domain Randomization Parameters}
\label{sub:ablations}

\begin{figure}[hptb]
\centerline{\includegraphics[width=1\linewidth]{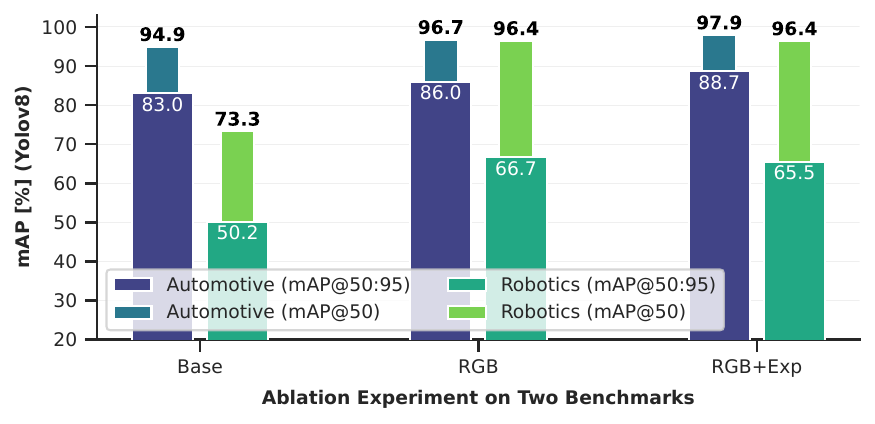}}
\caption{Ablation experiments on the automotive \cite{martinez2024scap} and robotics \cite{horvath2022object} benchmarks.}
\label{fig:DR_Ablation_Automotive_Robotics}
\end{figure}

This section evaluates a subset of the randomization parameters available in SynthRender. All tests carried out in this section follow the additive and reverse ablation configurations of ~\autoref{fig:DR_Ablation_Automotive_Robotics}. 

In the first ablation study, shown in~\autoref{fig:DR_Ablation_Automotive_Robotics}, a positive development of mAP values across the robotics \cite{horvath2022object} and automotive \cite{martinez2024scap} datasets is observed if RGB light color randomization is enabled in addition to the base configuration. Furthermore, if exponential sampling of light intensities within the allowed randomization range is also allowed, \deleted{, as per ~\autoref{sub:synthrender:scene_setup}} the mAP values of the automotive dataset further improve. In contrast, exponential light sampling did not improve performance on the robotics dataset compared to using only RGB light randomization.

\begin{figure*}[t]
\centering
\includegraphics[width=1\textwidth]{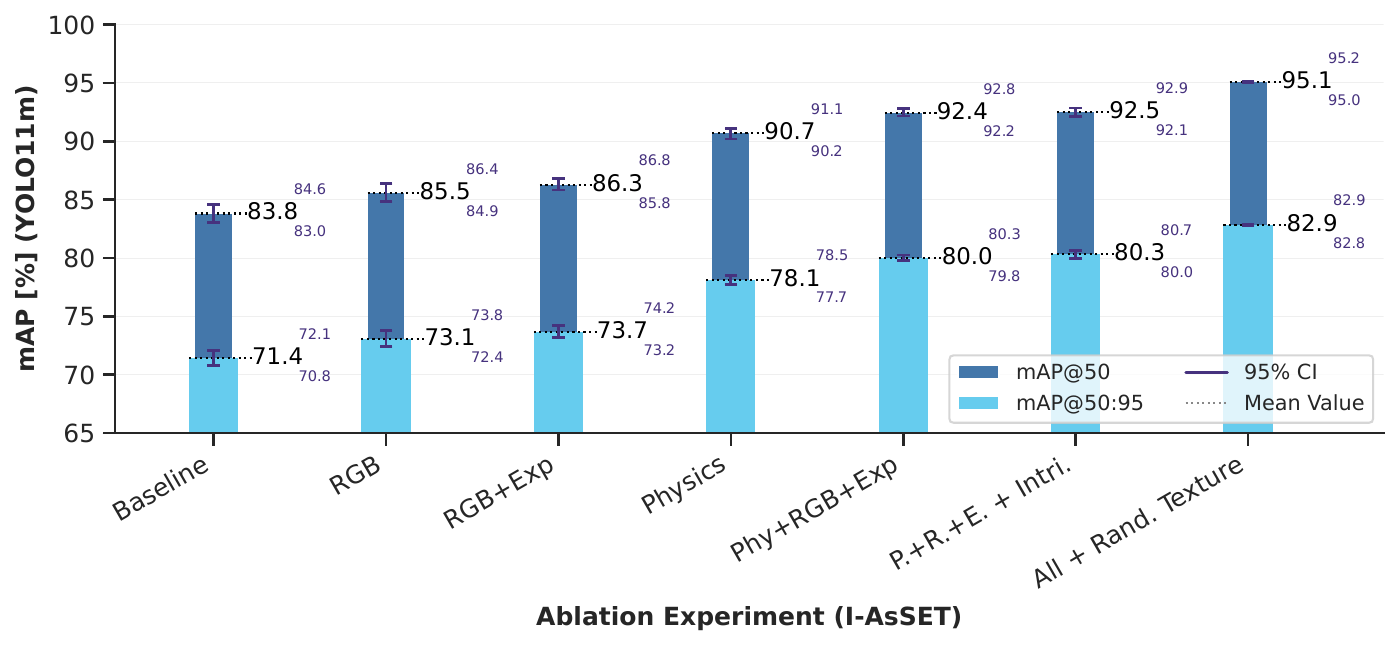}
\caption{Additive ablation study on \acrshort{iasset} using multiple SynthRender's configurations as per \autoref{tab:ablation_config}. Mean mAP values and respective 95\% confidence intervals are presented.}
\label{fig:DR_Ablation}
\end{figure*}

\begin{figure*}[t]
\centering
\includegraphics[width=1\textwidth]{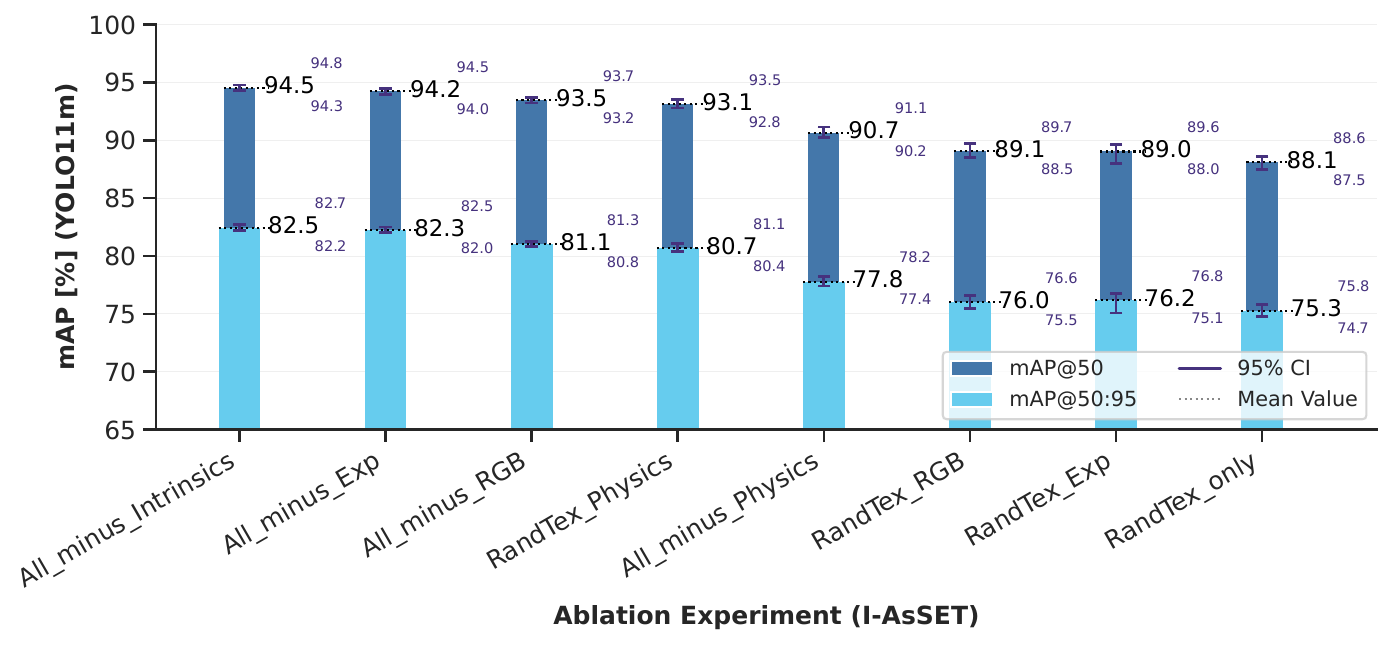}
\caption{Reverse ablation study on \acrshort{iasset} with SynthRender parameters. Mean mAP values and respective 95\% confidence intervals are presented.}
\label{fig:DR_Reverse_Ablation}
\end{figure*}

\autoref{fig:DR_Ablation} and \autoref{fig:DR_Reverse_Ablation} build upon the\added{se initial ablation} results \deleted{presented in \autoref{fig:DR_Ablation_Automotive_Robotics}} by expanding the ablation parameters for the \acrshort{iasset} dataset into an \added{additive and reverse ablation study}\deleted{The experiment names and their association with the SynthRender parameters (~\autoref{tab:synthrender_params}) are summarized in \autoref{tab:ablation_config}}. \added{Both figures show the mAP mean values for each experiment, in addition to 95\% percentile bootstrap confidence intervals from six independent training experiments with variable initialization seed \cite{efron1993bootstrap}.}

The ablation study in \autoref{fig:DR_Ablation} demonstrates that incorporating RGB illumination randomization improves detection performance relative to the baseline. The combination of chromatic light variations and exponential light sampling only yields marginal improvements. Furthermore, implementing physics simulation from the baseline configuration results in a more significant increase in mAP values than the two aforementioned lighting variables combined. Notably, the benefits of RGB randomization and exponential light sampling complement the gains achieved through physics simulation. \replaced{Similarly, using the fixed intrinsic matrix of the Zivid 2+ MR60 acquisition camera in the corresponding ablation configuration results in a further improvement in detection performance.}{In contrast, randomizing camera intrinsics (f-stop) within plausible ranges does not produce significant improvements in detection performance.} Finally, substituting manually-selected textures for the target object with randomized PBR materials further reduces the \acrshort{sim-to-real} gap. This configuration achieves the highest performance on the \acrshort{iasset} dataset, with an mAP@50 of 95.1 and an mAP@50:95 of 82.9.

To complement the progressive additive ablation, we additionally performed a
reverse ablation from the best-performing configuration in \autoref{fig:DR_Reverse_Ablation}. In this setting, one
\acrshort{DR} factor is removed at a time while the remaining configuration is kept unchanged. The reverse ablation supports the trend observed in the forward study: removing physics simulation produces the largest performance degradation, whereas removing RGB illumination variation, exponential light sampling, or camera intrinsics variation leads to smaller reductions. The isolated
RandTex-based variants further indicate that randomized texture alone is insufficient to recover the full performance. Overall, the forward and reverse ablations suggest that physics simulation is the dominant individual contributor, while lighting, intrinsics, and texture randomization
provide complementary gains within the complete \acrshort{DR} pipeline. These effects are therefore interpreted as marginal contributions under the evaluated configurations, rather than as fully independent causal effects. To ensure reproducibility, the synthetic train sets generated via SynthRender to carry out all experiments in \autoref{fig:DR_Ablation}, as well as the checkpoints of the two best-performing models are publicly available within the \acrshort{iasset} dataset. 

\begin{figure}[!t]
\centerline{\includegraphics[width=1\linewidth]{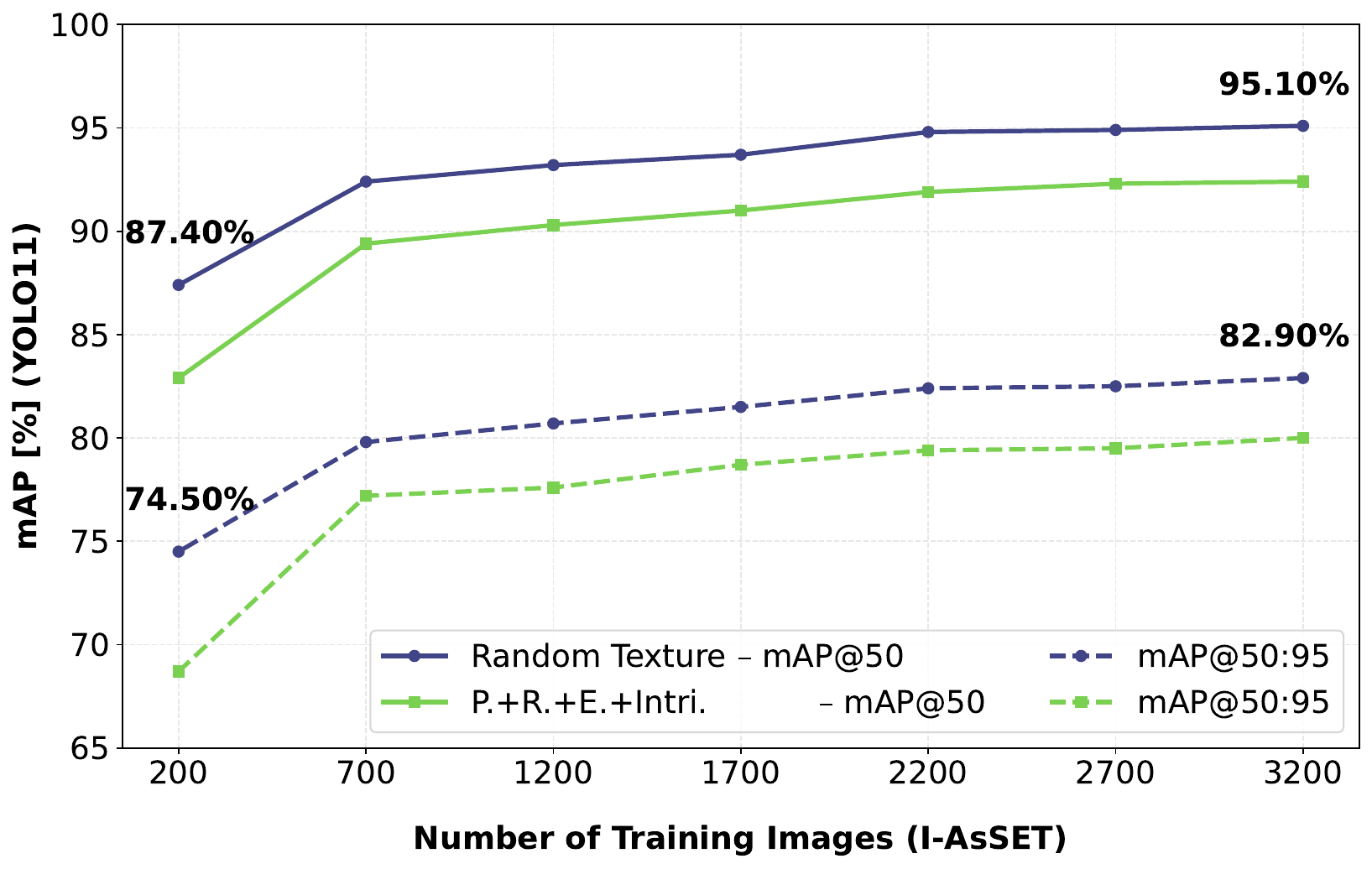}}
\caption{\label{fig:training_size_ablation_comparison}\replaced{Effect of the synthetic training-set size (200--3200 images) on real-world test performance using YOLO11m models for our two best synthetic configurations, \emph{Random Texture} and \emph{Manual Texture}. Solid lines report mAP@50 and dashed lines mAP@50:95, and color distinguishes the two configurations.}{Ablation on \acrshort{iasset} training set size (200–3200 images) using our two best-performing synthetic configurations.}}
\end{figure}

~Figure~\ref{fig:training_size_ablation_comparison} compares the effect of progressively increasing the number of synthetic training images for our two best-performing synthetic datasets from~\autoref{fig:DR_Ablation}. For both datasets, mAP@50 and mAP@50:95 increase in proportion to the amount of images in the train set. However, performance gains become smaller as more data samples are added
\deleted{ Major performance gains are attained at low thousands image regime, with both datasets retaining competitive accuracy in small train sets, demonstrating that high-fidelity synthetic assets can be effective even in low-data regimes.}
\added{exhibiting diminishing returns as the training set grows: within the evaluated range, the marginal gain generally decreases, and the final increment contributes less than 0.5~mAP@50 (the Random Texture dataset rises only from 94.9 to 95.1~mAP@50 between the two largest training-set sizes) (Figure~\ref{fig:training_size_ablation_comparison})}.
\added{This behaviour is consistent with prior work on the same robotics benchmark, in which doubling the synthetic set from 4{,}000 to 8{,}000 images changes performance by only $+0.3$~mAP@50 (YOLOv8; $96.1\rightarrow96.4$)~\cite{Zhu2025icra}, indicating that accuracy is governed more by the construction of synthetic variability than by dataset volume within this low-thousands-image regime.} This setting therefore provides the best balance
between coverage and \acrshort{sim-to-real} transfer.

\subsection{Few-Shot Finetuning on Fully-Synthetic Models}
\label{sub:few_shot}

\autoref{tab:fewshot_results} summarizes the effect of adding a small number of real samples to synthetic training data across our two best synthetic datasets. Both datasets show consistent improvement with additional real images, demonstrating the value of few-shot adaptation. For the \textit{Random Texture} dataset, performance improves from 95.1 mAP@50 (0-shot) to 98.8 mAP@50 (10-shot), with similar gains in mAP@50:95. The \textit{Manual Texture} dataset shows even more pronounced improvement, gaining 9.8\% mAP@50:95 with 10 real samples (80.3\% to 90.1\%).

These results confirm that even a single real image provides measurable improvement over pure synthetic training, and 5 real images are sufficient to bridge most of the \acrshort{sim-to-real} gap, achieving over 98\% mAP@50 on both datasets.

The aggregate improvement observed with few-shot fine-tuning is consistent across most classes; however, performance gains are expected to be non-uniform, with visually ambiguous and highly reflective classes such as \path{C_Steel_Ball} and \path{GF_Knurled_Screw_M8}, identified as low performers in~\autoref{fig:per_class_metrics}, likely driving the residual gap between synthetic-only and few-shot results.

\begin{table}[htbp]
\centering
\caption{Few-shot results on \acrshort{iasset} with 4000 synthetic images. Zero-shot uses only synthetic data; few-shot setups add 1–10 real samples to the training dataset.}
\label{tab:fewshot_results}
\renewcommand{\arraystretch}{1.2} 
\resizebox{\columnwidth}{!}{%
\begin{tabular}{c c c c c c c}
\hline
\multirow[c]{2}{*}{\textbf{Dataset}} & \multirow[c]{2}{*}{\textbf{mAP}} & \multicolumn{5}{c}{\textbf{Number of real images}} \\
\cline{3-7}
                  &                & \textbf{0} & \textbf{1} & \textbf{3} & \textbf{5} & \textbf{10} \\
\hline
\multirow{2}{*}{Random Texture} & @50     & 95.1 & 97.1 & 97.9 & 98.5 & \textbf{98.8} \\
                                 & @50:95  & 82.9 & 85.9 & 88.1 & 89.3 & \textbf{89.7} \\
\hline
\multirow{2}{*}{Manual Texture}  & @50     & 92.5 & 95.9 & 97.5 & 98.3 & \textbf{98.7} \\
                               & @50:95  & 80.3 & 84.9 & 87.5 & 89.2 & \textbf{90.1} \\
\hline
\end{tabular}
}
\end{table}

\added{Because both training and validation are performed on synthetic data, model convergence relies on the synthetic validation set, which raises the question of whether synthetic-validation accuracy is a reliable proxy for real-domain accuracy. We observe that it is not a discriminative one. Evaluated on their own synthetic validation splits, the two best configurations are near saturation; i.e., \emph{Random Texture} reaches $99.3$ mAP@50 ($97.2$ mAP@50:95) and \emph{Manual Texture} $99.3$ mAP@50 ($96.3$ mAP@50:95), yet their real-domain performance differs appreciably, with $95.1$ and $92.5$ mAP@50 respectively; as shown in \autoref{tab:fewshot_results}). This indicates an associated \acrshort{sim-to-real} gap of about $4$--$7$ mAP@50 and $14$--$16$ mAP@50:95 for our two best models, which is precisely what the real-world \acrshort{iasset} test set is designed to expose. In contrast, adding only one to five real images reduces this gap (\autoref{tab:fewshot_results}), mitigating the synthetic-domain overfitting.}

\subsection{3D Reconstruction as Domain Adaptation}
\label{sub:DA_results}

\autoref{tab:domain_adaptation_final} compares the results of the reconstruction methods described in ~\autoref{sec:techniques_for_DA}, being used as automated domain-adaptation strategies, against randomized and manual approaches. \deleted{This experiment reveals that manually modeled CAD geometries, whether
with manual or randomized textures, provide the strongest performance.}

\added{This experiment reveals that the strongest performance is obtained with manually modeled CAD geometry and randomized textures (95.1 mAP@50, 82.9 mAP@50:95). A notable result concerns 3DGS reconstruction: using the natively scanned 3DGS texture yields 86.6 mAP@50 (75.2 mAP@50:95), whereas replacing it with randomized PBR materials on the same reconstructed geometry raises performance to 92.8 mAP@50 (80.6 mAP@50:95), a $+6.2$ mAP@50 improvement. This randomized-texture 3DGS configuration statistically matches the fully manual CAD + PBR configuration (92.5 mAP@50, 80.3 mAP@50:95; overlapping 95\% confidence intervals) while requiring neither manual geometry modeling nor manual texture authoring. Automated geometry reconstruction combined with texture randomization therefore reaches parity with fully manual modeling and hand-crafted PBR texturing, trailing only the best manual configuration (CAD + randomized texture) by 2.3 mAP@50 and outperforming the single-image GenAI methods by roughly 6--7 mAP@50. This positions 3DGS with texture randomization as a fully automated alternative that reaches manual-modeling parity when accurate CAD models are unavailable. The gap between native-texture and randomized-texture 3DGS is consistent with the way 3DGS bakes capture-time illumination and cast shadows into the recovered surface appearance: using the scanned texture directly transfers these scan-specific lighting and reconstruction artifacts into the synthetic training distribution, whereas randomizing the texture removes this
dependence and forces the detector toward transferable geometric cues.}


\begin{table*}[htbp]
\centering
\caption{Comparison of \acrshort{real-to-sim} \acrshort{DA} and randomization strategies. Values reported as mean$^{\text{95\% CI upper}}_{\text{95\% CI lower}}$ over $n=6$ independent runs (percentile bootstrap).}
\label{tab:domain_adaptation_final}
\renewcommand{\arraystretch}{1.2}
\setlength{\tabcolsep}{4pt}
\begin{tabular}{m{2cm} ll c cc cc}
\hline
\multirow{2}{*}{\textbf{Background}}
 & \multicolumn{2}{c}{\textbf{Target Objects}} &  & \multicolumn{2}{c}{\textbf{mAP@}} & \multicolumn{2}{c}{\textbf{Detection Metrics}} \\
\cline{2-3}\cline{5-6}\cline{7-8}
 & \textbf{Geometry} & \textbf{Texture} &  & \textbf{50} & \textbf{50:95} & \textbf{Precision} & \textbf{Recall} \\
\hline
\multirow{7}{*}{\textbf{Randomized}}
  & Manual CAD          & Randomized     &  & $\mathbf{95.1}^{95.2}_{95.0}$ & $\mathbf{82.9}^{82.9}_{82.8}$ & $92.6^{92.8}_{92.3}$ & $91.6^{91.9}_{91.3}$ \\
  & Manual CAD          & Manual PBR     &  & $92.5^{92.9}_{92.1}$ & $80.3^{80.7}_{80.0}$ & $92.1^{92.4}_{91.9}$ & $87.2^{87.6}_{86.8}$ \\
  & 3DGS Reconstr.      & 3DGS Reconstr. &  & $86.6^{86.9}_{86.2}$ & $75.2^{75.4}_{74.8}$ & $85.0^{85.6}_{84.4}$ & $83.1^{83.4}_{82.7}$ \\
  
  & 3DGS Reconstr.      & Randomized. &  & $92.8^{93.2}_{92.4}$ & $80.6^{80.9}_{80.3}$ & $90.6^{90.9}_{90.3}$ & $87.9^{88.2}_{87.6}$ \\
  & Manual CAD          & MeshyAI GenAI  &  & $85.9^{86.1}_{85.7}$ & $74.6^{74.8}_{74.4}$ & $83.7^{84.2}_{83.2}$ & $78.9^{79.3}_{78.3}$ \\
  & TRELLIS GenAI       & TRELLIS GenAI  &  & $86.8^{87.2}_{86.3}$ & $74.8^{75.2}_{74.4}$ & $83.7^{85.2}_{82.8}$ & $83.5^{84.4}_{82.9}$ \\
\hline
\textbf{Reconstructed Scene (3DGS)}
  & Manual CAD          & Manual PBR      &  & $92.1^{92.5}_{91.7}$ & $79.5^{80.1}_{79.1}$ & $92.7^{93.2}_{92.2}$ & $86.8^{87.1}_{86.5}$ \\
\hline
\end{tabular}
\end{table*}

The lower-overhead, GenAI-based MeshyAI and TRELLIS, which can rely on a single image for geometry and texture reconstruction, perform slightly lower. In our empirical experiments with TRELLIS, geometry and texture consistency were found to depend strongly on the input image perspective, yielding lower reliability than the other approaches. However, results for GenAI-based reconstruction methods remain \deleted{above 86\%}\added{around 86\%} mAP@50, confirming that automated 2D-to-3D asset creation constitutes a valid alternative to manual modeling when CAD models are unavailable.

Furthermore, replacing randomized backgrounds with a more realistic scene setup using 3DGS scans of the real environment yields nearly identical performance. This aligns with previous research suggesting that background adaptation is less critical for the detection model to transfer effectively to real-world scenarios \cite{araya-martinez2025xai} and positions \acrshort{3DGS} as an alternative to CAD-based scene and asset creation.

\added{For practical industrial deployment, input imagery can be acquired in-house directly from the physical components: a high-resolution single view can support appearance generation or single-view GenAI reconstruction, whereas 3DGS benefits from overlapping turntable or camera-orbit views covering the object. These captures are used only to reconstruct a reusable 3D asset and require no object-detection annotation; the annotated detector-training images are subsequently generated synthetically by SynthRender. Supplier images may provide an alternative when physical access is unavailable, subject to usage rights and sufficient viewpoint coverage. Because single- or few-view GenAI methods infer unobserved geometry, assets intended for geometry-critical tasks should be validated against physical measurements or CAD tolerances, with calibrated 3D scanning or CAD remaining preferable when dimensional accuracy is required.}

To foster future \acrshort{real-to-sim} research, we release the synthetic datasets evaluated in \autoref{tab:domain_adaptation_final} \added{as part of I-AsSET}. Furthermore, we provide 3D assets for the 32 \acrshort{iasset} object classes generated via all investigated reconstruction methods, i.e. \acrshort{3DGS}, TRELLIS, and MeshyAI. To support the development of novel reconstruction techniques, we also supply the real multi-view 2D images used to generate the 3D meshes.

\subsection{Benchmarking Against the State-of-the-Art}
\label{sub:sota_benchmark}

\added{We compare our approach against state-of-the-art methods across three benchmark datasets under matched experimental budgets, model architectures, and image resolutions. For this comparison, we employ the same hyperparameters used for the ablation studies, as previously stated in \autoref{tab:hyperparams}, unless otherwise specified. All state-of-the-art comparisons use identical, off-the-shelf training conditions and report performance as {mAP}@50, {mAP}@50:95, recall and precision, following the PASCAL~VOC~\cite{everingham2010pascal} and COCO~\cite{lin2014microsoft} conventions. As with all other experiments in this work, each configuration is trained six times and the reported values include mean and 95\% bootstrap confidence intervals \cite{efron1993bootstrap}. Evaluation is performed exclusively on the corresponding real-world test set. Entries marked ``--'' denote metrics not reported by the original authors.}

\autoref{tab:Ours_vs_SOTA} shows that, \added{under a matched training budget
(YOLOv8, 4{,}000 synthetic images, $720\times720$),} the best-performing
configuration, using randomized \acrshort{PBR} materials applied to
\acrshort{CAD} models, \added{reaches 98.7~{mAP}@50 with a narrow 95\% confidence interval of 0.2 on the robotics
dataset~\cite{horvath2022object}, compared to an 96.1~{mAP}@50 for a prior
configuration~\cite{Zhu2025icra} under otherwise identical conditions, a
$+2.6$~{mAP}@50 improvement.} \deleted{Notably, this improvement is achieved
using only 4{,}000 synthetic training images, whereas the compared prior result
uses 8{,}000 images.} For both datasets, identical YOLOv8 training settings
were used as per \autoref{tab:hyperparams} \added{(500 epochs, batch 16, lr 0.01, patience of 30)}.
\added{To foster results reproducibility, we provide the synthetic robotics train set leading to this result in a
\href{https://huggingface.co/datasets/moiaraya/SynthRender_Robotics/}{robotics
repository}~\cite{araya2026synthrender_robotics}.}

\begin{table*}[ht]
\centering
\caption{State-of-the-art comparison on fully-synthetic training of robotics~\cite{horvath2022object}, automotive~\cite{martinez2024scap} and proposed \acrshort{iasset} benchmarks. \added{Automotive and Robotics experiments use 512x512 and 720×720 respectively to match the resolution of the compared prior work}. Values for \textit{Ours} are reported as mean$^{\text{95\% CI upper}}_{\text{95\% CI lower}}$ over $n=6$ independent runs.}
\label{tab:Ours_vs_SOTA}
\renewcommand{\arraystretch}{1.2}
\setlength{\tabcolsep}{4pt}
\begin{tabular}{l l cc cc c cccc}
\hline
 &  & \multicolumn{2}{c}{\textbf{mAP@}} & \multicolumn{2}{c}{\textbf{Detection Metrics}} & & \multicolumn{4}{c}{\textbf{Train Conditions}} \\
\cline{3-4} \cline{5-6} \cline{8-11}
\textbf{Dataset} &
\textbf{Method} &
\textbf{50} &
\textbf{50:95} &
\textbf{Prec.} &
\textbf{Rec.} &
&
\textbf{Model} &
\textbf{\# Img} &
\textbf{Res.} &
\textbf{Texture} \\
\hline
\multirow{2}{*}{Robotics~\cite{horvath2022object}}
 & Zhu~\cite{Zhu2025icra} & 96.1 & -- & -- & -- & & YOLOv8 & 4k & $720^2$ & Rand. PBR \\
 & Ours & $\mathbf{98.7}^{98.9}_{98.5}$ & $\mathbf{68.2}^{68.4}_{68.0}$ & $96.1^{97.4}_{94.7}$ & $97.3^{97.7}_{96.9}$ & & YOLOv8 & 4k & $720^2$ & Rand. PBR \\
\hline
\multirow{4}{*}{Automotive~\cite{martinez2024scap}}
 & Araya-Martinez~\cite{martinez2024scap} & -- & 75.0 & -- & -- & & YOLOv8 & 900 & $512^2$ & Manual \\
 & Araya-Martinez~\cite{araya-martinez2025xai} & 91.3 & 78.4 & -- & -- & & YOLOv8 & 900 & $512^2$ & Manual \\
 & Ours & $\mathbf{95.5}^{95.8}_{95.1}$ & $\mathbf{81.0}^{82.7}_{79.5}$ & $93.3^{95.4}_{91.6}$ & $92.2^{93.9}_{91.0}$ & & YOLOv8 & 900 & $512^2$ & Manual \\
 \cmidrule(lr){3-11}
 & Ours & $\mathbf{97.9}^{98.4}_{97.4}$ & $\mathbf{88.7}^{89.2}_{88.1}$ & $97.7^{98.1}_{97.3}$ & $95.3^{96.1}_{94.7}$ & & YOLOv8 & 4k & $512^2$ & Manual \\
\hline
\multirow{1}{*}{\acrshort{iasset}}
 & Ours & $95.1^{95.2}_{95.0}$ & $82.9^{82.9}_{82.8}$ & $92.6^{92.8}_{92.3}$ & $91.6^{91.9}_{91.3}$ & & YOLO11 & 4k & $1024^2$ & Rand. PBR \\
\hline
\end{tabular}
\end{table*}

The observed performance gains are not attributable to a single factor but rather to a combination of \acrshort{DR} parameters identified through the ablation studies presented in this work. Each individual parameter, RGB light color randomization, exponential light sampling, physics-based object placement, camera intrinsics variation, and randomized \acrshort{PBR} materials, contributes incrementally, as shown in the ablation results. Their combined effect yields a set of \acrshort{DR} features that collectively reduce the sim-to-real gap, as validated by the progressive mAP improvements reported in the additive ablation study (\autoref{fig:DR_Ablation}) and the progressive mAP degradation shown in the reverse ablation study (\autoref{fig:DR_Reverse_Ablation}). Light randomization captures the variation in illumination characteristic of industrial environments, while camera-to-object distance parameters are tuned to match the geometric conditions of the target setup. Together, these choices align the synthetic training distribution more closely with the \deleted{real} test distribution, rather than relying on broad, unconstrained randomization that may introduce out-of-context samples and dilute the feature coverage relevant to the target domain. This targeted, domain-aware generation strategy is considered the primary driver of the consistent improvements reported across all three benchmarks.

Consistent improvements are also observed on the automotive benchmark \cite{martinez2024scap} under identical test condition. Here, manually assigned textures are used instead of randomized PBR materials to match previously reported settings. 

\subsection{Inter-Class Ambiguity and Failure Modes}
\label{sub:failure_analysis}

\acrshort{iasset} was specifically designed to represent inter-class ambiguity, i.e., several object classes share almost the exact same shape and materials and differ primarily in size, such as the \path{C_O_Ring}, \path{GF_Slotted_Pin}, and \path{C_Steel_Ball} variants. These classes were intentionally included to reflect a realistic industrial setting and to test the limits of standard object detectors. Despite minor cross-class confusions within these families, the model demonstrates a robust overall ability to separate classes that differ purely by scale. This success is directly attributed to the synthetic data generation pipeline. By simulating physics to let objects rest naturally on a shared plane, and by restricting the camera-to-object distance and spatial parameters to realistic ranges defined by the target acquisition setup, the synthetic data provides highly consistent size cues. This prevents unrealistic perspective distortions, such as a small object appearing as large as a different class simply because the camera was placed unusually close. \deleted{Consequently, the network learns to reliably utilize apparent object scale as a discriminative feature.}

\added{To confirm that this scale separation is learned from object features rather than inferred from scene context or from the relative size of neighboring parts, we evaluated the \texttt{Random Texture} model on the 96 single-object test images, as introduced in \autoref{tab:iasset_scenes}, in which each part is imaged in isolation so that no co-occurring object is available as a size reference. Following the PASCAL VOC evaluation protocol, detections were considered true positives when matched to a ground-truth object with an $\text{IoU}\ge0.5$ \cite{everingham2010pascal}. During inference, predictions with confidence scores below $0.25$ were discarded, following the default setting of the Ultralytics YOLO implementation \cite{Jocher_Ultralytics_YOLO_2023}.}

\begin{figure*}[htbp]
    \centering
    \includegraphics[width=0.98\linewidth]{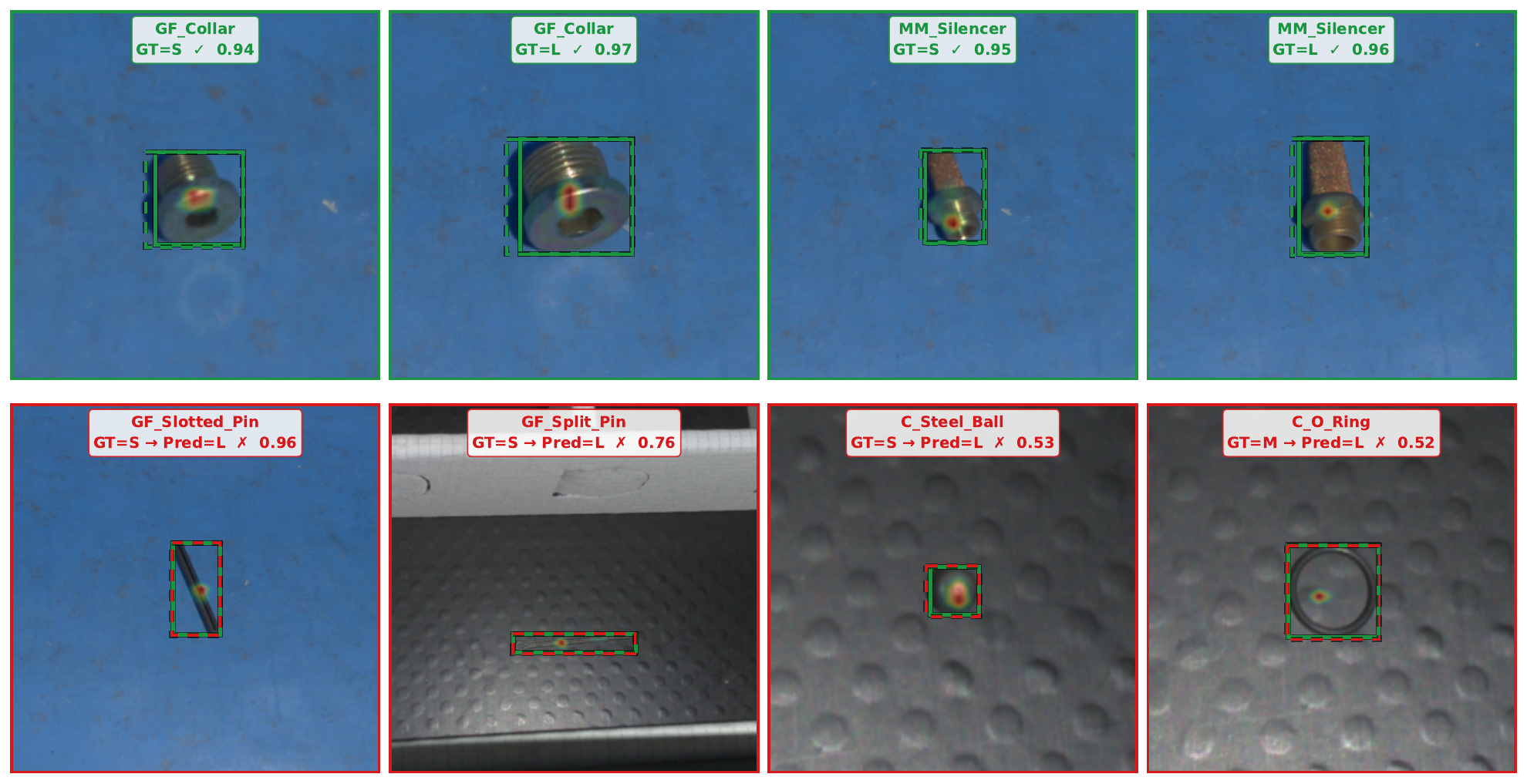}
    \caption{\added{Scale discrimination on the single-object test images, overlaid with LayerCAM evidence maps~\cite{jiang2021layercam} for the predicted class. Warm colors mark the image regions supporting the prediction. The ground-truth box is solid green and the predicted box is dashed, green where the class is correct and red where the size is mis-predicted. Top row: parts from the same size-ambiguous family imaged at different sizes, all correctly classified, showing that the detector recovers absolute scale without any relative-size reference. Bottom row: the residual within-family size confusions; the ground-truth and predicted boxes coincide, so localization is correct and only the size class flips. In every panel the evidence is confined to the part and concentrates on small local structures rather than on the surrounding background.}}
    \label{fig:scale_single_object}
\end{figure*}

\added{Even without any relative-size cue, the model predicts the correct class for $88.5\%$ of these images ($85/96$), and for $80.4\%$ ($41/51$) of the single-object images depicting the size-ambiguous families (\texttt{C\_\allowbreak Steel\_\allowbreak Ball}, \texttt{GF\_\allowbreak Split\_\allowbreak Pin}, \texttt{GF\_\allowbreak Collar}, \texttt{C\_\allowbreak O\_\allowbreak Ring}, \texttt{C\_\allowbreak Plastic\_\allowbreak Washer}, \texttt{MM\_\allowbreak Silencer}, \texttt{GF\_\allowbreak Slotted\_\allowbreak Pin}, and \texttt{C\_\allowbreak Washer}), well above the $\sim\!50\%$ chance level of the binary size families. Same-family variants imaged at different sizes are frequently both classified correctly (e.g., \texttt{GF\_\allowbreak Collar} and \texttt{MM\_\allowbreak Silencer}) as seen in Figure~\ref{fig:scale_single_object}, top row. The residual errors are almost exclusively within-family size confusions that systematically over-predict size (e.g., \texttt{GF\_\allowbreak Slotted\_\allowbreak Pin\_\allowbreak S} classified as \texttt{GF\_\allowbreak Slotted\_\allowbreak Pin\_\allowbreak L} at $0.96$ confidence) as depicted in Figure~\ref{fig:scale_single_object}, bottom row, and are concentrated on object classes lacking distinctive geometric features such as threads, internal bores, or junctions.}

\added{The evidence maps overlaid in Figure~\ref{fig:scale_single_object} make explicit which image regions drive these decisions. They are LayerCAM maps~\cite{jiang2021layercam} computed by taking gradients with respect to the score of the predicted class. For the correctly-predicted classes, the class evidence is confined to the part itself and, within it, to small local structures. The synthetic training images are rendered with randomized background textures, so the surfaces and scene composition of the real acquisition environment are not represented in the training distribution and could not act as a transferable cue in the first place. The maps agree with this: the supporting evidence remains on the part in every panel, in line with the earlier observation that background adaptation is not critical for transfer~\cite{araya-martinez2025xai}. Taken together, these observations indicate that the model is not exploiting contextual cues, relative object sizes, environmental priors, or scene textures to differentiate between ambiguous classes.}

\begin{figure}[thpb]
  \centering
  \includegraphics[width=1\linewidth]{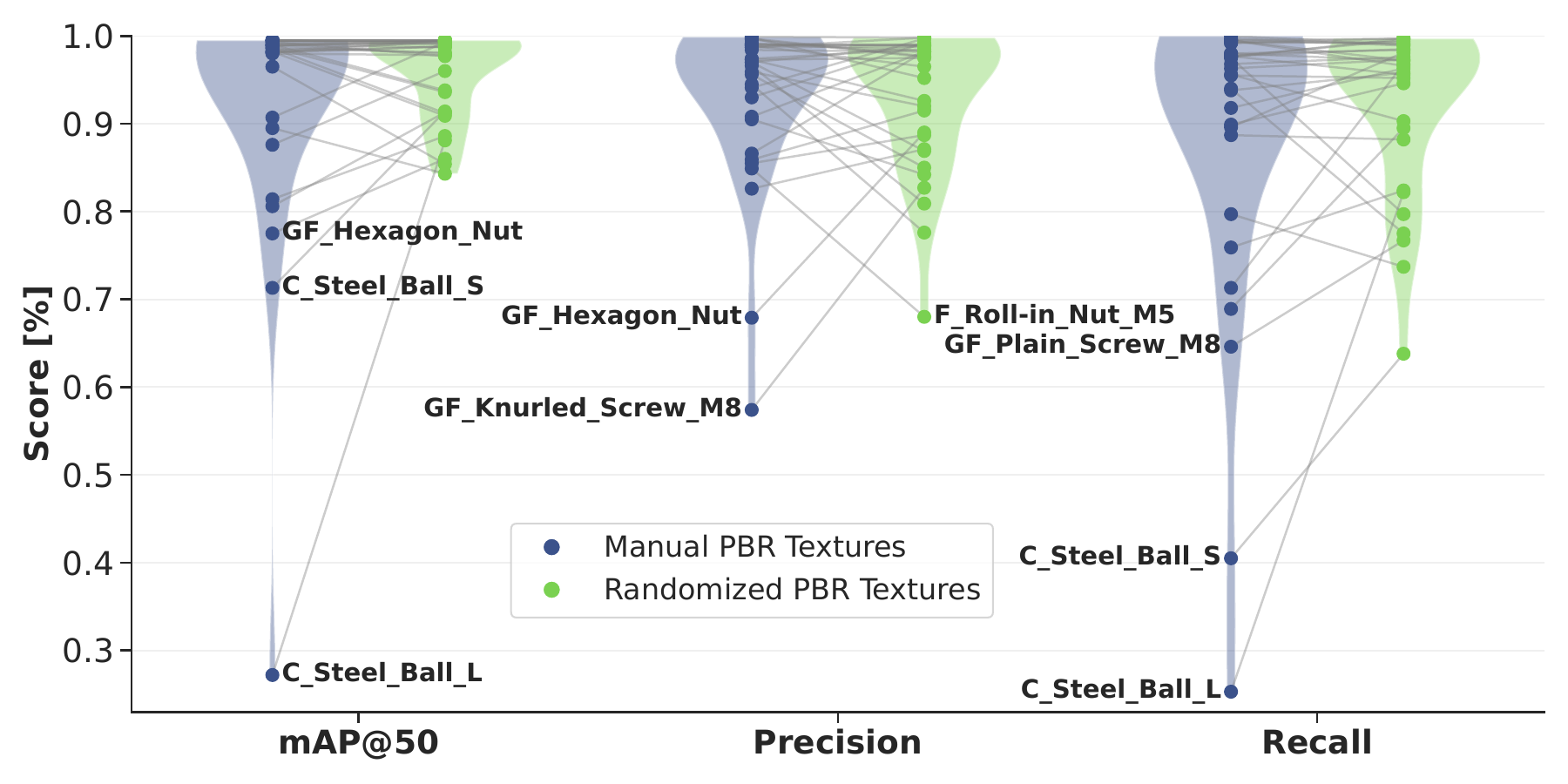}
  \caption{Per-class performance comparison of \acrshort{iasset} objects after training with manual and randomized textures.}
  \label{fig:per_class_metrics}
\end{figure}

\autoref{fig:per_class_metrics} compares the distributions of mAP @50, precision, and recall for the two top-performing datasets from \autoref{fig:DR_Ablation}, which also correspond to the two right-most columns of \autoref{tab:ablation_config}. These two datasets provide full parameter randomization, with the only difference being the manual or randomized texture assignment for the target objects. In the comparative graph, the lowest-performing \acrshort{iasset} classes are labeled in each metric. Precision distributions exhibit similar patterns across both datasets; however, mAP@50 and recall curves show more outliers and lower mean values in the manual dataset, particularly for the C\_Steel\_Ball\_X classes. These results suggest that randomized textures are advantageous for highly reflective surfaces, as the detection model is forced to rely on geometric cues, which remain more consistent across synthetic and real environments.

\begin{figure*}[!t]
    \centering
    \includegraphics[width=\linewidth]{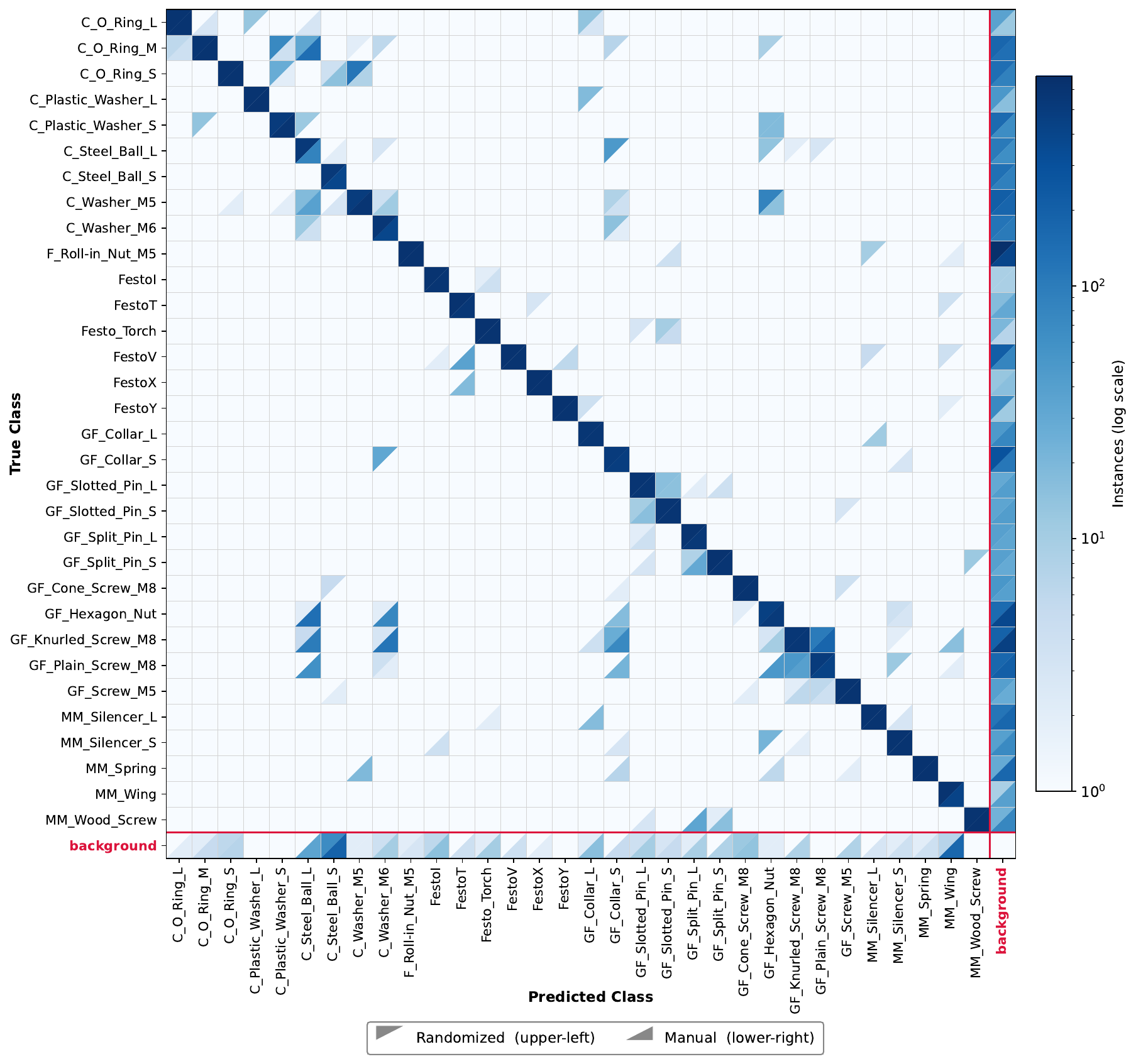}
    \caption{\label{fig:confusion_matrix_diff}\replaced{Per-class confusion matrix on the \acrshort{iasset} test set comparing the two texture-assignment strategies. Each cell is split diagonally: the upper-left triangle is the randomized-texture model and the lower-right triangle is the manually-textured model, shaded by instance count on a shared logarithmic scale to keep sparse (rare) confusions visible. The final \emph{background} row and column report, per model, the missed detections and the false detections; their counts are printed in each background cell (randomized top-left, manual bottom-right). The pronounced diagonal reflects the high accuracy of both models.}{Outlier aware per-class misclassification comparison between randomized and manual textures on the \acrshort{iasset} test set. Off diagonal cells show a z-score outlier criterion ($\tau = 2.0$): green indicates manual exceeds the threshold, yellow indicates randomized exceeds, purple means both exceed, and white means neither. The diagonal shows average normalized per class recall.}}
\end{figure*}

\replaced{The per-class breakdown in Figure~\mbox{\ref{fig:per_class_metrics}} and the dual confusion matrix in Figure~\mbox{\ref{fig:confusion_matrix_diff}}, in which every cell is split diagonally to display the randomized-texture model (upper-left triangle) alongside the manually-textured model (lower-right triangle), reveal a systematic shift in the error distribution between the two texture-assignment strategies rather than a uniform improvement. Comparing the two triangles of each off-diagonal cell isolates the class pairs on which the strategies disagree, while the added \mbox{\emph{background}} row and column report, for each model, the missed detections (ground-truth objects not detected) and the false detections (predictions with no matching object).}{The per-class breakdown in~\mbox{\autoref{fig:per_class_metrics}} and the outlier-aware pairwise difference map in~\mbox{Figure~\ref{fig:confusion_matrix_diff}} reveal a systematic shift in the error distribution between the two texture assignment strategies rather than a uniform improvement. To identify cells where the two models diverge beyond noise, each off-diagonal entry of the row-normalized confusion matrix is flagged as an outlier if it exceeds two standard deviations above the mean of all off-diagonal entries pooled across both models~\mbox{\cite{iglewicz1993detect, barnett1994outliers}}. Despite the visual salience of the color encoding, the flagged cells represent only 2.8\% of all matrix entries, underscoring that the colored cells reflect statistically rare but systematic failure modes rather than pervasive confusion.}

\replaced{Where the manually-textured model produces a darker lower-right triangle, i.e. more confusions than the randomized model, the errors are concentrated among intra-material class pairs such as \texttt{GF\_\allowbreak Knurled\_\allowbreak Screw\_\allowbreak M8} vs. \texttt{GF\_\allowbreak Plain\_\allowbreak Screw\_\allowbreak M8} and the \texttt{C\_\allowbreak Steel\_\allowbreak Ball} variants, consistent with overfitting to simulated specular reflections that do not transfer to real imagery; the \emph{background} row corroborates this, as the manual model misses far more \texttt{C\_\allowbreak Steel\_\allowbreak Ball\_\allowbreak S} instances than the randomized one. The randomized model resolves these confusions by forcing the network to rely on geometric edge profiles. Conversely, where the randomized model's upper-left triangle is darker, the added confusions are inter-material, such as \texttt{C\_\allowbreak O\_\allowbreak Ring} variants being misclassified as \texttt{C\_\allowbreak Washer} or \texttt{C\_\allowbreak Plastic\_\allowbreak Washer} instances when their 2D silhouettes briefly align. Cells in which both triangles are dark correspond to confusions that neither strategy resolves, arising from a superposition of both limitations: classes that simultaneously share geometric structure, similar scale, and ambiguous surface appearance; most notably the \texttt{GF\_\allowbreak Screw\_\allowbreak M8} pair and confusions between \texttt{C\_\allowbreak Steel\_\allowbreak Ball\_\allowbreak L} and other circular objects.}{The green cells in~\mbox{Figure~\ref{fig:confusion_matrix_diff}}, where only the manually textured model exceeds the misclassification threshold, are concentrated among intra-material class pairs such as \texttt{GF\_\allowbreak Knurled\_\allowbreak Screw\_\allowbreak M8} vs.\ \texttt{GF\_\allowbreak Plain\_\allowbreak Screw\_\allowbreak M8} and the \texttt{C\_\allowbreak Steel\_\allowbreak Ball} variants, consistent with overfitting to simulated specular reflections that do not transfer to real imagery. The randomized model resolves these confusions by forcing the network to rely on geometric edge profiles. However, the yellow cells reveal the complementary cost: removing material appearance as a discriminative feature introduces inter-material confusions that were previously absent, such as \texttt{C\_\allowbreak O\_\allowbreak Ring} variants being misclassified as \texttt{C\_\allowbreak Washer} or \texttt{C\_\allowbreak Plastic\_\allowbreak Washer} instances when their 2D silhouettes briefly align. Purple cells indicate confusions that neither strategy resolves, arising from a superposition of both limitations: classes that simultaneously share geometric structure, similar scale, and ambiguous surface appearance; most notably the \texttt{GF\_\allowbreak Screw\_\allowbreak M8} pair and confusions between \texttt{C\_\allowbreak Steel\_\allowbreak Ball\_\allowbreak L} and other circular objects.} Taken together, \acrshort{DR} imposes a fundamental trade-off: intra-material confusions driven by appearance overfitting are suppressed at the cost of a moderate increase in inter-material confusions driven by geometric similarity, while scale-based ambiguities remain as a strategy-independent error floor. 

\added{These residual within-family scale confusions constitute a
recognized limitation of the current approach: for classes that are
geometrically near-identical and differ only in absolute size, detection remains sensitive to the size cues encoded by the physics-based placement and constrained camera geometry, and a strategy-independent error floor persists even under full randomization. Overcoming this floor likely requires either metric-aware training signals or auxiliary depth-based scale reasoning, which we identify as a direction for future work.}

Practitioners targeting deployment robustness across varying real-world lighting and surface conditions should therefore prefer randomized texture assignment, whereas applications in controlled environments, where object materials are consistent and discriminative, may benefit from carefully curated manual textures that preserve material-based separability.

\section{Conclusions and Future Work}
\label{sec:conclusions}

We present a bidirectional sim-to-real framework for industrial object detection, combining 2D-to-3D reconstruction as \acrshort{DA} approach with programmatic \acrshort{DR} via SynthRender, and validate it on \acrshort{iasset}, a new 32-class dataset for \acrshort{sim-to-real} analysis in semi-uncontrolled industrial settings. \deleted{In line with previous research, our experiments conclude that the targeted construction of synthetic variability drives transfer performance more strongly than simply increasing dataset volume.} \added{Within the evaluated low-thousands-image regime and detector families, our experiments suggest that the construction of relevant synthetic variability contributes to transfer performance more than dataset volume alone.}
Furthermore, \acrshort{sim-to-real} transfer was observed to be valid across the three studied model architectures.

Ablation studies indicate that bounded \acrshort{DR}, combining physically plausible scene formation with diverse lighting spectra, nonlinear light intensity sampling, and randomized PBR materials, consistently outperforms baseline rendering. For highly reflective objects such as steel components, texture randomization encourages detectors to rely on transferable geometric cues, improving robustness. \deleted{Accuracy scales with synthetic dataset size but saturates in the low-thousands-image regime, demonstrating strong data efficiency even under semi-uncontrolled I-AsSET conditions.}
\added{Accuracy scales with model size and synthetic dataset size but shows diminishing returns beyond a few thousand images, consistent with prior work\deleted{on the same robotics benchmark}~\cite{Zhu2025icra}. We scope this observation to the evaluated benchmarks, dataset sizes and detector families rather than presenting it as a general scaling law.}


\added{Few-shot fine-tuning further reduces the remaining \acrshort{sim-to-real} gap: adding only one to five real images yields most of the achievable improvement, exceeding 98\% mAP@50 in the best synthetic configurations.} Among \acrshort{real-to-sim} asset strategies, manually curated CAD \deleted{models remain the strongest}\added{with randomized texture remains the strongest (95.1 mAP@50)}, but the differences are graded rather than binary.
\deleted{Multi-view \acrshort{3DGS} object reconstruction closes most of this gap, trailing the best CAD configuration by 4.1 mAP@50 (91.2 vs.\ 95.3) and the fully manual CAD+PBR configuration by 2.2 mAP@50 (91.2 vs.\ 93.4), while clearly outperforming single-image GenAI reconstruction (86.4 mAP@50). We therefore position \acrshort{3DGS} not as equivalent to manual CAD, but as the most
effective \emph{fully automated} alternative when CAD models are unavailable.} \added{Multi-view \acrshort{3DGS} reconstruction with randomized texture matches the fully manual CAD+PBR configuration (92.8 vs.\ 92.5 mAP@50, overlapping 95\% confidence intervals) and trails only the best manual configuration by 2.3 mAP@50, while clearly outperforming single-image GenAI reconstruction ($\approx$86 mAP@50). Notably, on identical \acrshort{3DGS} geometry, randomizing the texture improves performance by 6.2 mAP@50 over the natively scanned texture (86.6 $\rightarrow$ 92.8), indicating that texture randomization, rather than geometry fidelity alone, drives this parity. We therefore position
\acrshort{3DGS} with texture randomization not as equivalent to the best manual result, but as a \emph{fully automated} alternative that reaches manual-modeling parity when CAD models are unavailable.} Scene context behaves differently: replacing randomized backgrounds with \acrshort{3DGS} scans of the \deleted{real} test environment changes performance
by only \deleted{0.1 mAP@50 (93.4 vs.\ 93.3)}\added{0.4 mAP@50 (92.5 vs.\ 92.1)}, indicating that background realism is not a limiting factor for transfer in this setting.


Using the proposed framework and data generation guidelines, we achieve highly competitive performance on two established industrial benchmarks under matched evaluation protocols. These results support a bidirectional \acrshort{sim-to-real} workflow, where real observations refine assets and priors, and simulation provides controlled variability with minimal real supervision as the final calibrator.

\added{We also acknowledge several limitations of the current benchmark that motivate future extensions. First, \acrshort{iasset} comprises 32 object classes; these were deliberately selected to maximize inter-class similarity (including size-only variants and parts that share geometry or material) in order to challenge class discrimination under intra- and inter-class ambiguity rather than to maximize class count, yet catalogs with hundreds of items may exhibit additional similarity patterns that a 32-class study cannot fully capture. Second, images in the test set are acquired with a single industrial RGB-D sensor (Zivid~2 Plus MR60); although SynthRender exposes parameterized camera settings (including the acquisition-camera intrinsics evaluated in our ablation) and \acrshort{iasset} is designed to be extended with new imaging systems, multi-camera capture would further probe robustness to sensor variation. Third, the test scenes contain at most two instances of a given class, so denser same-class arrangements, such as bins of many identical parts, are not represented. \acrshort{iasset} is released as an extensible benchmark, and enlarging the class catalog, the sensor diversity, and the per-class instance counts are natural directions for future work.}
\added{Fourth, our quantitative findings are established under the specific evaluation protocol of this study, namely synthetic training sets of 4{,}000 images, six independent seeds per configuration, the YOLOv8, YOLO11, and DEIM detector families, and the randomization ranges of Table~4. The reported trends should therefore be interpreted within this regime rather than extrapolated to arbitrary dataset scales, detector architectures, or class counts. Fifth, although the evaluated 2D-to-3D reconstruction methods reduce the manual effort of asset preparation, obtaining suitable input imagery can still be a practical bottleneck: 3DGS benefits from overlapping multi-view captures and single-image
GenAI methods infer unobserved geometry, so physical access to the parts, or supplier imagery with sufficient viewpoint coverage and usage rights, remains a prerequisite. Reconstructing geometry directly from manufacturer or product images without physical access is a promising but not yet validated route.}

As future work, we plan to exploit the high-fidelity RGB-D data of \acrshort{iasset} and SynthRender's rendering capabilities to investigate robustness gains in sim-to-real RGB-D object detection, with particular focus on inter-class ambiguity and data efficiency. A per-class analysis of few-shot adaptation behavior, particularly for scale-ambiguous class pairs, remains an open question. This motivates future investigation into active domain adaptation strategies, where \acrshort{SDG} is guided by a small set of unannotated real images and iterative per-class performance feedback, progressively targeting the classes that benefit most from real-world feature grounding while minimizing annotation effort. In addition, extending the test set and performing a comprehensive performance analysis across multiple semi-uncontrolled \acrshort{iasset} scenarios could provide deeper insights into generalization under real-world conditions. Future work will also extend evaluation of the sim-to-real capabilities offered by SynthRender and \acrshort{iasset} on pose estimation and semantic segmentation tasks, as well as evaluating the contribution of the synthetically-trained object detection models in combination to foundation models for downstream applications for which 2D bounding boxes are a valuable input.

\deleted{Future work will also extend the ablation study to combinations of geometry reconstruction methods and texture assignment strategies, including \acrshort{3DGS} with randomized textures, to analyze their combined effect on sim-to-real transfer.}

\added{A further direction concerns the modeling of part-production artifacts and service-induced imperfections. Manually modeled CAD assets provide idealized surfaces that omit the scratches, rust, casting marks, and wear present on the real components in the I-AsSET test set. A growing body of work synthesizes such surface defects and imperfections procedurally within Blender-based pipelines, for example through procedural shader and geometry-node texturing for industrial surface inspection~\cite{schmedemann2022procedural}, an approach also reflected in the inspection-oriented procedural texturing of SynosIs~\cite{fulir2024synosis}. Integrating procedural artifact synthesis into SynthRender's material randomization would allow the synthetic distribution to better reflect the intra-class surface variation observed in real parts, and is a natural extension of the texture-randomization findings reported in this work.}

\section{Acknowledgments}\label{ackn}

This work was funded by the German Federal Ministry for Economic Affairs and Climate Action based on a resolution of the German Bundestag, and financed by the European Union. We gratefully acknowledge FATH GmbH, Festo SE \& Co. KG, GlobalFastener Inc., and McMaster-Carr Supply Co. for granting permission to include selected 3D models in the \acrshort{iasset} dataset. All copyrights remain with their respective owners. The authors also thank Michael Hernández and Heike Wohlfeld, from the ARENA2036 e.V. for providing crucial computational resources to conduct image rendering and model training during our experiments.

\printcredits

\section{Declaration of Generative AI Usage}

During the preparation of this work, the authors used Claude Sonnet 4.6 and GPT-4.1 solely to assist with grammar, written expression, and code debugging. All implementations, experimental procedures, analyses, and scientific interpretations are exclusively the authors' work, for which they take full responsibility.

\bibliographystyle{elsarticle-num}

\bibliography{cas-refs}

\end{document}